\pdfoutput=1

\documentclass[11pt]{article}

\usepackage[final]{acl}

\usepackage{times}
\usepackage{latexsym}

\usepackage[T1]{fontenc}

\usepackage[utf8]{inputenc}

\usepackage{microtype}

\usepackage{inconsolata}

\usepackage{graphicx}
\usepackage{placeins}
\usepackage{float}
\usepackage{titletoc}
 \usepackage{amsmath}
 \usepackage{amssymb}
 
\usepackage{booktabs} 
\usepackage{array}    
\usepackage{graphicx} 
\usepackage{multirow} 
\usepackage{makecell} 
\usepackage{xcolor}
\usepackage{colortbl}
\usepackage{pgfmath}

\definecolor{empcellcolor}{HTML}{E9F7EF}
\newcommand{\empcell}{\cellcolor{empcellcolor}}

\definecolor{palestred_new}{HTML}{cce1ff}      
\definecolor{verylightred_new}{HTML}{aed0ff}   
\definecolor{lightred_new}{HTML}{8EBBFF}      
\definecolor{mediumred_new}{HTML}{8EBBFF}     
\definecolor{deepred_new}{HTML}{6CA6FF}        
\definecolor{darkgreen}{RGB}{0,100,0}

\usepackage{pgfmath}

\definecolor{brightblue}{RGB}{30, 144, 255}
\definecolor{brightgreen}{RGB}{0, 200, 100}
\newcommand{\colordiffcell}[2]{%
  \ifdim #1pt > #2pt
    \ifdim #1pt > \dimexpr#2pt + 0.1pt
      \cellcolor{brightgreen!60}#1
    \else
      \ifdim #1pt > \dimexpr#2pt + 0.05pt
        \cellcolor{brightgreen!30}#1
      \else
        \cellcolor{brightgreen!15}#1
      \fi
    \fi
  \else
    \ifdim #1pt < #2pt
      \ifdim #1pt < \dimexpr#2pt - 0.1pt
        \cellcolor{brightblue!50}#1
      \else
        \ifdim #1pt < \dimexpr#2pt - 0.05pt
          \cellcolor{brightblue!30}#1
        \else
          \cellcolor{brightblue!15}#1
        \fi
      \fi
    \else
      \cellcolor{gray!3}#1
    \fi
  \fi
}

\title{Judging with Many Minds: Do More Perspectives Mean Less Prejudice? On Bias Amplification and Resistance in Multi-Agent Based LLM-as-Judge}

\author{Chiyu Ma \thanks{\protect Denotes equal contribution}, Enpei Zhang\footnotemark[1], Yilun Zhao, Wenjun Liu, Yaning Jia, \\
        \textbf{Peijun Qing,} \textbf{Lin Shi,} \textbf{Arman Cohan,} \textbf{Yujun Yan,} \textbf{Soroush Vosoughi}\\
Dartmouth College, Yale University\\
\small {\{chiyu.ma.gr, enpei.zhang.gr\}@dartmouth.edu, yilun.zhao@yale.edu}}

\begin{document}
\maketitle
\begin{abstract}

LLM-as-Judge has emerged as a scalable alternative to human evaluation, enabling large language models (LLMs) to provide reward signals in trainings. While recent work has explored multi-agent extensions such as multi-agent debate and meta-judging to enhance evaluation quality, the question of how intrinsic biases manifest in these settings remains underexplored. In this study, we conduct a systematic analysis of four diverse bias types: position bias, verbosity bias, chain-of-thought bias, and bandwagon bias. We evaluate these biases across two widely adopted multi-agent LLM-as-Judge frameworks: \emph{Multi-Agent-Debate} and \emph{LLM-as-Meta-Judge}. \textbf{Our results show that debate framework amplifies biases sharply after the initial debate, and this increased bias is sustained in subsequent rounds, while meta-judge approaches exhibit greater resistance.} We further investigate the incorporation of PINE, a leading single-agent debiasing method, as a bias-free agent within these systems. \textbf{The results reveal that this bias-free agent effectively reduces biases in debate settings but provides less benefit in meta-judge scenarios.} Our work provides a comprehensive study of bias behavior in multi-agent LLM-as-Judge systems and highlights the need for targeted bias mitigation strategies in collaborative evaluation settings. Our code and experimental data can be found at \url{https://github.com/Henrymachiyu/Multi_Agent_Judge_Bias}.


\end{abstract}

\section{Introduction}
\begin{figure}[t]
    \centering
    \includegraphics[scale=0.37]{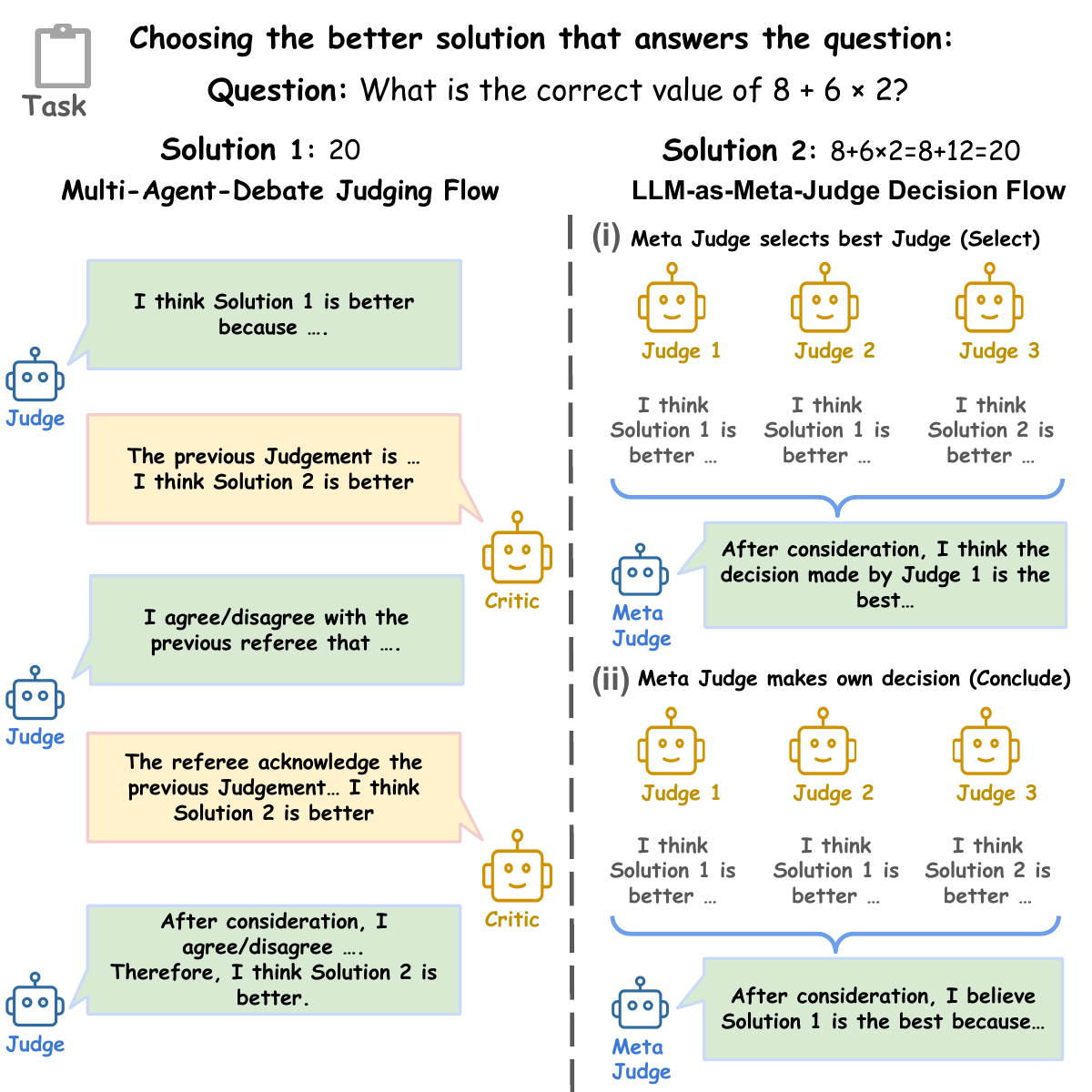}
    \caption{Demonstration of the multi-agent systems analyzed in our study under the LLM-as-Judge paradigm. \textit{Top}: Given responses from two different assistants to the same prompt, the Judge model is tasked with selecting the superior response. \textit{Left:} In the Multi-Agent-Debate framework, a judge provides an initial judgment and then debates with a referee. \textit{Right:} In the Meta-Judge framework, the meta-judge either selects the best judgment or generates its own judgment based on the outputs of multiple judges.}
    \label{intro}
\end{figure}

Large language models (LLMs) have achieved remarkable performance gains in recent years
\cite{kaplan2020scaling, hoffmann2022training, achiam2023gpt}, however, further progress is increasingly constrained by the limited availability of high-quality data and the escalating costs of data acquisition and annotation~\cite{villalobos2024position,shen2025will}. 
As task complexity grows, generating the massive amounts of labeled data required through human effort alone has become impractical. To address this bottleneck, recent research has explored using LLMs themselves as sources of supervision. Notably, \citeauthor{burns2023weak} observed that supervision from weaker models has potential to help elicit the capabilities of stronger models. These findings have further motivated the shift toward automated, scalable feedback mechanisms that reduce reliance on costly human annotation. Within this line of work, the LLM-as-Judge paradigm \cite{chan2023chateval, gu2024survey, chen2024mllm, li2025leveraging, saha2025learning}, which uses LLMs to automatically evaluate responses and provide reward signals, has attracted particular interest due to its adaptability to custom evaluation criteria and its ability to generate human-understandable explanations of decisions.

Building upon this paradigm, recent efforts have begun to investigate whether incorporating multi-agent frameworks—which have proven effective in complex reasoning tasks~\cite{liang2023encouraging, du2023improving, hong2023metagpt, chen2023agentverse}—can further enhance the capabilities of LLM-as-Judge pipelines.
Structures such as multi-agent debate \cite{chan2023chateval, du2023improving, liu2024groupdebate} and meta-reasoning \cite{wu2024meta, li2025leveraging} offer promising avenues for boosting evaluation quality.
Despite these advances, concerns remain. 
Several studies reveal that even single-agent LLM judges have been shown to exhibit intrinsic biases, including position bias and verbosity bias, which can undermine the reliability of their evaluations \cite{huang2023large,shi2024judging,ye2024justice}. 
However, there has been limited focused or systematic investigation into whether such biases persist, diminish, or are amplified in multi-agent settings. 

In response to these concerns, this study systematically examines how intrinsic biases manifest in multi-agent LLM-as-Judge systems. As demonstrated in \autoref{intro}, we investigate two widely adopted paradigms: LLM-as-Judge with \emph{Multi-Agent-Debate} and \emph{LLM-as-Meta-Judge}. For each paradigm, we also evaluate the performance of different LLMs as base evaluators. Our analysis focuses on four diverse types of bias: position bias, verbosity bias, chain-of-thought (CoT) bias, and bandwagon bias, ensuring a comprehensive examination across key dimensions of judgment behavior. Interestingly, our results show that intrinsic biases manifest differently across the two multi-agent frameworks. \textbf{The Multi-Agent-Debate framework amplifies biases sharply after the initial debate, and this increased bias is sustained in subsequent rounds. In contrast, the LLM-as-Meta-Judge approach exhibits greater resistance to these intrinsic biases.}  When the meta-judge selects the best judgment from a set of candidates, the resulting bias levels are comparable to those observed in single-agent settings. However, when the meta-judge generates a new judgment based on the candidate judgments, a more substantial reduction in bias is achieved.

Additionally, we examine whether a leading single-agent bias mitigation strategy can be effectively extended to improve evaluation reliability in more complex multi-agent settings, particularly when direct modifications to the judge model and its prompt are impractical or undesirable. To this end, we incorporate PINE \cite{wang2024eliminating}, a method shown to eliminate position bias through modifications to position embeddings and causal masking, as a \textit{bias-free} agent within a multi-agent LLM-as-Judge system. \textbf{Our results show that introducing a \textit{bias-free} agent into the debate setting yields consistent improvements as the number of conversation rounds increases. In contrast, its effect in the meta-judge framework remains limited, regardless of whether the meta-judge selects an existing judgment or generates its own.} These findings underscore the importance of accounting for bias dynamics in multi-agent evaluation and highlight the need for mitigation strategies specifically tailored to collaborative decision-making settings.

\section{Related Work}


\subsection{LLM-as-Judges}

Recent advances in automated model evaluation have established LLMs as viable alternatives to human annotators. \citeauthor{zheng2023judging} introduced the concept of LLM-as-Judge through MT-Bench, which evaluates alignment between LLM-generated judgments and human preferences on open-ended questions. This line of research has since expanded with benchmarks such as RewardBench \cite{lambert2024rewardbench} and JudgeBench \cite{tan2024judgebench}, which assess model alignment on more complex, reasoning-intensive tasks. In parallel, researchers have explored improved prompting strategies \cite{stahl2024exploring, dubois2023alpacafarm,wang2025mcts}. Beyond prompting, fine-tuning on domain-specific datasets has been proposed to improve evaluation performance and mitigate biases \cite{zhu2023judgelm, park2024offsetbias, liu2024skywork, ye2025learning}. Most recently, EvalPlanner \cite{saha2025learning} introduced a planning-then-judging framework that incorporates agentic structure into the LLM-as-Judge paradigm.


\subsection{Multi-Agent LLM-as-Judges Framework}
Motivated by the effectiveness of multi-agent systems in complex reasoning tasks \cite{liang2023encouraging, du2023improving, hong2023metagpt, chen2023agentverse, huang2023agentcoder, liu2024groupdebate, zhang2025layercraft}, recent studies have applied multi-agent approaches to LLM-based evaluation. Multi-agent debate has emerged as a popular technique, with studies such as employing structured interactions among groups of LLM juries to reach consensus \cite{xu2023towards, li2023prd, verga2024replacing, zhang2025crowd}. In particular, ChatEval \cite{chan2023chateval} aimed to more closely mirror the human evaluation process by facilitating structured debates among agents with diverse roles. In its setup, “general public” agents provided initial judgments, engaged in discussions, and delivered final verdicts, while “referee” agents acted as critics or experts to challenge and refine the reasoning process. Building on this design, our work adopts a similar structure in which judge agents interact with a designated critic who offers counterpoints intended to improve the quality of final decisions.

In parallel, meta-thinking has emerged as a promising direction for enhancing LLM-as-Judge systems \cite{shinn2023reflexion, gao2024meta, sui2025meta, wan2025rema}. Meta-thinking, often described as ``thinking about thinking,'' involves reflecting on and evaluating the reasoning process itself. The concept of a meta-judge was first introduced by \citeauthor{wu2024meta}, where it served as a self-reward model during post-training. Moreover, Self-Rationalization \cite{trivedi2024self} proposed an iterative fine-tuning approach in which the judge improves by comparing and refining its own rationales. More recently, \citeauthor{li2025leveraging} expanded this idea by assigning different LLMs to act as judges and a meta-judge, facilitating collaborative evaluation across diverse models. Our work adopts a similar design with two modes: the meta-judge either selects the best judgment from multiple judges or generates a new judgment generated from their outputs.

\subsection{Intrinsic Biases and Mitigation}
LLMs have been shown to exhibit a variety of intrinsic biases, including position bias, verbosity bias, self-enhancement bias, bandwagon bias, and chain-of-thought bias. These biases have been identified through large-scale evaluations and benchmark studies \cite{zheng2023judging, wang2024large, ye2024justice, shi2024judging, wan2024positional, bao2024autobench, tian2025identifying, li2025preference, gulati2025uncovering}. More recently, CALM \cite{ye2024justice} evaluated twelve types of biases in single-agent LLMs using carefully designed experiments. Our study closely follows the bias evaluation procedures introduced by \citeauthor{ye2024justice}. However, there is limited understanding of how these biases behave in multi-agent settings. While \citeauthor{wu2024meta} observed that certain biases may intensify during iterative training, \citeauthor{li2023prd} found that collaborative interactions among agents can reduce specific biases such as position and self-enhancement. These contrasting observations highlight a critical gap in understanding bias dynamics in collaborative contexts, a gap our study seeks to address.

Among the various biases, position bias has received particular attention due to its substantial impact across a wide range of tasks. Mitigation strategies to date have included data augmentation combined with fine-tuning \cite{zhu2023judgelm, park2024offsetbias} and prompt alignment techniques \cite{li2023split}. More recently, the PINE framework \cite{wang2024eliminating} was introduced to completely eliminate position bias by modifying causal masking and positional embeddings. While these methods have demonstrated strong performance, their effectiveness in multi-agent contexts remains largely unexplored. Our work seeks to fill this gap by leveraging PINE as a \textit{bias-free} agent to assess its impact on position bias in collaborative evaluation scenarios. This is particularly important in settings where modifying the prompts or internal structure of judge models is impractical or undesirable.

\section{Method}


\subsection{Evaluative Settings}

In this study, we adopt the score-based pairwise comparison setting to stick with that of ChatEval \cite{chan2023chateval}. To ensure consistency, we apply this evaluation setting across all multi-agent LLM-as-Judge experiments and bias assessments. Similar to the single LLM-as-Judge setting, each model in the multi-agent LLM-as-Judge framework is tasked with evaluating two candidate solutions generated by different LLMs for a given prompt, although with different roles. The judges assign numerical scores and provide feedback for each solution, with the higher-scored solution subsequently selected as superior. We assign the solution as a tie, when the scores are the same. For clarity, \autoref{intro} illustrates only the pairwise comparison process. In the actual study, however, each judge model in the framework are prompted to provide explicit scores for each solution. The framework then determines and outputs the final verdict. Importantly, the identities of the LLMs that produced the candidate solutions are not disclosed to the multi-agent judges.

\subsection{Multi-Agent-Debate}
In the Multi-Agent-Debate setting,
we adopt and implement the “General Public and Critic” structure, as originally proposed in ChatEval \cite{chan2023chateval}, and widely used in several recent studies~\cite{feng2024m,arif2024fellowship,zhang2025if}. The General Public agent is analogous to the judge model in the single-agent LLM-as-Judge setting. It is tasked with making an initial judgment, which includes the scores and reasoning for both solutions. The Critic agent, in contrast, first reads and comprehends the evaluation task and the judgment provided by the General Public agent. It then offers its critique of the initial judgment along with its own independent assessment. To mirror real-world debate scenarios, the Critic agent is encouraged to provide a different judgment from that of the General Public agent. In subsequent rounds of conversation, the General Public agent must consider the Critic’s feedback and may choose either to revise or to maintain its original judgment. The Critic agent continues to generate additional critiques and independent assessments based on the evolving conversation. At the end of the debate, we use the judgment of the General Public agent as the final verdict. A demonstration of the debate flow is shown in the left panel of \autoref{intro}. The prompts used for the General Public and Critic agents are provided in Appendix \ref{debate_prompt_examples}.

We focus on the “General Public and Critic” structure rather than incorporating additional agents such as Analyst or Supporter agents, as proposed by \citeauthor{chan2023chateval}. This is because adding such roles, which typically represent mutually aligned perspectives, tends to offer limited additional judgment diversity and may weaken the intended role of the Critic agent within the debate. Additionally, our preliminary study finds that introducing more agents and extending the length of conversations increases the difficulty of instruction following, particularly for medium-sized LLMs. Taking these considerations into account, and aiming to isolate and measure the effect of actual debate between agents within the framework, our study specifically focuses on the “General Public and Critic” configuration within the Multi-Agent-Debate LLM-as-Judge framework.

\subsection{LLM-as-Meta-Judge}
In the LLM-as-Meta-Judge setting, we focus on the fundamental concept of \textit{“Thinking about thinking.”} Inspired by the frameworks proposed by \citeauthor{li2025leveraging} and \citeauthor{wu2024meta}, we adopt diverse LLMs as judges, each providing an independent judgment to form a judgment pool based on the evaluation task. We further examine two variations of the Meta-Judge framework, as shown in the right panel of \autoref{intro}. In the “Select” setting, the Meta-Judge selects the best judgment from the pool and uses it as the final verdict. This setting evaluates the Meta-Judge’s ability to discern and identify the highest-quality answer among diverse alternatives, testing its competence as an evaluator and ranker of peer outputs. In contrast, the “Conclude” setting requires the Meta-Judge to provide its own judgment as the final verdict based on the judgment pool. This setting assesses the Meta-Judge’s ability to integrate, reason over, and reconcile potentially conflicting or complementary information, acting as a collaborative decision-maker. It is also similar to the “summarization” role that was widely used in group discussion settings \cite{liu2024groupdebate,li2025leveraging}. The prompts used for the Meta Judges are provided in Appendix~\ref{meta-judge_prompt_examples}. 

We focus on the most basic structure of the Meta-Judges framework, intentionally avoiding techniques such as Rubric Design and Majority Rubric proposed by \citeauthor{li2025leveraging}, as our primary goal is to measure how bias manifests within the framework itself. The inclusion of such techniques could introduce confounding effects that obscure this measurement. Additionally, position bias of the Meta-Judge itself could act as a confounder. To mitigate this, we shuffle the positions of the judge model outputs within the judgment pool. Furthermore, we gradually expand the size of the judge pool to examine the effect of pool size on the observed biases.

\subsection{Biases and Measurement}
\label{bias_intro}
\begin{figure}[ht]
    \centering
    \includegraphics[scale=0.37]{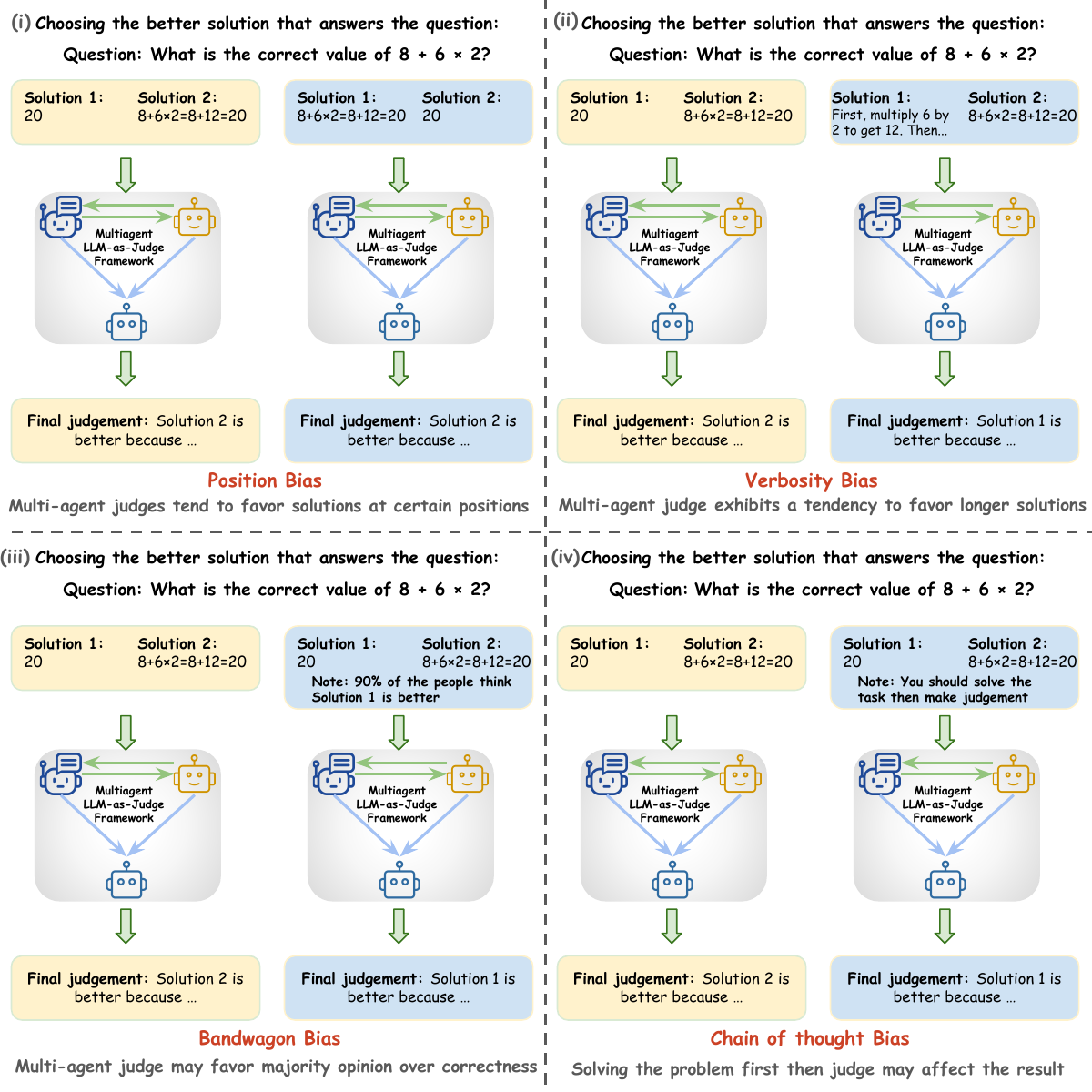}
    \caption{Demonstration of the biases measured in our study. We kept the multi-agent LLM-as-Judge framework unchanged throughout each evaluation process. Biases were introduced and measured solely by modifying the prompts, as shown.
    }
    \label{bias}
\end{figure}

In our study, we primarily focus on four types of intrinsic bias: position bias, verbosity bias, bandwagon bias, and chain-of-thought bias. Illustrative examples of these biases are provided in \autoref{bias}. We selected these biases because they span some of the most critical and generalizable dimensions of intrinsic model behavior: input ordering, response length, reasoning stability, and social conformity. To measure them, we follow the procedures proposed by CALM \cite{ye2024justice}, which assess bias by comparing the consistency of judgments before and after specific prompt modifications designed to elicit each bias. In the Multi-Agent LLM-as-Judge framework, all agents are exposed to these prompt modifications when measuring biases. Examples of these prompts can be found in Appendix~\ref{bias_prompts}. The consistency rate is defined as:
\begin{equation}
\text{CR} = \frac{1}{N} \sum_{i=1}^{N} \mathbb{I}(y^i = y^i_{\text{modified}})
\end{equation}
where $N$ denotes the number of samples, $y^i$ represents the original judgment for sample $i$, and $y^i_{\text{modified}}$ represents the judgment after prompt modification for the same sample. 

\noindent \textbf{Position bias} reflects the tendency of LLM judges to favor solutions based on their positions rather than content quality. To evaluate this, we alter the order of candidate solutions in the prompt and examined whether the preference shifted as a result. 

\noindent  \textbf{Verbosity bias} occurs when judges prefer longer responses over shorter, equally or more accurate alternatives. We generate extended responses following the approach of \citeauthor{ye2024justice}. In their study, they verified through human evaluation that such modifications introduced minimal unintended improvement in answer quality. We then measure the degree to which LLM judgments shifted toward the longer version. 

\noindent \textbf{Bandwagon bias} describes the inclination of LLM judges to align with majority opinions. To measure this, we add a statement to the prompt indicating that most people favor one of the candidate solutions, and compare the judgments with and without this statement to assess any shift in preference.

\noindent \textbf{Chain-of-thought (CoT) bias} refers to changes in evaluation outcomes when judges are first prompted to reason through the task independently before assessing candidate answers. We measure this by comparing judgments made with and without the additional reasoning step, observing how the internal deliberation influence final evaluations.

\section{Experiment}



\definecolor{verylightgray}{gray}{0.9}
\begin{table*}[ht!]
    \centering
    \setlength{\tabcolsep}{10pt}
    \scalebox{0.55}{
    \begin{tabular}{lc|c|llll|llll}
    \toprule
    \multirow{2}{*}{\textbf{Judge}} & \multirow{2}{*}{\textbf{Critic}} & \multirow{2}{*}{\textbf{Round}} & \multicolumn{4}{c}{\textbf{MTBench}} & \multicolumn{4}{c}{\textbf{CALM-Alignment}} \\
    \cline{4-11}
     &  &  & $Pos.$ & $Verbo.$ & $Band.$ & $Cot$ & $Pos.$ & $Verbo.$ & $Band.$ & $Cot$ \\
    \hline
     &  &\cellcolor{verylightgray} 0 &\cellcolor{verylightgray} 0.772 &\cellcolor{verylightgray} 0.736 &\cellcolor{verylightgray} 0.728 &\cellcolor{verylightgray} 0.795 &\cellcolor{verylightgray} 0.723 &\cellcolor{verylightgray} 0.732 &\cellcolor{verylightgray} 0.689 &\cellcolor{verylightgray} 0.826 \\
    \textbf{R1-Distilled-} & \textbf{R1-Distilled-} & 1 & \cellcolor{deepred_new}0.696 $\Downarrow$ & \cellcolor{deepred_new}0.654 $\Downarrow$ & \cellcolor{mediumred_new}0.687 $\downarrow$ & \cellcolor{deepred_new}0.723 $\Downarrow$ & \cellcolor{deepred_new}0.670 $\Downarrow$ & \cellcolor{deepred_new}0.613 $\Downarrow$ & \cellcolor{mediumred_new}0.645 $\Downarrow$ & \cellcolor{deepred_new}0.745 $\Downarrow$ \\
    \textbf{Qwen-32B} & \textbf{Qwen-32B} & 2 & \cellcolor{palestred_new}0.700 $\uparrow$ & \cellcolor{lightred_new}0.652 $\downarrow$ & \cellcolor{lightred_new}0.681 $\downarrow$ & \cellcolor{palestred_new}0.729 $\uparrow$ & \cellcolor{palestred_new}0.672 $\uparrow$ & \cellcolor{palestred_new}0.620 $\uparrow$ & \cellcolor{mediumred_new}0.633 $\downarrow$ & \cellcolor{palestred_new}0.756 $\uparrow$ \\
     &  & 3 & \cellcolor{palestred_new}0.702 $\uparrow$ & \cellcolor{palestred_new}0.658 $\uparrow$ & \cellcolor{palestred_new}0.685 $\uparrow$ & \cellcolor{verylightred_new}0.729 $\rightarrow$ & \cellcolor{palestred_new}0.677 $\uparrow$ & \cellcolor{palestred_new}0.633 $\uparrow$ & \cellcolor{lightred_new}0.626 $\downarrow$ & \cellcolor{lightred_new}0.754 $\downarrow$ \\
    \hline

     &  &\cellcolor{verylightgray} 0 &\cellcolor{verylightgray} 0.772 &\cellcolor{verylightgray} 0.736 &\cellcolor{verylightgray} 0.728 &\cellcolor{verylightgray} 0.795 &\cellcolor{verylightgray} 0.723 &\cellcolor{verylightgray} 0.732 &\cellcolor{verylightgray} 0.689 &\cellcolor{verylightgray} 0.826 \\
    \textbf{R1-Distilled-} & \textbf{GPT4o-mini} & 1 & \cellcolor{deepred_new}0.621 $\Downarrow$ & \cellcolor{deepred_new}0.606 $\Downarrow$ & \cellcolor{deepred_new}0.656 $\Downarrow$ & \cellcolor{deepred_new}0.660 $\Downarrow$ & \cellcolor{deepred_new}0.626 $\Downarrow$ & \cellcolor{deepred_new}0.608 $\Downarrow$ & \cellcolor{deepred_new}0.624 $\Downarrow$ & \cellcolor{deepred_new}0.688 $\Downarrow$ \\
    \textbf{Qwen-32B} &  & 2 & \cellcolor{palestred_new}0.637 $\uparrow$ & \cellcolor{palestred_new}0.608 $\rightarrow$ & \cellcolor{mediumred_new}0.640 $\downarrow$ & \cellcolor{lightred_new}0.656 $\downarrow$ & \cellcolor{lightred_new}0.617 $\downarrow$ & \cellcolor{verylightred_new}0.608 $\rightarrow$ & \cellcolor{palestred_new}0.642 $\uparrow$ & \cellcolor{palestred_new}0.699 $\uparrow$ \\
     &  & 3 & \cellcolor{palestred_new}0.642 $\uparrow$ & \cellcolor{palestred_new}0.610 $\uparrow$ & \cellcolor{palestred_new}0.644 $\uparrow$ & \cellcolor{palestred_new}0.665 $\uparrow$ & \cellcolor{palestred_new}0.629 $\uparrow$ & \cellcolor{lightred_new}0.604 $\downarrow$ & \cellcolor{verylightred_new}0.642 $\rightarrow$ & \cellcolor{lightred_new}0.695 $\downarrow$ \\
    \hline

     &  &\cellcolor{verylightgray} 0 &\cellcolor{verylightgray} 0.772 &\cellcolor{verylightgray} 0.736 &\cellcolor{verylightgray} 0.728 &\cellcolor{verylightgray} 0.795 &\cellcolor{verylightgray} 0.723 &\cellcolor{verylightgray} 0.732 &\cellcolor{verylightgray} 0.689 &\cellcolor{verylightgray} 0.826 \\
    \textbf{R1-Distilled-} & \textbf{Llama-3.3-70B-} & 1 & \cellcolor{mediumred_new}0.754 $\downarrow$ & \cellcolor{deepred_new}0.660 $\Downarrow$ & \cellcolor{lightred_new}0.721 $\downarrow$ & \cellcolor{mediumred_new}0.752 $\downarrow$ & \cellcolor{mediumred_new}0.697 $\downarrow$ & \cellcolor{deepred_new}0.645 $\Downarrow$ & \cellcolor{mediumred_new}0.665 $\Downarrow$ & \cellcolor{deepred_new}0.745 $\downarrow$ \\
    \textbf{Qwen-32B} & \textbf{Instruct-Turbo} & 2 & \cellcolor{verylightred_new}0.754 $\rightarrow$ & \cellcolor{lightred_new}0.656 $\downarrow$ & \cellcolor{palestred_new}0.727 $\uparrow$ & \cellcolor{palestred_new}0.767 $\uparrow$ & \cellcolor{lightred_new}0.692 $\downarrow$ & \cellcolor{palestred_new}0.656 $\uparrow$ & \cellcolor{palestred_new}0.667 $\uparrow$ & \cellcolor{mediumred_new}0.733 $\downarrow$ \\
     &  & 3 & \cellcolor{verylightred_new}0.754 $\rightarrow$ & \cellcolor{palestred_new}0.658 $\uparrow$ & \cellcolor{palestred_new}0.731 $\uparrow$ & \cellcolor{lightred_new}0.765 $\downarrow$ & \cellcolor{verylightred_new}0.692 $\rightarrow$ & \cellcolor{palestred_new}0.670 $\uparrow$ & \cellcolor{palestred_new}0.672 $\uparrow$ & \cellcolor{verylightred_new}0.733 $\rightarrow$ \\
    \hline

     &  &\cellcolor{verylightgray} 0 &\cellcolor{verylightgray} 0.772 &\cellcolor{verylightgray} 0.736 &\cellcolor{verylightgray} 0.728 &\cellcolor{verylightgray} 0.795 &\cellcolor{verylightgray} 0.723 &\cellcolor{verylightgray} 0.732 &\cellcolor{verylightgray} 0.689 &\cellcolor{verylightgray} 0.826 \\
    \textbf{R1-Distilled-} & \textbf{DeepSeek-V3} & 1 & \cellcolor{deepred_new}0.594 $\Downarrow$ & \cellcolor{deepred_new}0.640 $\Downarrow$ & \cellcolor{deepred_new}0.642 $\Downarrow$ & \cellcolor{deepred_new}0.644 $\Downarrow$ & \cellcolor{deepred_new}0.581 $\Downarrow$ & \cellcolor{deepred_new}0.649 $\Downarrow$ & \cellcolor{deepred_new}0.592 $\Downarrow$ & \cellcolor{deepred_new}0.626 $\Downarrow$ \\
    \textbf{Qwen-32B} &  & 2 & \cellcolor{palestred_new}0.598 $\uparrow$ & \cellcolor{palestred_new}0.656 $\uparrow$ & \cellcolor{palestred_new}0.660 $\uparrow$ & \cellcolor{lightred_new}0.640 $\downarrow$ & \cellcolor{lightred_new}0.574 $\downarrow$ & \cellcolor{palestred_new}0.658 $\uparrow$ & \cellcolor{palestred_new}0.595 $\uparrow$ & \cellcolor{palestred_new}0.640 $\uparrow$ \\
     &  & 3 & \cellcolor{palestred_new}0.617 $\uparrow$ & \cellcolor{mediumred_new}0.644 $\downarrow$ & \cellcolor{palestred_new}0.675 $\uparrow$ & \cellcolor{palestred_new}0.646 $\uparrow$ & \cellcolor{palestred_new}0.588 $\uparrow$ & \cellcolor{lightred_new}0.651 $\downarrow$ & \cellcolor{palestred_new}0.608 $\uparrow$ & \cellcolor{palestred_new}0.651 $\uparrow$ \\
    \hline

     &  &\cellcolor{verylightgray} 0 &\cellcolor{verylightgray} 0.793 &\cellcolor{verylightgray} 0.716 &\cellcolor{verylightgray} 0.746 &\cellcolor{verylightgray} 0.835 &\cellcolor{verylightgray} 0.793 &\cellcolor{verylightgray} 0.789 &\cellcolor{verylightgray} 0.740 &\cellcolor{verylightgray} 0.869 \\
    \textbf{GPT4o-mini} & \textbf{GPT4o-mini} & 1 & \cellcolor{deepred_new}0.667 $\Downarrow$ & \cellcolor{deepred_new}0.637 $\Downarrow$ & \cellcolor{deepred_new}0.667 $\Downarrow$ & \cellcolor{deepred_new}0.731 $\Downarrow$ & \cellcolor{deepred_new}0.663 $\Downarrow$ & \cellcolor{deepred_new}0.704 $\Downarrow$ & \cellcolor{deepred_new}0.663 $\Downarrow$ & \cellcolor{deepred_new}0.763 $\Downarrow$ \\
     &  & 2 & \cellcolor{mediumred_new}0.648 $\downarrow$ & \cellcolor{mediumred_new}0.617 $\downarrow$ & \cellcolor{palestred_new}0.692 $\uparrow$ & \cellcolor{palestred_new}0.740 $\uparrow$ & \cellcolor{palestred_new}0.681 $\uparrow$ & \cellcolor{verylightred_new}0.704 $\rightarrow$ & \cellcolor{palestred_new}0.683 $\uparrow$ & \cellcolor{palestred_new}0.781 $\uparrow$ \\
     &  & 3 & \cellcolor{verylightred_new}0.648 $\rightarrow$ & \cellcolor{mediumred_new}0.598 $\downarrow$ & \cellcolor{palestred_new}0.704 $\uparrow$ & \cellcolor{mediumred_new}0.719 $\downarrow$ & \cellcolor{palestred_new}0.692 $\uparrow$ & \cellcolor{palestred_new}0.708 $\uparrow$ & \cellcolor{palestred_new}0.704 $\uparrow$ & \cellcolor{palestred_new}0.784 $\uparrow$ \\
    \hline

     &  &\cellcolor{verylightgray} 0 &\cellcolor{verylightgray} 0.793 &\cellcolor{verylightgray} 0.716 &\cellcolor{verylightgray} 0.746 &\cellcolor{verylightgray} 0.835 &\cellcolor{verylightgray} 0.793 &\cellcolor{verylightgray} 0.789 &\cellcolor{verylightgray} 0.740 &\cellcolor{verylightgray} 0.869 \\
    \textbf{GPT4o-mini} & \textbf{Llama-3.3-70B-} & 1 & \cellcolor{mediumred_new}0.744 $\downarrow$ & \cellcolor{mediumred_new}0.681 $\downarrow$ & \cellcolor{mediumred_new}0.735 $\downarrow$ & \cellcolor{deepred_new}0.777 $\downarrow$ & \cellcolor{deepred_new}0.715 $\downarrow$ & \cellcolor{deepred_new}0.708 $\downarrow$ & \cellcolor{mediumred_new}0.697 $\downarrow$ & \cellcolor{deepred_new}0.797 $\downarrow$ \\
     & \textbf{Instruct-Turbo} & 2 & \cellcolor{palestred_new}0.762 $\uparrow$ & \cellcolor{palestred_new}0.692 $\uparrow$ & \cellcolor{palestred_new}0.767 $\uparrow$ & \cellcolor{lightred_new}0.775 $\downarrow$ & \cellcolor{palestred_new}0.743 $\uparrow$ & \cellcolor{lightred_new}0.699 $\downarrow$ & \cellcolor{palestred_new}0.702 $\uparrow$ & \cellcolor{palestred_new}0.804 $\uparrow$ \\
     &  & 3 & \cellcolor{palestred_new}0.765 $\uparrow$ & \cellcolor{palestred_new}0.712 $\uparrow$ & \cellcolor{lightred_new}0.760 $\downarrow$ & \cellcolor{palestred_new}0.787 $\uparrow$ & \cellcolor{mediumred_new}0.722 $\downarrow$ & \cellcolor{palestred_new}0.713 $\uparrow$ & \cellcolor{palestred_new}0.718 $\uparrow$ & \cellcolor{lightred_new}0.795 $\downarrow$ \\
    \hline

     &  &\cellcolor{verylightgray} 0 &\cellcolor{verylightgray} 0.814 &\cellcolor{verylightgray} 0.720 &\cellcolor{verylightgray} 0.783 &\cellcolor{verylightgray} 0.821 &\cellcolor{verylightgray} 0.823 &\cellcolor{verylightgray} 0.773 &\cellcolor{verylightgray} 0.768 &\cellcolor{verylightgray} 0.839 \\
    \textbf{Llama-3.3-70B-} & \textbf{GPT4o-mini} & 1 & \cellcolor{mediumred_new}0.765 $\downarrow$ & \cellcolor{deepred_new}0.646 $\Downarrow$ & \cellcolor{mediumred_new}0.771 $\downarrow$ & \cellcolor{deepred_new}0.696 $\Downarrow$ & \cellcolor{deepred_new}0.706 $\Downarrow$ & \cellcolor{deepred_new}0.670 $\Downarrow$ & \cellcolor{lightred_new}0.761 $\downarrow$ & \cellcolor{deepred_new}0.747 $\Downarrow$ \\
    \textbf{Instruct-Turbo} &  & 2 & \cellcolor{lightred_new}0.756 $\downarrow$ & \cellcolor{lightred_new}0.644 $\downarrow$ & \cellcolor{mediumred_new}0.731 $\Downarrow$ & \cellcolor{palestred_new}0.698 $\uparrow$ & \cellcolor{mediumred_new}0.692 $\downarrow$ & \cellcolor{mediumred_new}0.638 $\downarrow$ & \cellcolor{deepred_new}0.711 $\Downarrow$ & \cellcolor{palestred_new}0.756 $\uparrow$ \\
     &  & 3 & \cellcolor{palestred_new}0.775 $\uparrow$ & \cellcolor{palestred_new}0.679 $\uparrow$ & \cellcolor{palestred_new}0.746 $\uparrow$ & \cellcolor{palestred_new}0.721 $\uparrow$ & \cellcolor{palestred_new}0.708 $\uparrow$ & \cellcolor{palestred_new}0.656 $\uparrow$ & \cellcolor{lightred_new}0.702 $\downarrow$ & \cellcolor{lightred_new}0.749 $\downarrow$ \\
    \hline

     &  &\cellcolor{verylightgray} 0 &\cellcolor{verylightgray} 0.814 &\cellcolor{verylightgray} 0.720 &\cellcolor{verylightgray} 0.783 &\cellcolor{verylightgray} 0.821 &\cellcolor{verylightgray} 0.823 &\cellcolor{verylightgray} 0.773 &\cellcolor{verylightgray} 0.768 &\cellcolor{verylightgray} 0.839 \\
    \textbf{Llama-3.3-70B-} & \textbf{Llama-3.3-70B-} & 1 & \cellcolor{mediumred_new}0.773 $\downarrow$ & \cellcolor{deepred_new}0.665 $\Downarrow$ & \cellcolor{mediumred_new}0.756 $\downarrow$ & \cellcolor{deepred_new}0.692 $\Downarrow$ & \cellcolor{deepred_new}0.713 $\Downarrow$ & \cellcolor{deepred_new}0.690 $\downarrow$ & \cellcolor{mediumred_new}0.736 $\downarrow$ & \cellcolor{deepred_new}0.688 $\Downarrow$ \\
    \textbf{Instruct-Turbo} & \textbf{Instruct-Turbo} & 2 & \cellcolor{mediumred_new}0.733 $\downarrow$ & \cellcolor{lightred_new}0.662 $\downarrow$ & \cellcolor{mediumred_new}0.742 $\downarrow$ & \cellcolor{palestred_new}0.744 $\uparrow$ & \cellcolor{lightred_new}0.704 $\downarrow$ & \cellcolor{mediumred_new}0.677 $\downarrow$ & \cellcolor{mediumred_new}0.692 $\Downarrow$ & \cellcolor{palestred_new}0.711 $\uparrow$ \\
     &  & 3 & \cellcolor{lightred_new}0.727 $\downarrow$ & \cellcolor{lightred_new}0.656 $\downarrow$ & \cellcolor{palestred_new}0.760 $\uparrow$ & \cellcolor{palestred_new}0.760 $\uparrow$ & \cellcolor{mediumred_new}0.690 $\downarrow$ & \cellcolor{mediumred_new}0.656 $\downarrow$ & \cellcolor{palestred_new}0.708 $\uparrow$ & \cellcolor{palestred_new}0.715 $\uparrow$ \\
    \hline

     &  & \cellcolor{verylightgray} 0 & \cellcolor{verylightgray} 0.821 & \cellcolor{verylightgray} 0.746 & \cellcolor{verylightgray} 0.748 & \cellcolor{verylightgray} 0.883 & \cellcolor{verylightgray} 0.809 & \cellcolor{verylightgray} 0.718 & \cellcolor{verylightgray} 0.688 & \cellcolor{verylightgray} 0.868 \\
    \textbf{DeepSeek-V3} & \textbf{DeepSeek-V3} & 1 & \cellcolor{deepred_new}0.715 $\Downarrow$ & \cellcolor{deepred_new}0.687 $\Downarrow$ & \cellcolor{mediumred_new}0.721 $\downarrow$ & \cellcolor{deepred_new}0.758 $\Downarrow$ & \cellcolor{deepred_new}0.665 $\Downarrow$ & \cellcolor{deepred_new}0.658 $\Downarrow$ & \cellcolor{mediumred_new}0.645 $\Downarrow$ & \cellcolor{deepred_new}0.738 $\Downarrow$ \\
     &  & 2 & \cellcolor{palestred_new}0.717 $\uparrow$ & \cellcolor{palestred_new}0.696 $\uparrow$ & \cellcolor{palestred_new}0.733 $\uparrow$ & \cellcolor{mediumred_new}0.733 $\downarrow$ & \cellcolor{palestred_new}0.667 $\uparrow$ & \cellcolor{mediumred_new}0.613 $\downarrow$ & \cellcolor{palestred_new}0.649 $\uparrow$ & \cellcolor{mediumred_new}0.713 $\downarrow$ \\
     &  & 3 & \cellcolor{verylightred_new}0.717 $\rightarrow$ & \cellcolor{palestred_new}0.704 $\uparrow$ & \cellcolor{mediumred_new}0.719 $\downarrow$ & \cellcolor{mediumred_new}0.717 $\downarrow$ & \cellcolor{palestred_new}0.679 $\uparrow$ & \cellcolor{palestred_new}0.654 $\uparrow$ & \cellcolor{lightred_new}0.642 $\downarrow$ & \cellcolor{lightred_new}0.704 $\downarrow$ \\
    \bottomrule
    \end{tabular}
    }
    \caption{Consistency scores for each bias (Positivity, Verbosity, Bandwagon, Cot) across three rounds of multi-agent debate, shown for different Judge–Critic model pairs on MTBench and CALM-Alignment. Arrows indicate trends across rounds (
    \colorbox{deepred_new}{$\Downarrow$ sharp decrease}, 
    \colorbox{lightred_new}{$\downarrow$ decrease},
    \colorbox{verylightred_new}{$\rightarrow$ stable}, 
    \colorbox{palestred_new}{$\uparrow$ increase}).
    Round 0 indicates the consistency rate of judges without debate. The higher consistency rate the better resistance to biases. 
    }
    \label{MAD_result}
    \vspace{-3mm}
\end{table*}

\subsection{Evaluation Dataset}
We conduct our experiments on two benchmarks: MT-Bench \cite{zheng2023judging} and the Alignment dataset from CALM \cite{ye2024justice}, which we refer to as CALM-Alignment. MT-Bench consists of responses from 30 LLMs to each of 8 tasks, with 10 questions per task. For each question, we sample 60 response pairs per task along with their corresponding questions, yielding a total of 480 samples. This sampling strategy balances computational cost with comparability to the CALM-Alignment dataset. CALM-Alignment, sampled from various DPO datasets, contains 439 samples. Each response pair includes a chosen and a rejected label based on human feedback. More details on the sub-sampled MT-Bench and CALM-Alignment datasets are provided in Appendix~\ref{benchmark_details}. We apply prompt modifications as described in Sec. \ref{bias_intro}. Notably, since the CALM-Alignment dataset includes full labels, we specifically extend the solution for the rejected responses (which always appear as the second solution in the pair in original order). This allows us to measure how many previously correct judgments are changed to incorrect ones. For consistency, we also apply the prompt extension to the second solution in each pair in MT-Bench. To further support our findings, we also include results on Reward Bench \cite{lambert2024rewardbench} in \autoref{Reward_Bench_result}. 


\subsection{Experiments Setup}
In our experiments, we evaluate four LLMs: R1-Distilled-Qwen 32B \cite{deepseekai2025deepseekr1incentivizingreasoningcapability}, LLaMA-3.3-70B-Instruct-Turbo (abbreviated as Llama-3.3-70B) \cite{grattafiori2024llama}, 4o-mini \cite{achiam2023gpt}, and DeepSeek-V3 \cite{liu2024deepseek}. For the multi-agent debate experiment, we adopt a reference-focused strategy, with R1-Distilled-Qwen 32B most frequently serving as the Judge model and other models acting as Critics. To explore cross-model dynamics, we also conduct selective pairings with LLaMA 3.3 70B, GPT-4o-mini, and DeepSeek-V3. Both self-play and cross-model pairings are included, as summarized in \autoref{MAD_result}. Each debate is limited to three rounds per sample to ensure consistency across experiments while keeping computational costs manageable. We do not account for early convergence in judgment between the agent and the judge, as allowing variable-length debates would reduce comparability across different model combinations. For the Meta-Judge setting, GPT-4o-mini, LLaMA-3.3-70B , and DeepSeek-V3 serve as the Meta-Judge. We use R1-Distilled-Qwen 32B as reference in the judgement pool, and thus we do not use it as a Meta-Judge. To assess how the size of the judgment pool affects biases, we vary the pool from two models up to the full set of models providing initial judgments. The Meta-Judge then either selects from these judgments or concludes its own, based on the judgements from the judgment pool. The full study is summarized in \autoref{Meta_result}.

To assess the impact of a bias-free agent, we apply the PINE method \cite{wang2024eliminating} to Qwen1.5-32B-Chat \cite{qwen}, strictly following the setup used in the original work. We then deploy it in debates against GPT-4o-mini (which exhibited the largest drop in position bias consistency) and LLaMA 3.3 70B (which showed the least drop). Leveraging PINE’s position bias elimination, we investigate whether a bias-free agent can reduce bias in judge models through interaction. In our experiments, the bias-free agent serves as a critic in the Multi-Agent Debate and is included in the judgment pool under Meta-Judge settings. This experiment targets scenarios where modifying the judge model or its prompt is impractical, such as self-reward settings during training \cite{zhang2024rest, wu2024meta}. Full results are presented in \autoref{PINE_debate} and \autoref{PINE_meta}. Case studies and examples can be found in \autoref{Case_study}. 


\begin{table*}[ht!]
    \centering
    \setlength{\tabcolsep}{8pt}
    \scalebox{0.56}{
    \begin{tabular}{lllll|llll|llll}
    \toprule
     \multicolumn{5}{c|}{\textbf{Meta Judge: 4o-mini}} & \multicolumn{4}{c|}{\textbf{MT-Bench}} & \multicolumn{4}{c}{\textbf{CALM-Alignment}}  \\
    \cline{6-13}
    R1-Distilled-Qwen 32B & 4o-mini & Llama3.3 70B & DeepSeek-V3  & $Mode$ & $Pos.$ & $Verb.$ & $Band.$ & $Cot.$ & $Pos.$ & $Verb.$ & $Band.$ & $Cot.$ \\
    \hline
      No & No &  No  & No & Judge &
    0.793 & 0.716 & 0.746 & 0.835 & 0.793 & 0.789 & 0.740 & 0.869 \\
    \empcell \textbf{Yes} & \empcell \textbf{Yes} & No & No & Select &
    \colordiffcell{0.817}{0.793} & \colordiffcell{0.737}{0.716} & \colordiffcell{0.735}{0.746} & \colordiffcell{0.783}{0.835} &
    \colordiffcell{0.793}{0.793} & \colordiffcell{0.772}{0.789} & \colordiffcell{0.677}{0.740} & \colordiffcell{0.861}{0.869} \\
    \empcell \textbf{Yes} & No & \empcell \textbf{Yes} & No & Select &
    \colordiffcell{0.798}{0.793} & \colordiffcell{0.706}{0.716} & \colordiffcell{0.746}{0.746} & \colordiffcell{0.798}{0.835} &
    \colordiffcell{0.811}{0.793} & \colordiffcell{0.745}{0.789} & \colordiffcell{0.754}{0.740} & \colordiffcell{0.841}{0.869} \\
    \empcell \textbf{Yes} & No & No & \empcell \textbf{Yes} & Select &
    \colordiffcell{0.806}{0.793} & \colordiffcell{0.756}{0.716} & \colordiffcell{0.706}{0.746} & \colordiffcell{0.819}{0.835} &
    \colordiffcell{0.804}{0.793} & \colordiffcell{0.715}{0.789} & \colordiffcell{0.658}{0.740} & \colordiffcell{0.866}{0.869} \\
    \empcell \textbf{Yes} & No & \empcell \textbf{Yes} & \empcell \textbf{Yes} & Select &
    \colordiffcell{0.835}{0.793} & \colordiffcell{0.721}{0.716} & \colordiffcell{0.733}{0.746} & \colordiffcell{0.844}{0.835} &
    \colordiffcell{0.836}{0.793} & \colordiffcell{0.731}{0.789} & \colordiffcell{0.731}{0.740} & \colordiffcell{0.852}{0.869} \\
    \empcell \textbf{Yes} & \empcell \textbf{Yes} & \empcell \textbf{Yes} & \empcell \textbf{Yes} & Select &
    \colordiffcell{0.823}{0.793} & \colordiffcell{0.729}{0.716} & \colordiffcell{0.740}{0.746} & \colordiffcell{0.798}{0.835} &
    \colordiffcell{0.831}{0.793} & \colordiffcell{0.738}{0.789} & \colordiffcell{0.715}{0.740} & \colordiffcell{0.863}{0.869} \\
    \hline
    \empcell \textbf{Yes} & \empcell \textbf{Yes} & No & No & Conclude &
    \colordiffcell{0.762}{0.793} & \colordiffcell{0.727}{0.716} & \colordiffcell{0.750}{0.746} & \colordiffcell{0.825}{0.835} &
    \colordiffcell{0.786}{0.793} & \colordiffcell{0.743}{0.789} & \colordiffcell{0.713}{0.740} & \colordiffcell{0.868}{0.869} \\
    \empcell \textbf{Yes} & No & \empcell \textbf{Yes} & No & Conclude &
    \colordiffcell{0.775}{0.793} & \colordiffcell{0.692}{0.716} & \colordiffcell{0.737}{0.746} & \colordiffcell{0.823}{0.835} &
    \colordiffcell{0.825}{0.793} & \colordiffcell{0.761}{0.789} & \colordiffcell{0.779}{0.740} & \colordiffcell{0.877}{0.869} \\
    \empcell \textbf{Yes} & No & No & \empcell \textbf{Yes} & Conclude &
    \colordiffcell{0.808}{0.793} & \colordiffcell{0.721}{0.716} & \colordiffcell{0.698}{0.746} & \colordiffcell{0.819}{0.835} &
    \colordiffcell{0.831}{0.793} & \colordiffcell{0.720}{0.789} & \colordiffcell{0.722}{0.740} & \colordiffcell{0.845}{0.869} \\
    \empcell \textbf{Yes} & No & \empcell \textbf{Yes} & \empcell \textbf{Yes} & Conclude &
    \colordiffcell{0.846}{0.793} & \colordiffcell{0.748}{0.716} & \colordiffcell{0.758}{0.746} & \colordiffcell{0.852}{0.835} &
    \colordiffcell{0.859}{0.793} & \colordiffcell{0.752}{0.789} & \colordiffcell{0.749}{0.740} & \colordiffcell{0.859}{0.869} \\
    \empcell \textbf{Yes} & \empcell \textbf{Yes} & \empcell \textbf{Yes} & \empcell \textbf{Yes} & Conclude &
    \colordiffcell{0.819}{0.793} & \colordiffcell{0.762}{0.716} & \colordiffcell{0.777}{0.746} & \colordiffcell{0.837}{0.835} &
    \colordiffcell{0.854}{0.793} & \colordiffcell{0.768}{0.789} & \colordiffcell{0.752}{0.740} & \colordiffcell{0.866}{0.869}
\\
    \hline
    \multicolumn{5}{c|}{\textbf{Meta Judge: Llama3.3 70B}} &  &   &   &   &   &   &   &  \\
    R1-Distilled-Qwen 32B & 4o-mini & Llama3.3 70B & DeepSeek-V3  & $Mode$ & $Pos.$ & $Verb.$ & $Band.$ & $Cot.$ & $Pos.$ & $Verb.$ & $Band.$ & $Cot.$ \\
    \hline
      No & No &  No  & No & Judge &
    0.814 & 0.720 & 0.783 & 0.821 & 0.823 & 0.773 & 0.768 & 0.839 \\
    \empcell \textbf{Yes} & \empcell \textbf{Yes} &  No  & No & Select &
    \colordiffcell{0.767}{0.814} & \colordiffcell{0.712}{0.720} & \colordiffcell{0.733}{0.783} & \colordiffcell{0.812}{0.821} &
    \colordiffcell{0.804}{0.823} & \colordiffcell{0.754}{0.773} & \colordiffcell{0.720}{0.768} & \colordiffcell{0.841}{0.839} \\
    \empcell \textbf{Yes} & No & \empcell \textbf{Yes} & No & Select &
    \colordiffcell{0.787}{0.814} & \colordiffcell{0.702}{0.720} & \colordiffcell{0.744}{0.783} & \colordiffcell{0.785}{0.821} &
    \colordiffcell{0.804}{0.823} & \colordiffcell{0.727}{0.773} & \colordiffcell{0.740}{0.768} & \colordiffcell{0.836}{0.839} \\
    \empcell \textbf{Yes} & No & No & \empcell \textbf{Yes} & Select &
    \colordiffcell{0.800}{0.814} & \colordiffcell{0.719}{0.720} & \colordiffcell{0.729}{0.783} & \colordiffcell{0.815}{0.821} &
    \colordiffcell{0.809}{0.823} & \colordiffcell{0.704}{0.773} & \colordiffcell{0.699}{0.768} & \colordiffcell{0.838}{0.839} \\
    \empcell \textbf{Yes} & \empcell \textbf{Yes} & No & \empcell \textbf{Yes} & Select &
    \colordiffcell{0.831}{0.814} & \colordiffcell{0.735}{0.720} & \colordiffcell{0.717}{0.783} & \colordiffcell{0.815}{0.821} &
    \colordiffcell{0.800}{0.823} & \colordiffcell{0.738}{0.773} & \colordiffcell{0.722}{0.768} & \colordiffcell{0.863}{0.839} \\
    \empcell \textbf{Yes} & \empcell \textbf{Yes} & \empcell \textbf{Yes} & \empcell \textbf{Yes} & Select &
    \colordiffcell{0.812}{0.814} & \colordiffcell{0.721}{0.720} & \colordiffcell{0.758}{0.783} & \colordiffcell{0.796}{0.821} &
    \colordiffcell{0.806}{0.823} & \colordiffcell{0.738}{0.773} & \colordiffcell{0.754}{0.768} & \colordiffcell{0.882}{0.839} \\
    \hline
    \empcell \textbf{Yes} & \empcell \textbf{Yes} & No & No & Conclude &
    \colordiffcell{0.754}{0.814} & \colordiffcell{0.690}{0.720} & \colordiffcell{0.715}{0.783} & \colordiffcell{0.790}{0.821} &
    \colordiffcell{0.772}{0.823} & \colordiffcell{0.747}{0.773} & \colordiffcell{0.733}{0.768} & \colordiffcell{0.793}{0.839} \\
    \empcell \textbf{Yes} & No & \empcell \textbf{Yes} & No & Conclude &
    \colordiffcell{0.806}{0.814} & \colordiffcell{0.717}{0.720} & \colordiffcell{0.746}{0.783} & \colordiffcell{0.792}{0.821} &
    \colordiffcell{0.786}{0.823} & \colordiffcell{0.733}{0.773} & \colordiffcell{0.738}{0.768} & \colordiffcell{0.818}{0.839} \\
    \empcell \textbf{Yes} & No & No & \empcell \textbf{Yes} & Conclude &
    \colordiffcell{0.806}{0.814} & \colordiffcell{0.706}{0.720} & \colordiffcell{0.687}{0.783} & \colordiffcell{0.800}{0.821} &
    \colordiffcell{0.788}{0.823} & \colordiffcell{0.688}{0.773} & \colordiffcell{0.697}{0.768} & \colordiffcell{0.809}{0.839} \\
    \empcell \textbf{Yes} & \empcell \textbf{Yes} & No & \empcell \textbf{Yes} & Conclude &
    \colordiffcell{0.804}{0.814} & \colordiffcell{0.723}{0.720} & \colordiffcell{0.725}{0.783} & \colordiffcell{0.787}{0.821} &
    \colordiffcell{0.804}{0.823} & \colordiffcell{0.720}{0.773} & \colordiffcell{0.686}{0.768} & \colordiffcell{0.838}{0.839} \\
    \empcell \textbf{Yes} & \empcell \textbf{Yes} & \empcell \textbf{Yes} & \empcell \textbf{Yes} & Conclude &
    \colordiffcell{0.833}{0.814} & \colordiffcell{0.712}{0.720} & \colordiffcell{0.731}{0.783} & \colordiffcell{0.810}{0.821} &
    \colordiffcell{0.822}{0.823} & \colordiffcell{0.733}{0.773} & \colordiffcell{0.722}{0.768} & \colordiffcell{0.806}{0.839}
\\
    \hline
    \multicolumn{5}{c|}{\textbf{Meta Judge: DeepSeek-V3 }} &  &   &   &   &   &   &   &  \\
    R1-Distilled-Qwen 32B & 4o-mini & Llama3.3 70B & DeepSeek-V3  & $Mode$ & $Pos.$ & $Verb.$ & $Band.$ & $Cot.$ & $Pos.$ & $Verb.$ & $Band.$ & $Cot.$ \\
    \hline
      No & No &  No  & No & Judge &
    0.821 & 0.746 & 0.748 & 0.883 & 0.809 & 0.718 & 0.688 & 0.868 \\
    \empcell \textbf{Yes} & \empcell \textbf{Yes} &  No  & No & Select &
    \colordiffcell{0.773}{0.821} & \colordiffcell{0.708}{0.746} & \colordiffcell{0.687}{0.748} & \colordiffcell{0.817}{0.883} &
    \colordiffcell{0.788}{0.809} & \colordiffcell{0.745}{0.718} & \colordiffcell{0.686}{0.688} & \colordiffcell{0.852}{0.868} \\
    \empcell \textbf{Yes} & No & \empcell \textbf{Yes} & No & Select &
    \colordiffcell{0.775}{0.821} & \colordiffcell{0.721}{0.746} & \colordiffcell{0.737}{0.748} & \colordiffcell{0.796}{0.883} &
    \colordiffcell{0.797}{0.809} & \colordiffcell{0.736}{0.718} & \colordiffcell{0.695}{0.688} & \colordiffcell{0.836}{0.868} \\
    \empcell \textbf{Yes} & No & No & \empcell \textbf{Yes} & Select &
    \colordiffcell{0.802}{0.821} & \colordiffcell{0.721}{0.746} & \colordiffcell{0.725}{0.748} & \colordiffcell{0.827}{0.883} &
    \colordiffcell{0.813}{0.809} & \colordiffcell{0.729}{0.718} & \colordiffcell{0.663}{0.688} & \colordiffcell{0.854}{0.868} \\
    \empcell \textbf{Yes} & \empcell \textbf{Yes} & \empcell \textbf{Yes} & No & Select &
    \colordiffcell{0.815}{0.821} & \colordiffcell{0.719}{0.746} & \colordiffcell{0.725}{0.748} & \colordiffcell{0.821}{0.883} &
    \colordiffcell{0.831}{0.809} & \colordiffcell{0.759}{0.718} & \colordiffcell{0.672}{0.688} & \colordiffcell{0.856}{0.868} \\
    \empcell \textbf{Yes} & \empcell \textbf{Yes} & \empcell \textbf{Yes} & \empcell \textbf{Yes} & Select &
    \colordiffcell{0.840}{0.821} & \colordiffcell{0.715}{0.746} & \colordiffcell{0.735}{0.748} & \colordiffcell{0.825}{0.883} &
    \colordiffcell{0.838}{0.809} & \colordiffcell{0.738}{0.718} & \colordiffcell{0.638}{0.688} & \colordiffcell{0.870}{0.868} \\
    \hline
    \empcell \textbf{Yes} & \empcell \textbf{Yes} & No & No & Conclude &
    \colordiffcell{0.800}{0.821} & \colordiffcell{0.760}{0.746} & \colordiffcell{0.687}{0.748} & \colordiffcell{0.823}{0.883} &
    \colordiffcell{0.795}{0.809} & \colordiffcell{0.779}{0.718} & \colordiffcell{0.674}{0.688} & \colordiffcell{0.854}{0.868} \\
    \empcell \textbf{Yes} & No & \empcell \textbf{Yes} & No & Conclude &
    \colordiffcell{0.817}{0.821} & \colordiffcell{0.731}{0.746} & \colordiffcell{0.754}{0.748} & \colordiffcell{0.823}{0.883} &
    \colordiffcell{0.804}{0.809} & \colordiffcell{0.754}{0.718} & \colordiffcell{0.711}{0.688} & \colordiffcell{0.847}{0.868} \\
    \empcell \textbf{Yes} & No & No & \empcell \textbf{Yes} & Conclude &
    \colordiffcell{0.827}{0.821} & \colordiffcell{0.740}{0.746} & \colordiffcell{0.692}{0.748} & \colordiffcell{0.860}{0.883} &
    \colordiffcell{0.841}{0.809} & \colordiffcell{0.718}{0.718} & \colordiffcell{0.672}{0.688} & \colordiffcell{0.850}{0.868} \\
    \empcell \textbf{Yes} & \empcell \textbf{Yes} & \empcell \textbf{Yes} & No & Conclude &
    \colordiffcell{0.812}{0.821} & \colordiffcell{0.767}{0.746} & \colordiffcell{0.746}{0.748} & \colordiffcell{0.848}{0.883} &
    \colordiffcell{0.815}{0.809} & \colordiffcell{0.754}{0.718} & \colordiffcell{0.695}{0.688} & \colordiffcell{0.852}{0.868} \\
    \empcell \textbf{Yes} & \empcell \textbf{Yes} & \empcell \textbf{Yes} & \empcell \textbf{Yes} & Conclude &
    \colordiffcell{0.825}{0.821} & \colordiffcell{0.765}{0.746} & \colordiffcell{0.733}{0.748} & \colordiffcell{0.873}{0.883} &
    \colordiffcell{0.859}{0.809} & \colordiffcell{0.736}{0.718} & \colordiffcell{0.692}{0.688} & \colordiffcell{0.845}{0.868}
\\
     \bottomrule
    \end{tabular}
    }
    \caption{Consistency scores for LLM-as-Meta-Judge across biases, pool sizes, and modes. "Yes" indicates inclusion of the model in the judgment pool; "No" indicates exclusion (single judge). 
    Cell colors indicate the difference compared with the single judge: \textcolor{brightgreen}{green} denotes higher and \textcolor{brightblue}{blue} denotes lower.
    }
    \label{Meta_result}
\vspace{-3mm}
\end{table*}
\section{Results and Analysis}

\begin{table}[ht]
    \centering
    \setlength{\tabcolsep}{4pt}
    \scalebox{0.71}{
    \begin{tabular}{cc|c|ll}
    \toprule
     Judge & Critic & Round &\textbf{MTBench} & \textbf{CALM} \\    
    \hline
     &  &   1&   0.817
&  0.786
\\
    4o-mini & PINE &  2&  0.825 $\uparrow$& 0.804 $\uparrow$\\
    & &  3&  0.831 $\uparrow$& 0.815 $\uparrow$\\
    \hline
    & &   1&   0.802&  0.781\\
    Llama3.3 70B& PINE &  2&  0.812 $\uparrow$& 0.786 $\uparrow$\\
    & &  3&  0.821 $\uparrow$& 0.788 $\uparrow$\\
    \bottomrule
    
    \end{tabular}
    }
    \caption{Position bias consistency score when bias-free agent are served as a critic. Round 0 indicates the consistency rate of
judges without debate.}
    \label{PINE_debate}
    \vspace{-4mm}
\end{table}

\begin{table*}[ht]
    \centering
    \setlength{\tabcolsep}{15pt}
    \scalebox{0.6}{
    \begin{tabular}{lllll|cc|cc}
    \hline
     \multicolumn{5}{c|}{\textbf{Meta Judge: 4o-mini}} & \multicolumn{2}{c|}{\textbf{MT-Bench}} & \multicolumn{2}{c}{\textbf{CALM-Alignment}}  \\
    \cline{6-7} \cline{8-9}
     4o-mini  & DeepSeek-V3 & Llama3.3 70B & R1-Distilled-Qwen 32B  & PINE & Select & Conclude & Select & Conclude \\
    \hline
     \empcell \textbf{Yes}& No & No  & No & \empcell \textbf{Yes}& 0.833& 0.794& 0.815&0.829
    \\
        & No &  No  & \empcell \textbf{Yes}& \empcell \textbf{Yes}&     0.802&     0.781&     0.800&  0.827
    \\
        No & No &  \empcell \textbf{Yes}& \empcell \textbf{Yes}& \empcell \textbf{Yes}&     0.833&     0.823&     \textcolor{teal}{\textbf{0.843}}&  0.847
    \\
        No & \empcell \textbf{Yes}&  \empcell \textbf{Yes}& \empcell \textbf{Yes}& \empcell \textbf{Yes}&     \textcolor{teal}{\textbf{0.840}}&     \textcolor{teal}{\textbf{0.848}}&     0.831&  0.852
    \\
    \empcell \textbf{Yes}& \empcell \textbf{Yes}&  \empcell \textbf{Yes}& \empcell \textbf{Yes}& \empcell \textbf{Yes}&     0.804&     0.817&     0.841&  \textcolor{teal}{\textbf{0.863}}\\
    \hline
    \multicolumn{5}{c|}{\textbf{Meta Judge: Llama3.3 70B}} &      &      &      &  \\
   4o-mini  & DeepSeek-V3 & Llama3.3 70B & R1-Distilled-Qwen 32B  & PINE & Select & Conclude & Select & Conclude \\
    \hline
     No & No  & \empcell \textbf{Yes}& No & \empcell \textbf{Yes}& 0.815& \textcolor{teal}{\textbf{0.827}}& 0.843&0.784
    \\
        & No &  No  & \empcell \textbf{Yes}& \empcell \textbf{Yes}&     0.775&     0.771&     0.777&  0.800
    \\
        No & No &  \empcell \textbf{Yes}& \empcell \textbf{Yes}& \empcell \textbf{Yes}&     0.790&     0.775&     0.815&  0.806
    \\
        No & \empcell \textbf{Yes}&  \empcell \textbf{Yes}& \empcell \textbf{Yes}& \empcell \textbf{Yes}&     0.806&     0.783&     \textcolor{teal}{\textbf{0.861}}&  \textcolor{teal}{\textbf{0.827}}\\
        \empcell \textbf{Yes}& \empcell \textbf{Yes}&  \empcell \textbf{Yes}& \empcell \textbf{Yes}& \empcell \textbf{Yes}&     \textcolor{teal}{\textbf{0.842}}&     0.808&     0.825&  0.806
    \\
    \hline
        \multicolumn{5}{c|}{\textbf{Meta Judge: DeepSeek-V3}} &      &      &      &  \\
   4o-mini  & DeepSeek-V3 & Llama3.3 70B & R1-Distilled-Qwen 32B  & PINE & Select & Conclude & Select & Conclude \\
    \hline
     No & \empcell \textbf{Yes}& No & No & \empcell \textbf{Yes}& 0.819& 0.792& 0.809&0.790
    \\
        & No &  No  & \empcell \textbf{Yes}& \empcell \textbf{Yes}&     0.767&     0.792&     0.781&  0.795
    \\
        No & No &  \empcell \textbf{Yes}& \empcell \textbf{Yes}& \empcell \textbf{Yes}&     0.802&     0.787&     0.827&  0.784
    \\
        No & \empcell \textbf{Yes}&  \empcell \textbf{Yes}& \empcell \textbf{Yes}& \empcell \textbf{Yes}&     0.817&     0.817&     \textcolor{teal}{\textbf{0.868}}&  0.836
    \\
    \empcell \textbf{Yes}& \empcell \textbf{Yes}&  \empcell \textbf{Yes}& \empcell \textbf{Yes}& \empcell \textbf{Yes}&     \textcolor{teal}{\textbf{0.840}}&     \textcolor{teal}{\textbf{0.827}}&     0.845&  \textcolor{teal}{\textbf{0.850}}\\
    \hline
    \end{tabular}
    }
    \caption{Position bias consistency score when bias-free agent is included in the judgment pool. "Yes" indicates inclusion of the model in the judgment pool; "No" indicates exclusion. The highest consistency is shown in \textcolor{teal}{\textbf{bold}}.} 
    \label{PINE_meta}
\end{table*}

\subsection{Multi-Agent-Debate}


\textbf{Multi-Agent Debate Can Amplify Bias Over Rounds.}
As shown in \autoref{MAD_result}, introducing Multi-Agent Debate into LLM evaluation leads to a sharp increase in bias immediately after the first round. Multiple pairwise t-tests across all judge–critic model pairings, both across and within different bias types, reveal a statistically significant drop in consistency rate from round 0 to round 1. Notably, this elevated level of bias persists in subsequent rounds, with no significant further increase or signs of recovery. This pattern is consistent across all bias types we tested. These findings suggest that the initial introduction of debate—rather than its continuation—poses the greatest risk for bias amplification in multi-agent evaluation. Moreover, additional rounds of debate appear insufficient to recover the amplified bias once introduced. Detailed analyses and statistical results are provided in Appendix~\ref{Stats_MAD}.

\textbf{Bias Amplification Is a General Phenomenon, May Not Tied to Weak Models.} Our results show that the amplification of bias in Multi-Agent-Debate is not limited to weaker or smaller models, but is a phenomenon observed across all judge-critic pairings. For example, DeepSeek-V3, despite being one of the largest and most advanced models in our experiments, exhibited a substantial and statistically significant drop in consistency rate after the introduction of debate, with position consistency falling from 0.82 pre-debate to 0.71 after the first round. Similar patterns were observed for other leading models such as Llama-3.3-70B and GPT4o-mini, indicating that increased model size and capability may not confer immunity to such bias amplification. This generality across architectures and scales underscores that the challenge of collaborative bias amplification is intrinsic to the debate framework itself, rather than a limitation of particular models.


\subsection{LLM-as-Meta-Judge}

\textbf{LLM-as-Meta-Judge Frameworks Achieve Bias Consistency Comparable to Single-Model Judgment.} As shown in \autoref{Meta_result}, across both the MT-Bench and CALM-Alignment benchmarks, and for most bias types, the Select and Conclude modes with diverse judge pools generally yield consistency scores comparable to, and in some cases better than, those of a strong single-model judge. However, comprehensive pairwise t-tests comparing all pooled Meta-Judge settings against their single-judge baselines reveal a statistically significant drop in overall consistency. gowever, the magnitude of this drop is relatively small—typically within 2\%. Additionally, for position consistency, the Meta-Judge framework shows no significant difference from, and sometimes even outperforms, the single-judge model under certain mode and size of judgment pool. Overall, this suggests that the Meta-Judge achieves bias consistency levels comparable to those of the single-judge model. Detailed discussions and results can be found in  Appenidx~ \ref{Stats_Meta}.

\textbf{Larger Judgment Pools Yield Comparable Consistency, with Gains in Position Bias.} For example, using 4o-mini as the Meta-Judge on the CALM-Alignment benchmark, position bias consistency rises from 0.793 to 0.854 in conclude mode. Statistical analysis confirms that this improvement is significant for position bias. Comprehensive pairwise t-tests across different pool sizes suggest that with a pool size of two, the Meta-Judge generally yields significantly lower consistency across all bias types. However, as the pool size increases, consistency becomes comparable to that of single-judge models—and even better in terms of position consistency. These results suggest that increasing pool size is beneficial in Meta-Judge settings. However, there is no evidence that it consistently outperforms the single-judge model overall. Full details of the analysis are provided in Appendix~\ref{Stats_Meta}.

\subsection{Bias-Free Experiment} 

\textbf{Although the bias-free agent performs well in debate settings, it has minimal effect when used as a Meta-Judge.} As shown in the \autoref{PINE_debate}, we observe a continuous improvement in position bias consistency with increasing rounds of debate. This is significantly different from the what we observed without such bias-free agent. On the other hand, as shown in \autoref{PINE_meta}, there is no considerable improvement to the consistency score when including this bias-free agent into the judgment pool. Our analysis further shows that, in select mode, the Meta-Judge shows no favor to the bias-free agent. This might suggest that the Meta-Judge itself may not be sensitive enough to recognize or value the unbiased nature of the agent’s outputs, instead treating all pool members as equally valid. As a result, the bias-free agent’s potential for debiasing is diluted within the collective judgment process. More details and analysis can be found in Appendix~\ref{Stats_pine}. More results of PINE with Llama3 as base model can be found in \autoref{More_PINE}.
\section{Conclusion}

In conclusion, our experiments show that Multi-Agent Debate can amplify intrinsic biases in LLMs through agent interactions, while the LLM-as-Meta-Judge approach demonstrates better consistency, and sometimes improvement, against these biases, especially with larger judgment pools. Notably, the presence of a bias-free agent makes the Multi-Agent Debate framework even more promising, at times surpassing the Meta-Judge approach. However, these results highlight the importance of developing bias-free agents that can effectively debias within fewer rounds of interaction. Overall, our findings offer guidance for deploying Multi-Agent LLM-as-Judge frameworks and emphasize the need for universal debiasing techniques capable of addressing multiple biases simultaneously, particularly in collaborative evaluation settings.
\section*{Limitations}

This study focuses on understanding how biases manifest in multi-agent frameworks when used in the context of LLM-as-a-Judge systems. We specifically examine four types of bias: position bias, bandwagon bias, chain-of-thought bias, and verbosity bias. While these do not represent the full spectrum of potential biases, they are representative and have received substantial attention in prior work. Our goal is not to provide a comprehensive taxonomy of biases, but rather to initiate and encourage deeper exploration into how bias operates in collaborative LLM settings. Investigating additional forms of bias remains an important direction for future research.

To maintain clarity and control over confounding variables, we adopt a deliberately simplified multi-agent framework. Rather than replicating all existing frameworks, we focus on a core structure that is generalizable to more complex designs. This choice allows us to isolate bias-related effects without interference from more elaborate coordination strategies, but it also limits the direct applicability of our results to more sophisticated multi-agent architectures. Future work could investigate how different agent designs and interaction protocols influence the emergence or mitigation of bias.

In terms of mitigation, our study evaluates only a single method for reducing position bias. Although more advanced strategies—such as modifying judge prompt structures—may yield stronger results, our focus is on scenarios where altering individual judge models is not feasible. We believe this reflects a more realistic and widely applicable constraint in real-world deployments. Our findings suggest that even under such constraints, meaningful bias mitigation remains a critical challenge, and we emphasize the need for more general and robust solutions that can be applied across multiple bias types in collaborative LLM systems.

Finally, our experiments are conducted on two benchmarks. While other benchmarks exist—such as Judge-Bench \cite{tan2024judgebench}, which emphasizes problem-solving ability—we chose not to include them due to the risk of confounding bias measurements with task-solving failures. Accurately attributing selection outcomes to bias becomes difficult when a model’s ability to answer the question is in question. Moreover, multi-agent inference is computationally intensive and demands strong instruction-following capabilities, which limits the scale of evaluation. We test four different model types across a diverse set of permutations, resulting in approximately 288 total instances. While this setup is sufficient for our core analyses, it presents challenges for fine-grained statistical testing, where smaller subgroup sizes can lead to greater variance and less reliable conclusions.



\section*{Ethical Statement}

This work involves the use of a publicly available dataset, CALM alignment \cite{ye2024justice}, which contains a small portion of content labeled as NSFW. According to the dataset authors, all data has been manually reviewed and curated to be appropriate for research use. In accordance with the dataset's license and usage guidelines, we have not redistributed or publicly shared any part of the dataset. These data are used solely for the purpose of evaluating bias in AI models. We encourage responsible use of this work, with attention to ethical standards and potential social implications.
\section*{Acknowledgment}
This research was supported in part by a generous grant from the John Templeton Foundation.

\bibliography{custom}

\begin{thebibliography}{57}
\providecommand{\natexlab}[1]{#1}

\bibitem[{Achiam et~al.(2023)Achiam, Adler, Agarwal, Ahmad, Akkaya, Aleman, Almeida, Altenschmidt, Altman, Anadkat et~al.}]{achiam2023gpt}
Josh Achiam, Steven Adler, Sandhini Agarwal, Lama Ahmad, Ilge Akkaya, Florencia~Leoni Aleman, Diogo Almeida, Janko Altenschmidt, Sam Altman, Shyamal Anadkat, and 1 others. 2023.
\newblock Gpt-4 technical report.
\newblock \emph{arXiv preprint arXiv:2303.08774}.

\bibitem[{Arif et~al.(2024)Arif, Farid, Azeemi, Athar, and Raza}]{arif2024fellowship}
Samee Arif, Sualeha Farid, Abdul~Hameed Azeemi, Awais Athar, and Agha~Ali Raza. 2024.
\newblock The fellowship of the llms: Multi-agent workflows for synthetic preference optimization dataset generation.
\newblock \emph{arXiv preprint arXiv:2408.08688}.

\bibitem[{Bai et~al.(2023)Bai, Bai, Chu, Cui, Dang, Deng, Fan, Ge, Han, Huang, Hui, Ji, Li, Lin, Lin, Liu, Liu, Lu, Lu, Ma, Men, Ren, Ren, Tan, Tan, Tu, Wang, Wang, Wang, Wu, Xu, Xu, Yang, Yang, Yang, Yang, Yao, Yu, Yuan, Yuan, Zhang, Zhang, Zhang, Zhang, Zhou, Zhou, Zhou, and Zhu}]{qwen}
Jinze Bai, Shuai Bai, Yunfei Chu, Zeyu Cui, Kai Dang, Xiaodong Deng, Yang Fan, Wenbin Ge, Yu~Han, Fei Huang, Binyuan Hui, Luo Ji, Mei Li, Junyang Lin, Runji Lin, Dayiheng Liu, Gao Liu, Chengqiang Lu, Keming Lu, and 29 others. 2023.
\newblock Qwen technical report.
\newblock \emph{arXiv preprint arXiv:2309.16609}.

\bibitem[{Bao et~al.(2024)Bao, Huang, Wang, Ye, Wang, Chen, Zhao, Zhou, Elhoseiny, and Zhang}]{bao2024autobench}
Han Bao, Yue Huang, Yanbo Wang, Jiayi Ye, Xiangqi Wang, Xiuying Chen, Yue Zhao, Tianyi Zhou, Mohamed Elhoseiny, and Xiangliang Zhang. 2024.
\newblock Autobench-v: Can large vision-language models benchmark themselves?
\newblock \emph{arXiv preprint arXiv:2410.21259}.

\bibitem[{Burns et~al.(2023)Burns, Izmailov, Kirchner, Baker, Gao, Aschenbrenner, Chen, Ecoffet, Joglekar, Leike et~al.}]{burns2023weak}
Collin Burns, Pavel Izmailov, Jan~Hendrik Kirchner, Bowen Baker, Leo Gao, Leopold Aschenbrenner, Yining Chen, Adrien Ecoffet, Manas Joglekar, Jan Leike, and 1 others. 2023.
\newblock Weak-to-strong generalization: Eliciting strong capabilities with weak supervision.
\newblock \emph{arXiv preprint arXiv:2312.09390}.

\bibitem[{Chan et~al.(2023)Chan, Chen, Su, Yu, Xue, Zhang, Fu, and Liu}]{chan2023chateval}
Chi-Min Chan, Weize Chen, Yusheng Su, Jianxuan Yu, Wei Xue, Shanghang Zhang, Jie Fu, and Zhiyuan Liu. 2023.
\newblock Chateval: Towards better llm-based evaluators through multi-agent debate.
\newblock \emph{arXiv preprint arXiv:2308.07201}.

\bibitem[{Chen et~al.(2024)Chen, Chen, Zhang, Wang, Liu, Zhou, Zhang, Wan, Zhou, and Sun}]{chen2024mllm}
Dongping Chen, Ruoxi Chen, Shilin Zhang, Yaochen Wang, Yinuo Liu, Huichi Zhou, Qihui Zhang, Yao Wan, Pan Zhou, and Lichao Sun. 2024.
\newblock Mllm-as-a-judge: Assessing multimodal llm-as-a-judge with vision-language benchmark.
\newblock In \emph{Forty-first International Conference on Machine Learning}.

\bibitem[{Chen et~al.(2023)Chen, Su, Zuo, Yang, Yuan, Qian, Chan, Qin, Lu, Xie et~al.}]{chen2023agentverse}
Weize Chen, Yusheng Su, Jingwei Zuo, Cheng Yang, Chenfei Yuan, Chen Qian, Chi-Min Chan, Yujia Qin, Yaxi Lu, Ruobing Xie, and 1 others. 2023.
\newblock Agentverse: Facilitating multi-agent collaboration and exploring emergent behaviors in agents.
\newblock \emph{arXiv preprint arXiv:2308.10848}, 2(4):6.

\bibitem[{DeepSeek-AI(2025)}]{deepseekai2025deepseekr1incentivizingreasoningcapability}
DeepSeek-AI. 2025.
\newblock \href {https://arxiv.org/abs/2501.12948} {Deepseek-r1: Incentivizing reasoning capability in llms via reinforcement learning}.
\newblock \emph{Preprint}, arXiv:2501.12948.

\bibitem[{Du et~al.(2023)Du, Li, Torralba, Tenenbaum, and Mordatch}]{du2023improving}
Yilun Du, Shuang Li, Antonio Torralba, Joshua~B Tenenbaum, and Igor Mordatch. 2023.
\newblock Improving factuality and reasoning in language models through multiagent debate.
\newblock In \emph{Forty-first International Conference on Machine Learning}.

\bibitem[{Dubois et~al.(2023)Dubois, Li, Taori, Zhang, Gulrajani, Ba, Guestrin, Liang, and Hashimoto}]{dubois2023alpacafarm}
Yann Dubois, Chen~Xuechen Li, Rohan Taori, Tianyi Zhang, Ishaan Gulrajani, Jimmy Ba, Carlos Guestrin, Percy~S Liang, and Tatsunori~B Hashimoto. 2023.
\newblock Alpacafarm: A simulation framework for methods that learn from human feedback.
\newblock \emph{Advances in Neural Information Processing Systems}, 36:30039--30069.

\bibitem[{Feng et~al.(2024)Feng, Su, Zheng, Ren, Zhang, Wu, Wang, and Liu}]{feng2024m}
Zhaopeng Feng, Jiayuan Su, Jiamei Zheng, Jiahan Ren, Yan Zhang, Jian Wu, Hongwei Wang, and Zuozhu Liu. 2024.
\newblock M-mad: Multidimensional multi-agent debate framework for fine-grained machine translation evaluation.
\newblock \emph{arXiv preprint arXiv:2412.20127}.

\bibitem[{Gao et~al.(2024)Gao, Xie, Mao, Wu, Xia, Mi, and Wei}]{gao2024meta}
Peizhong Gao, Ao~Xie, Shaoguang Mao, Wenshan Wu, Yan Xia, Haipeng Mi, and Furu Wei. 2024.
\newblock Meta reasoning for large language models.
\newblock \emph{arXiv preprint arXiv:2406.11698}.

\bibitem[{Grattafiori et~al.(2024)Grattafiori, Dubey, Jauhri, Pandey, Kadian, Al-Dahle, Letman, Mathur, Schelten, Vaughan et~al.}]{grattafiori2024llama}
Aaron Grattafiori, Abhimanyu Dubey, Abhinav Jauhri, Abhinav Pandey, Abhishek Kadian, Ahmad Al-Dahle, Aiesha Letman, Akhil Mathur, Alan Schelten, Alex Vaughan, and 1 others. 2024.
\newblock The llama 3 herd of models.
\newblock \emph{arXiv preprint arXiv:2407.21783}.

\bibitem[{Gu et~al.(2024)Gu, Jiang, Shi, Tan, Zhai, Xu, Li, Shen, Ma, Liu et~al.}]{gu2024survey}
Jiawei Gu, Xuhui Jiang, Zhichao Shi, Hexiang Tan, Xuehao Zhai, Chengjin Xu, Wei Li, Yinghan Shen, Shengjie Ma, Honghao Liu, and 1 others. 2024.
\newblock A survey on llm-as-a-judge.
\newblock \emph{arXiv preprint arXiv:2411.15594}.

\bibitem[{Gulati et~al.(2025)Gulati, D'Inc{\`a}, Sebe, Lepri, and Oliver}]{gulati2025uncovering}
Aditya Gulati, Moreno D'Inc{\`a}, Nicu Sebe, Bruno Lepri, and Nuria Oliver. 2025.
\newblock Uncovering an attractiveness bias in multimodal large language models: A case study with llava.
\newblock \emph{arXiv preprint arXiv:2504.16104}.

\bibitem[{Hoffmann et~al.(2022)Hoffmann, Borgeaud, Mensch, Buchatskaya, Cai, Rutherford, Casas, Hendricks, Welbl, Clark et~al.}]{hoffmann2022training}
Jordan Hoffmann, Sebastian Borgeaud, Arthur Mensch, Elena Buchatskaya, Trevor Cai, Eliza Rutherford, Diego de~Las Casas, Lisa~Anne Hendricks, Johannes Welbl, Aidan Clark, and 1 others. 2022.
\newblock Training compute-optimal large language models.
\newblock \emph{arXiv preprint arXiv:2203.15556}.

\bibitem[{Hong et~al.(2023)Hong, Zheng, Chen, Cheng, Wang, Zhang, Wang, Yau, Lin, Zhou et~al.}]{hong2023metagpt}
Sirui Hong, Xiawu Zheng, Jonathan Chen, Yuheng Cheng, Jinlin Wang, Ceyao Zhang, Zili Wang, Steven Ka~Shing Yau, Zijuan Lin, Liyang Zhou, and 1 others. 2023.
\newblock Metagpt: Meta programming for multi-agent collaborative framework.
\newblock \emph{arXiv preprint arXiv:2308.00352}, 3(4):6.

\bibitem[{Huang et~al.(2023{\natexlab{a}})Huang, Zhang, Luck, Bu, Qing, and Cui}]{huang2023agentcoder}
Dong Huang, Jie~M Zhang, Michael Luck, Qingwen Bu, Yuhao Qing, and Heming Cui. 2023{\natexlab{a}}.
\newblock Agentcoder: Multi-agent-based code generation with iterative testing and optimisation.
\newblock \emph{arXiv preprint arXiv:2312.13010}.

\bibitem[{Huang et~al.(2023{\natexlab{b}})Huang, Chen, Mishra, Zheng, Yu, Song, and Zhou}]{huang2023large}
Jie Huang, Xinyun Chen, Swaroop Mishra, Huaixiu~Steven Zheng, Adams~Wei Yu, Xinying Song, and Denny Zhou. 2023{\natexlab{b}}.
\newblock Large language models cannot self-correct reasoning yet.
\newblock \emph{arXiv preprint arXiv:2310.01798}.

\bibitem[{Kaplan et~al.(2020)Kaplan, McCandlish, Henighan, Brown, Chess, Child, Gray, Radford, Wu, and Amodei}]{kaplan2020scaling}
Jared Kaplan, Sam McCandlish, Tom Henighan, Tom~B Brown, Benjamin Chess, Rewon Child, Scott Gray, Alec Radford, Jeffrey Wu, and Dario Amodei. 2020.
\newblock Scaling laws for neural language models.
\newblock \emph{arXiv preprint arXiv:2001.08361}.

\bibitem[{Lambert et~al.(2024)Lambert, Pyatkin, Morrison, Miranda, Lin, Chandu, Dziri, Kumar, Zick, Choi et~al.}]{lambert2024rewardbench}
Nathan Lambert, Valentina Pyatkin, Jacob Morrison, LJ~Miranda, Bill~Yuchen Lin, Khyathi Chandu, Nouha Dziri, Sachin Kumar, Tom Zick, Yejin Choi, and 1 others. 2024.
\newblock Rewardbench: Evaluating reward models for language modeling.
\newblock \emph{arXiv preprint arXiv:2403.13787}.

\bibitem[{Li et~al.(2025{\natexlab{a}})Li, Sun, Huang, Zhong, Jiang, Han, Zhang, Wang, and Liu}]{li2025preference}
Dawei Li, Renliang Sun, Yue Huang, Ming Zhong, Bohan Jiang, Jiawei Han, Xiangliang Zhang, Wei Wang, and Huan Liu. 2025{\natexlab{a}}.
\newblock Preference leakage: A contamination problem in llm-as-a-judge.
\newblock \emph{arXiv preprint arXiv:2502.01534}.

\bibitem[{Li et~al.(2023{\natexlab{a}})Li, Patel, and Du}]{li2023prd}
Ruosen Li, Teerth Patel, and Xinya Du. 2023{\natexlab{a}}.
\newblock Prd: Peer rank and discussion improve large language model based evaluations.
\newblock \emph{arXiv preprint arXiv:2307.02762}.

\bibitem[{Li et~al.(2025{\natexlab{b}})Li, Mohamud, Sun, Wu, and Boulet}]{li2025leveraging}
Yuran Li, Jama~Hussein Mohamud, Chongren Sun, Di~Wu, and Benoit Boulet. 2025{\natexlab{b}}.
\newblock Leveraging llms as meta-judges: A multi-agent framework for evaluating llm judgments.
\newblock \emph{arXiv preprint arXiv:2504.17087}.

\bibitem[{Li et~al.(2023{\natexlab{b}})Li, Wang, Ma, Wu, Wang, Gao, and Liu}]{li2023split}
Zongjie Li, Chaozheng Wang, Pingchuan Ma, Daoyuan Wu, Shuai Wang, Cuiyun Gao, and Yang Liu. 2023{\natexlab{b}}.
\newblock Split and merge: Aligning position biases in large language model based evaluators.
\newblock \emph{arXiv preprint arXiv:2310.01432}.

\bibitem[{Liang et~al.(2023)Liang, He, Jiao, Wang, Wang, Wang, Yang, Shi, and Tu}]{liang2023encouraging}
Tian Liang, Zhiwei He, Wenxiang Jiao, Xing Wang, Yan Wang, Rui Wang, Yujiu Yang, Shuming Shi, and Zhaopeng Tu. 2023.
\newblock Encouraging divergent thinking in large language models through multi-agent debate.
\newblock \emph{arXiv preprint arXiv:2305.19118}.

\bibitem[{Liu et~al.(2024{\natexlab{a}})Liu, Feng, Xue, Wang, Wu, Lu, Zhao, Deng, Zhang, Ruan et~al.}]{liu2024deepseek}
Aixin Liu, Bei Feng, Bing Xue, Bingxuan Wang, Bochao Wu, Chengda Lu, Chenggang Zhao, Chengqi Deng, Chenyu Zhang, Chong Ruan, and 1 others. 2024{\natexlab{a}}.
\newblock Deepseek-v3 technical report.
\newblock \emph{arXiv preprint arXiv:2412.19437}.

\bibitem[{Liu et~al.(2024{\natexlab{b}})Liu, Zeng, Liu, Yan, He, Wang, Yan, Liu, and Zhou}]{liu2024skywork}
Chris~Yuhao Liu, Liang Zeng, Jiacai Liu, Rui Yan, Jujie He, Chaojie Wang, Shuicheng Yan, Yang Liu, and Yahui Zhou. 2024{\natexlab{b}}.
\newblock Skywork-reward: Bag of tricks for reward modeling in llms.
\newblock \emph{arXiv preprint arXiv:2410.18451}.

\bibitem[{Liu et~al.(2024{\natexlab{c}})Liu, Wang, Huang, Xu, Zeng, Jiang, Yang, and Li}]{liu2024groupdebate}
Tongxuan Liu, Xingyu Wang, Weizhe Huang, Wenjiang Xu, Yuting Zeng, Lei Jiang, Hailong Yang, and Jing Li. 2024{\natexlab{c}}.
\newblock Groupdebate: Enhancing the efficiency of multi-agent debate using group discussion.
\newblock \emph{arXiv preprint arXiv:2409.14051}.

\bibitem[{Park et~al.(2024)Park, Jwa, Meiying, Kim, and Choi}]{park2024offsetbias}
Junsoo Park, Seungyeon Jwa, Ren Meiying, Daeyoung Kim, and Sanghyuk Choi. 2024.
\newblock Offsetbias: Leveraging debiased data for tuning evaluators.
\newblock In \emph{Findings of the Association for Computational Linguistics: EMNLP 2024}, pages 1043--1067.

\bibitem[{Saha et~al.(2025)Saha, Li, Ghazvininejad, Weston, and Wang}]{saha2025learning}
Swarnadeep Saha, Xian Li, Marjan Ghazvininejad, Jason Weston, and Tianlu Wang. 2025.
\newblock Learning to plan \& reason for evaluation with thinking-llm-as-a-judge.
\newblock \emph{arXiv preprint arXiv:2501.18099}.

\bibitem[{Shen et~al.(2025)Shen, Zhu, Zhao, Wu, and Wu}]{shen2025will}
Tao Shen, Didi Zhu, Ziyu Zhao, Chao Wu, and Fei Wu. 2025.
\newblock Will llms scaling hit the wall? breaking barriers via distributed resources on massive edge devices.
\newblock \emph{arXiv preprint arXiv:2503.08223}.

\bibitem[{Shi et~al.(2024)Shi, Ma, Liang, Ma, and Vosoughi}]{shi2024judging}
Lin Shi, Chiyu Ma, Wenhua Liang, Weicheng Ma, and Soroush Vosoughi. 2024.
\newblock Judging the judges: A systematic investigation of position bias in pairwise comparative assessments by llms.
\newblock \emph{arXiv preprint arXiv:2406.07791}.

\bibitem[{Shinn et~al.(2023)Shinn, Cassano, Gopinath, Narasimhan, and Yao}]{shinn2023reflexion}
Noah Shinn, Federico Cassano, Ashwin Gopinath, Karthik Narasimhan, and Shunyu Yao. 2023.
\newblock Reflexion: Language agents with verbal reinforcement learning.
\newblock \emph{Advances in Neural Information Processing Systems}, 36:8634--8652.

\bibitem[{Stahl et~al.(2024)Stahl, Biermann, Nehring, and Wachsmuth}]{stahl2024exploring}
Maja Stahl, Leon Biermann, Andreas Nehring, and Henning Wachsmuth. 2024.
\newblock Exploring llm prompting strategies for joint essay scoring and feedback generation.
\newblock \emph{arXiv preprint arXiv:2404.15845}.

\bibitem[{Sui et~al.(2025)Sui, He, Cao, Han, and Hooi}]{sui2025meta}
Yuan Sui, Yufei He, Tri Cao, Simeng Han, and Bryan Hooi. 2025.
\newblock Meta-reasoner: Dynamic guidance for optimized inference-time reasoning in large language models.
\newblock \emph{arXiv preprint arXiv:2502.19918}.

\bibitem[{Tan et~al.(2024)Tan, Zhuang, Montgomery, Tang, Cuadron, Wang, Popa, and Stoica}]{tan2024judgebench}
Sijun Tan, Siyuan Zhuang, Kyle Montgomery, William~Y Tang, Alejandro Cuadron, Chenguang Wang, Raluca~Ada Popa, and Ion Stoica. 2024.
\newblock Judgebench: A benchmark for evaluating llm-based judges.
\newblock \emph{arXiv preprint arXiv:2410.12784}.

\bibitem[{Tian et~al.(2025)Tian, Zou, Yang, and Zhang}]{tian2025identifying}
Xinyu Tian, Shu Zou, Zhaoyuan Yang, and Jing Zhang. 2025.
\newblock Identifying and mitigating position bias of multi-image vision-language models.
\newblock \emph{arXiv preprint arXiv:2503.13792}.

\bibitem[{Trivedi et~al.(2024)Trivedi, Gulati, Molenschot, Rajeev, Ramamurthy, Stevens, Chaudhery, Jambholkar, Zou, and Rajani}]{trivedi2024self}
Prapti Trivedi, Aditya Gulati, Oliver Molenschot, Meghana~Arakkal Rajeev, Rajkumar Ramamurthy, Keith Stevens, Tanveesh~Singh Chaudhery, Jahnavi Jambholkar, James Zou, and Nazneen Rajani. 2024.
\newblock Self-rationalization improves llm as a fine-grained judge.
\newblock \emph{arXiv preprint arXiv:2410.05495}.

\bibitem[{Verga et~al.(2024)Verga, Hofstatter, Althammer, Su, Piktus, Arkhangorodsky, Xu, White, and Lewis}]{verga2024replacing}
Pat Verga, Sebastian Hofstatter, Sophia Althammer, Yixuan Su, Aleksandra Piktus, Arkady Arkhangorodsky, Minjie Xu, Naomi White, and Patrick Lewis. 2024.
\newblock Replacing judges with juries: Evaluating llm generations with a panel of diverse models.
\newblock \emph{arXiv preprint arXiv:2404.18796}.

\bibitem[{Villalobos et~al.(2024)Villalobos, Ho, Sevilla, Besiroglu, Heim, and Hobbhahn}]{villalobos2024position}
Pablo Villalobos, Anson Ho, Jaime Sevilla, Tamay Besiroglu, Lennart Heim, and Marius Hobbhahn. 2024.
\newblock Position: Will we run out of data? limits of llm scaling based on human-generated data.
\newblock In \emph{Forty-first International Conference on Machine Learning}.

\bibitem[{Wan et~al.(2024)Wan, Vig, Bansal, and Joty}]{wan2024positional}
David Wan, Jesse Vig, Mohit Bansal, and Shafiq Joty. 2024.
\newblock On positional bias of faithfulness for long-form summarization.
\newblock \emph{arXiv preprint arXiv:2410.23609}.

\bibitem[{Wan et~al.(2025)Wan, Li, Song, Wang, Yang, Schmidt, Wang, Zhang, Hu, and Wen}]{wan2025rema}
Ziyu Wan, Yunxiang Li, Yan Song, Hanjing Wang, Linyi Yang, Mark Schmidt, Jun Wang, Weinan Zhang, Shuyue Hu, and Ying Wen. 2025.
\newblock Rema: Learning to meta-think for llms with multi-agent reinforcement learning.
\newblock \emph{arXiv preprint arXiv:2503.09501}.

\bibitem[{Wang et~al.(2024{\natexlab{a}})Wang, Li, Chen, Cai, Zhu, Lin, Cao, Kong, Liu, Liu et~al.}]{wang2024large}
Peiyi Wang, Lei Li, Liang Chen, Zefan Cai, Dawei Zhu, Binghuai Lin, Yunbo Cao, Lingpeng Kong, Qi~Liu, Tianyu Liu, and 1 others. 2024{\natexlab{a}}.
\newblock Large language models are not fair evaluators.
\newblock In \emph{Proceedings of the 62nd Annual Meeting of the Association for Computational Linguistics (Volume 1: Long Papers)}, pages 9440--9450.

\bibitem[{Wang et~al.(2025)Wang, Ji, Yang, Li, Hu, Li, and Sartoretti}]{wang2025mcts}
Yutong Wang, Pengliang Ji, Chaoqun Yang, Kaixin Li, Ming Hu, Jiaoyang Li, and Guillaume Sartoretti. 2025.
\newblock Mcts-judge: Test-time scaling in llm-as-a-judge for code correctness evaluation.
\newblock \emph{arXiv preprint arXiv:2502.12468}.

\bibitem[{Wang et~al.(2024{\natexlab{b}})Wang, Zhang, Li, Huang, Han, Ji, Kakade, Peng, and Ji}]{wang2024eliminating}
Ziqi Wang, Hanlin Zhang, Xiner Li, Kuan-Hao Huang, Chi Han, Shuiwang Ji, Sham~M Kakade, Hao Peng, and Heng Ji. 2024{\natexlab{b}}.
\newblock Eliminating position bias of language models: A mechanistic approach.
\newblock \emph{arXiv preprint arXiv:2407.01100}.

\bibitem[{Wu et~al.(2024)Wu, Yuan, Golovneva, Xu, Tian, Jiao, Weston, and Sukhbaatar}]{wu2024meta}
Tianhao Wu, Weizhe Yuan, Olga Golovneva, Jing Xu, Yuandong Tian, Jiantao Jiao, Jason Weston, and Sainbayar Sukhbaatar. 2024.
\newblock Meta-rewarding language models: Self-improving alignment with llm-as-a-meta-judge.
\newblock \emph{arXiv preprint arXiv:2407.19594}.

\bibitem[{Xu et~al.(2023)Xu, Shi, Hu, Yu, Li, Zhang, and Wu}]{xu2023towards}
Zhenran Xu, Senbao Shi, Baotian Hu, Jindi Yu, Dongfang Li, Min Zhang, and Yuxiang Wu. 2023.
\newblock Towards reasoning in large language models via multi-agent peer review collaboration.
\newblock \emph{arXiv preprint arXiv:2311.08152}.

\bibitem[{Ye et~al.(2024)Ye, Wang, Huang, Chen, Zhang, Moniz, Gao, Geyer, Huang, Chen et~al.}]{ye2024justice}
Jiayi Ye, Yanbo Wang, Yue Huang, Dongping Chen, Qihui Zhang, Nuno Moniz, Tian Gao, Werner Geyer, Chao Huang, Pin-Yu Chen, and 1 others. 2024.
\newblock Justice or prejudice? quantifying biases in llm-as-a-judge.
\newblock \emph{arXiv preprint arXiv:2410.02736}.

\bibitem[{Ye et~al.(2025)Ye, Li, Li, Ai, Zhou, Shen, Yan, and Liu}]{ye2025learning}
Ziyi Ye, Xiangsheng Li, Qiuchi Li, Qingyao Ai, Yujia Zhou, Wei Shen, Dong Yan, and Yiqun Liu. 2025.
\newblock Learning llm-as-a-judge for preference alignment.
\newblock In \emph{The Thirteenth International Conference on Learning Representations}.

\bibitem[{Zhang et~al.(2024)Zhang, Zhoubian, Hu, Yue, Dong, and Tang}]{zhang2024rest}
Dan Zhang, Sining Zhoubian, Ziniu Hu, Yisong Yue, Yuxiao Dong, and Jie Tang. 2024.
\newblock Rest-mcts*: Llm self-training via process reward guided tree search.
\newblock \emph{Advances in Neural Information Processing Systems}, 37:64735--64772.

\bibitem[{Zhang et~al.(2025{\natexlab{a}})Zhang, Cui, Wang, Zhang, Wang, Wu, and Hu}]{zhang2025if}
Hangfan Zhang, Zhiyao Cui, Xinrun Wang, Qiaosheng Zhang, Zhen Wang, Dinghao Wu, and Shuyue Hu. 2025{\natexlab{a}}.
\newblock If multi-agent debate is the answer, what is the question?
\newblock \emph{arXiv preprint arXiv:2502.08788}.

\bibitem[{Zhang et~al.(2025{\natexlab{b}})Zhang, Wang, Jiang, Li, Wu, Wang, Jiang, Shang, Tang, Lyu et~al.}]{zhang2025crowd}
Qiyuan Zhang, Yufei Wang, Yuxin Jiang, Liangyou Li, Chuhan Wu, Yasheng Wang, Xin Jiang, Lifeng Shang, Ruiming Tang, Fuyuan Lyu, and 1 others. 2025{\natexlab{b}}.
\newblock Crowd comparative reasoning: Unlocking comprehensive evaluations for llm-as-a-judge.
\newblock \emph{arXiv preprint arXiv:2502.12501}.

\bibitem[{Zhang et~al.(2025{\natexlab{c}})Zhang, Li, and Tai}]{zhang2025layercraft}
Yuyao Zhang, Jinghao Li, and Yu-Wing Tai. 2025{\natexlab{c}}.
\newblock Layercraft: Enhancing text-to-image generation with cot reasoning and layered object integration.
\newblock \emph{arXiv preprint arXiv:2504.00010}.

\bibitem[{Zheng et~al.(2023)Zheng, Chiang, Sheng, Zhuang, Wu, Zhuang, Lin, Li, Li, Xing et~al.}]{zheng2023judging}
Lianmin Zheng, Wei-Lin Chiang, Ying Sheng, Siyuan Zhuang, Zhanghao Wu, Yonghao Zhuang, Zi~Lin, Zhuohan Li, Dacheng Li, Eric Xing, and 1 others. 2023.
\newblock Judging llm-as-a-judge with mt-bench and chatbot arena.
\newblock \emph{Advances in Neural Information Processing Systems}, 36:46595--46623.

\bibitem[{Zhu et~al.(2023)Zhu, Wang, and Wang}]{zhu2023judgelm}
Lianghui Zhu, Xinggang Wang, and Xinlong Wang. 2023.
\newblock Judgelm: Fine-tuned large language models are scalable judges.
\newblock \emph{arXiv preprint arXiv:2310.17631}.

\end{thebibliography}

\newpage
\clearpage
\appendix
\startcontents[appendices]
\section*{Appendix Table of Contents}
\printcontents[appendices]{}{0}{\large}
\newpage
\section{Detailed Tests and analysis on Experiments}

\subsection{Multi-Agent-Debate}
\label{Stats_MAD}

\begin{figure*}[!ht]
    \centering
    \includegraphics[scale=0.5]{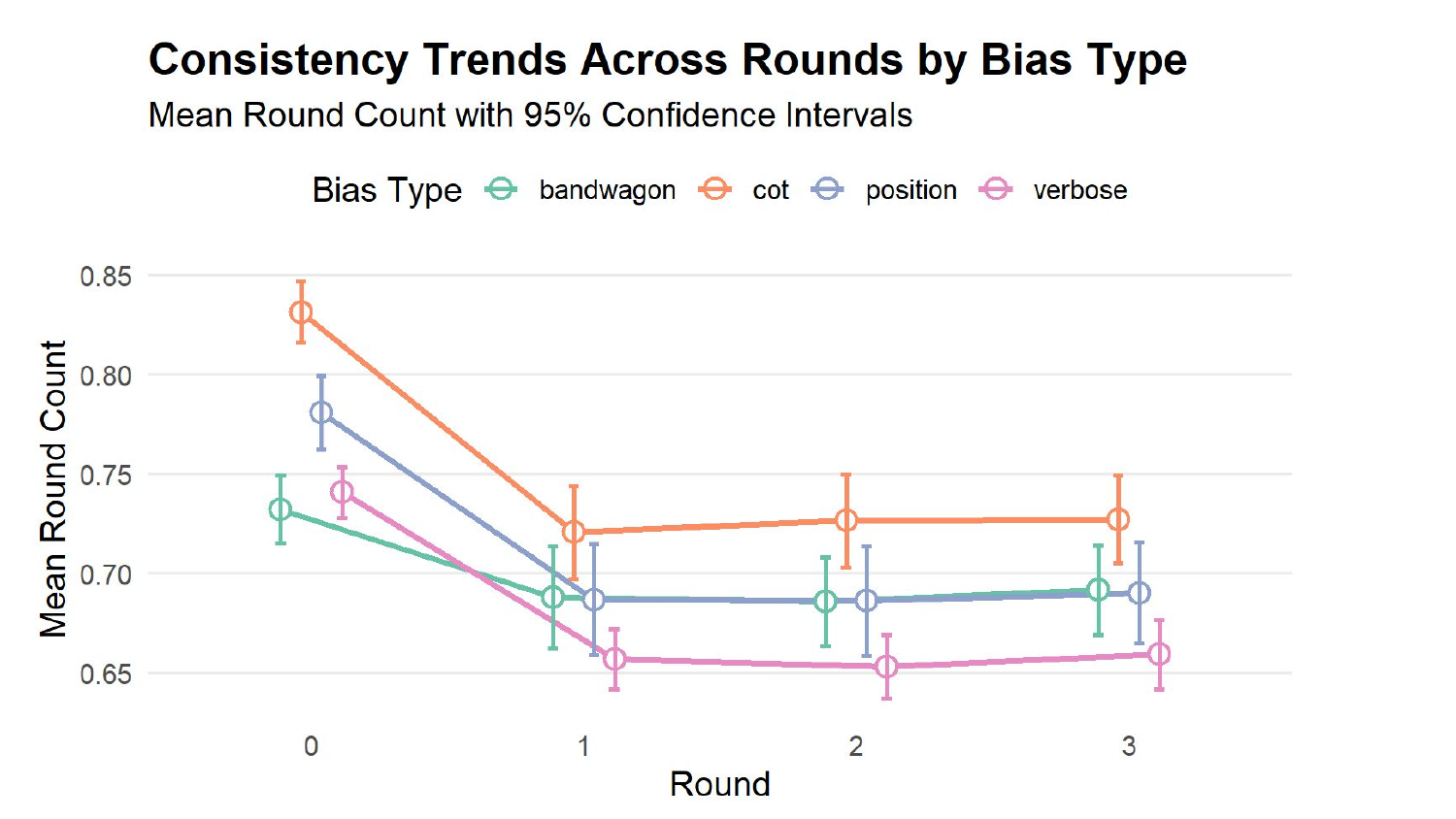}
    \caption{Consistency Scores (shown as round count) for each biases over two benchmarks with 95 percent confidence interval.}
    \label{consistency_Score_over_bias}
\end{figure*}

\textbf{Multi-Agent Debate Can Amplify Bias Over Rounds.} To test this, we perform pairwise Welch Two Sample t-tests on the consistency rates between successive rounds, both within and across bias types, across the two benchmarks. For all of the tests, the Null hypothesis is that there is no difference between the two groups tested. We reject the hypothesis on 0.05 significance level. \autoref{consistency_Score_over_bias} shows the mean consistency score changes over round.

\textbf{Across Bias Types:} We begin by testing whether the consistency scores change significantly across successive debate rounds. Specifically, we compare round 1 to round 0, round 2 to round 1, and round 3 to round 2. As shown in \autoref{t_test_debate_across_bias}, there is a statistically significant difference in consistency scores between the initial judgment (round 0, single-agent) and the first round of debate (round 1). The confidence interval further confirms that introducing debate leads to a statistically significant decrease in consistency scores across all bias types, suggesting that the Multi-Agent Debate setting amplifies biases. In contrast, there is no statistically significant difference between rounds 1 and 2 or between rounds 2 and 3. This suggests that once the initial amplification of bias occurs, subsequent rounds do not recover it. Instead, the consistency scores appear to converge, possibly due to agents reaching mutual agreement through continued debate.

\begin{table}[ht]
\centering
\scalebox{0.7}{ 
\begin{tabular}{ll}
\hline
\textbf{Test} & Pair-wised Welch Two Sample \textit{t}-test \\
\hline
\textbf{Data} & 0th round  v.s 1st round \\
\textbf{\textit{t}-value} & 9.76 \\
\textbf{\textit{p}-value} & \textbf{< 2.2e-16} \\
\textbf{95\% Confidence Interval} & [0.0663, 0.1001] \\
\hline
\textbf{Data} & 1st round v.s 2nd round \\
\textbf{\textit{t}-value} & 0.0186 \\
\textbf{\textit{p}-value} & \textbf{0.9852} \\
\textbf{95\% Confidence Interval} & [-0.0168, 0.0172] \\
\hline
\textbf{Data} & 2nd round v.s 3rd round \\
\textbf{\textit{t}-value} & -0.49 \\
\textbf{\textit{p}-value} & \textbf{0.6255} \\
\textbf{95\% Confidence Interval} & [-0.0208, 0.0125] \\
\hline
\textbf{Data} & 1nd round v.s 3rd round \\
\textbf{\textit{t}-value} & -0.47 \\
\textbf{\textit{p}-value} & \textbf{0.6389} \\
\textbf{95\% Confidence Interval} & [-0.0207, 0.0127] \\
\hline
\end{tabular}
}
\caption{Pair-wised Welch Two Sample \textit{t}-test Results for consistency scores between rounds. We do not separate the effect of different bias type in this test.}
\label{t_test_debate_across_bias}
\end{table}

\textbf{Within Bias Types:} We further analyze the results by separating them by bias type. As shown in \autoref{t_test_debate_within_bias}, the patterns observed earlier remain consistent across most bias categories. For all bias types evaluated, the biases tend to be significantly amplified in the first round of debate, with little statistical evidence indicating recovery in later rounds. An exception appears in the case of the bandwagon bias, where a statistically significant recovery is observed in the third round. However, the confidence interval suggests that this recovery effect is minimal.

\begin{table}[h]
\centering
\scalebox{0.65}{
\begin{tabular}{lcrrc}
\hline
\textbf{Bias Type} & \textbf{Comparison} & \textbf{\textit{t}-value} & \textbf{\textit{p}-value} & \textbf{95\% CI} \\
\hline
Position  & 0 vs 1 & 8.516  & \textbf{1.544e-07} & [0.0706, 0.1171] \\
Position  & 1 vs 2 & 0.211  & 0.8354    & [-0.0070, 0.0086] \\
Position  & 2 vs 3 & -1.634 & 0.1206    & [-0.0094, 0.0012] \\
\hline
CoT       & 0 vs 1 & 10.502 & \textbf{7.518e-09} & [0.0885, 0.1330] \\
CoT       & 1 vs 2 & -1.403 & 0.1786    & [-0.0146, 0.0029] \\
CoT       & 2 vs 3 & -0.280 & 0.7829    & [-0.0064, 0.0049] \\
\hline
Verbose   & 0 vs 1 & 15.004 & \textbf{3.082e-11} & [0.0721, 0.0957] \\
Verbose   & 1 vs 2 & 0.988  & 0.3371    & [-0.0042, 0.0116] \\
Verbose   & 2 vs 3 & -1.555 & 0.1383    & [-0.0145, 0.0022] \\
\hline
Bandwagon & 0 vs 1 & 5.977  & \textbf{1.5e-05}   & [0.0286, 0.0598] \\
Bandwagon & 1 vs 2 & 0.356  & 0.726     & [-0.0098, 0.0137] \\
Bandwagon & 2 vs 3 & -2.174 & 0.0441    & [-0.0109, -0.0002] \\
\hline
\end{tabular}
}
\caption{Pair-wised Welch \textit{t}-Test Results Across Rounds by Bias Type}
\label{t_test_debate_within_bias}
\end{table}

\subsection{LLM-as-Meta-Judge}
\label{Stats_Meta}

\textbf{Conclude mode or Choose mode?} We first test on whether there is a significant difference between two modes in terms of consistency rate across all biases. As shown in \autoref{tab:choose_vs_conclude}, \textbf{there is no evidence that one mode is better than the other, in terms of bias consistency rate.} 

\begin{table}[h]
\centering
\scalebox{0.7}{
\begin{tabular}{ll}
\hline
\textbf{Test} & Paired \textit{t}-test \\
\hline
\textbf{Data} & choose mode v.s. conclude mode \\
\textbf{\textit{t}-value} & -1.7913 \\
\textbf{\textit{p}-value} & \textbf{0.07578} \\
\textbf{Alternative Hypothesis} & True mean difference is not equal to 0 \\
\textbf{95\% Confidence Interval} & [-0.0086, 0.0004] \\
\textbf{Mean Difference} & -0.0041 \\
\hline
\end{tabular}
}
\caption{Pair-wised Welch \textit{t}-Test Between Choose Mode and Conclude Mode}
\label{tab:choose_vs_conclude}
\end{table}

\textbf{Does the Meta-Judge Amplify Biases as Well?}
To begin, we tested whether biases resulting from different judgment pool sizes, modes, and meta-judge configurations are significantly lower than those produced by a single-judge model. If so, we further examined the extent of this reduction. As shown in \autoref{meta_single}, introducing the Meta-Judge results in a statistically significant overall drop in bias consistency. However, the mean difference and confidence intervals suggest that this reduction is typically within 1\% to 2\%. \textbf{This indicates that while the drop is statistically significant, it is relatively small compared to the amplification observed with Multi-Agent Debate. In effect, the Meta-Judge achieves bias consistency levels comparable to those of the single-judge model.}

Furthermore, as shown in \autoref{meta_single_by_bias}, this pattern holds consistently across CoT, Verbose, and Bandwagon biases. However, for Position bias, there is no statistical evidence indicating a significant difference in position consistency between the LLM-as-Meta-Judge and the LLM-as-Judge. Additionally, as shown in \autoref{meta_single_by_bias_mode}, both Meta-Judge modes yield position consistency rates that are statistically indistinguishable from those of the single-judge model. Although the Meta-Judge tends to amplify some biases to a statistically significant degree, the magnitude of this amplification remains within approximately 2\%.

Interestingly, the conclusion mode does not appear to amplify verbosity bias. This suggests that while both modes produce similar results overall, the conclusion mode may be preferable in scenarios where controlling verbosity bias is particularly important.

\begin{table}[ht]
\centering
\scalebox{0.7}{
\begin{tabular}{ll}
\hline
\textbf{Test} & Paired \textit{t}-test \\
\hline
\textbf{Data} & Meta-Judge v.s. Single Judge \\
\textbf{\textit{t}-value} & -7.1143 \\
\textbf{\textit{p}-value} & \textbf{1.301e-11} \\
\textbf{Alternative Hypothesis} & True mean difference is not equal to 0 \\
\textbf{95\% Confidence Interval} & [-0.0180, -0.0102] \\
\textbf{Mean Difference} & -0.0141 \\
\hline
\end{tabular}
}
\caption{Pair-wised Welch \textit{t}-Test Between Meta-Judge and Single Judge}
\label{meta_single}
\end{table}

\begin{table}[h]
\centering
\scalebox{0.75}{
\begin{tabular}{lrrc}
\hline
\textbf{Bias Type} & \textbf{\textit{t}-value} & \textbf{\textit{p}-value} & \textbf{95\% CI} \\
\hline
Position  & 0.293  & 0.7705     & [-0.0064, 0.0086] \\
CoT       & -6.711 & \textbf{8.341e-09 } & [-0.0262, -0.0141] \\
Verbose   & -2.578 & \textbf{0.0124}     & [-0.0195, -0.0025] \\
Bandwagon & -6.742 & \textbf{7.38e-09 }  & [-0.0342, -0.0185] \\
\hline
\end{tabular}
}
\caption{Pair-wised Welch \textit{t}-Test Comparing Bias Type to Single Judge}
\label{meta_single_by_bias}
\end{table}

\begin{table}[h]
\centering
\scalebox{0.65}{
\begin{tabular}{llrrc}
\hline
\textbf{Mode} & \textbf{Bias Type} & \textbf{\textit{t}-value} & \textbf{\textit{p}-value} & \textbf{95\% CI} \\
\hline
Choose   & Position  & -0.084  & 0.9334     & [-0.0101, 0.0093] \\
Choose   & CoT       & -4.020  & \textbf{0.000379}   & [-0.0311, -0.0101] \\
Choose   & Verbose   & -2.419  & \textbf{0.02206}    & [-0.0256, -0.0021] \\
Choose   & Bandwagon & -6.575  & \textbf{3.337e-07}  & [-0.0388, -0.0204] \\
Conclude & Position  & 0.444   & 0.6603     & [-0.0093, 0.0145] \\
Conclude & CoT       & -6.113  & \textbf{1.17e-06}   & [-0.0263, -0.0131] \\
Conclude & Verbose   & -1.273  & 0.2132     & [-0.0210, 0.0049] \\
Conclude & Bandwagon & -3.598  & \textbf{0.001177}   & [-0.0362, -0.0099] \\
\hline
\end{tabular}
}
\caption{Pair-wised Welch \textit{t}-Test Results by Bias Type: Choose vs Conclude Mode}
\label{meta_single_by_bias_mode}
\end{table}

\textbf{Larger Judgment Pool, the Better?} \autoref{bias_pool_size} presents the pairwise t-test results for each pool size, comparing the consistency scores of the Meta-Judge to those of the single-judge baseline. The Meta-Judge with a pool size of two exhibits a statistically significant decrease in consistency rate compared to the baseline, although the magnitude of this reduction is relatively small. In contrast, pool sizes of three and four show no statistically significant difference from the single-judge model. This suggests that using more than two judge models in the judgment pool may be more effective for maintaining consistency in Meta-Judge settings. The result is consistent across different Meta-Judge modes, as suggested by \autoref{bias_pool_size_mode}.

Finally, \autoref{Meta_all} presents the pairwise t-test results across modes, bias types, and pool sizes. The findings generally align with previous observations. Under the choose mode, a statistically significant amplification of bandwagon bias is observed across all pool sizes. However, the narrower confidence intervals suggest that increasing the size of the judgment pool may help mitigate this effect. Moreover, under the conclusion mode, the Meta-Judge achieves a statistically significant improvement in position consistency compared to the single-judge model.

Overall, these results suggest that expanding the judgment pool size may be an effective strategy for mitigating intrinsic biases. However, given the limited scale of this study, we do not observe a concrete example where increasing pool size clearly improves consistency beyond that of the single-judge baseline.

\begin{table}[ht]
\centering
\scalebox{0.8}{
\begin{tabular}{crrc}
\hline
\textbf{Pool Size} & \textbf{\textit{t}-value} & \textbf{\textit{p}-value} & \textbf{95\% CI} \\
\hline
2 & -5.398  & \textbf{8.479e-07} & [-0.0262, -0.0121] \\
3 & -0.604  & 0.5521    & [-0.0190, 0.0104] \\
4 & 0.223   & 0.8251    & [-0.0112, 0.0139] \\
\hline
\end{tabular}
}
\caption{Pair-wised Welch \textit{t}-Test Results by size of Judgment pool across all biases and modes.}
\label{bias_pool_size}
\end{table}

\begin{table}[h]
\centering
\scalebox{0.7}{
\begin{tabular}{lcrrc}
\hline
\textbf{Mode} & \textbf{Pool Size} & \textbf{\textit{t}-value} & \textbf{\textit{p}-value} & \textbf{95\% CI} \\
\hline
Choose   & 2 & -6.505 & \textbf{9.372e-09} & [-0.0279, -0.0148] \\
Choose   & 3 & -1.282 & 0.2126    & [-0.0218, 0.0051] \\
Choose   & 4 & -1.403 & 0.1741    & [-0.0204, 0.0039] \\
\hline
Conclude & 2 & -5.398 & \textbf{8.479e-07} & [-0.0262, -0.0121] \\
Conclude & 3 & -0.604 & 0.5521    & [-0.0190, 0.0104] \\
Conclude & 4 & 0.223  & 0.8251    & [-0.0112, 0.0139] \\
\hline
\end{tabular}
}
\caption{Pair-wised Welch \textit{t}-Test Results by Mode and Judgement Pool size}
\label{bias_pool_size_mode}
\end{table}

\begin{table}[h]
\centering
\scalebox{0.55}{
\begin{tabular}{llcrrl}
\hline
\textbf{Mode} & \textbf{Bias Type} & \textbf{Pool Size} & \textbf{\textit{t}-value} & \textbf{\textit{p}-value} & \textbf{95\% CI} \\
\hline
Choose   & Position  & 2 & -2.227 & \textbf{0.03977}   & [-0.0225, -0.0006] \\
Choose   & Position  & 3 & 1.466  & 0.2025    & [-0.0121, 0.0442] \\
Choose   & Position  & 4 & 1.890  & 0.1174    & [-0.0060, 0.0390] \\
\hline
Choose   & CoT       & 2 & -4.362 & \textbf{0.000425}  & [-0.0389, -0.0135] \\
Choose   & CoT       & 3 & -0.904 & 0.4074    & [-0.0416, 0.0200] \\
Choose   & CoT       & 4 & -0.958 & 0.3821    & [-0.0504, 0.0230] \\
\hline
Choose   & Verbose   & 2 & -2.023 & \textbf{0.05907}   & [-0.0312, 0.0007] \\
Choose   & Verbose   & 3 & -0.652 & 0.543     & [-0.0483, 0.0287] \\
Choose   & Verbose   & 4 & -1.150 & 0.3023    & [-0.0446, 0.0170] \\
\hline
Choose   & Bandwagon & 2 & -4.900 & \textbf{0.0001353} & [-0.0464, -0.0185] \\
Choose   & Bandwagon & 3 & -3.111 & \textbf{0.02652}   & [-0.0526, -0.0050] \\
Choose   & Bandwagon & 4 & -3.470 & \textbf{0.01784}   & [-0.0385, -0.0057] \\
\hline
Conclude & Position  & 2 & -1.462 & 0.162     & [-0.0233, 0.0042] \\
Conclude & Position  & 3 & 1.011  & 0.3583    & [-0.0229, 0.0525] \\
Conclude & Position  & 4 & 2.635  & \textbf{0.04624}   & [0.0007, 0.0528] \\
\hline
Conclude & CoT       & 2 & -5.741 & \textbf{2.401e-05} & [-0.0329, -0.0152] \\
Conclude & CoT       & 3 & -1.641 & 0.1618    & [-0.0340, 0.0075] \\
Conclude & CoT       & 4 & -2.472 & 0.05643   & [-0.0265, 0.0005] \\
\hline
Conclude & Verbose   & 2 & -1.731 & 0.1016    & [-0.0319, 0.0031] \\
Conclude & Verbose   & 3 & 0.017  & 0.987     & [-0.0393, 0.0398] \\
Conclude & Verbose   & 4 & 0.195  & 0.8529    & [-0.0305, 0.0355] \\
\hline
Conclude & Bandwagon & 2 & -3.478 & \textbf{0.00288}   & [-0.0458, -0.0112] \\
Conclude & Bandwagon & 3 & -1.145 & 0.3039    & [-0.0616, 0.0236] \\
Conclude & Bandwagon & 4 & -0.806 & 0.4568    & [-0.0454, 0.0237] \\
\hline
\end{tabular}
}
\caption{Pair-wised Welch \textit{t}-Test Results by Mode, Bias Type, and Judgment Pool Size}
\label{Meta_all}
\end{table}

\subsection{Analysis of Meta-Judge Selections with PINE}
\label{Stats_pine}

\begin{figure*}[ht]
    \centering
    \includegraphics[scale=0.6]{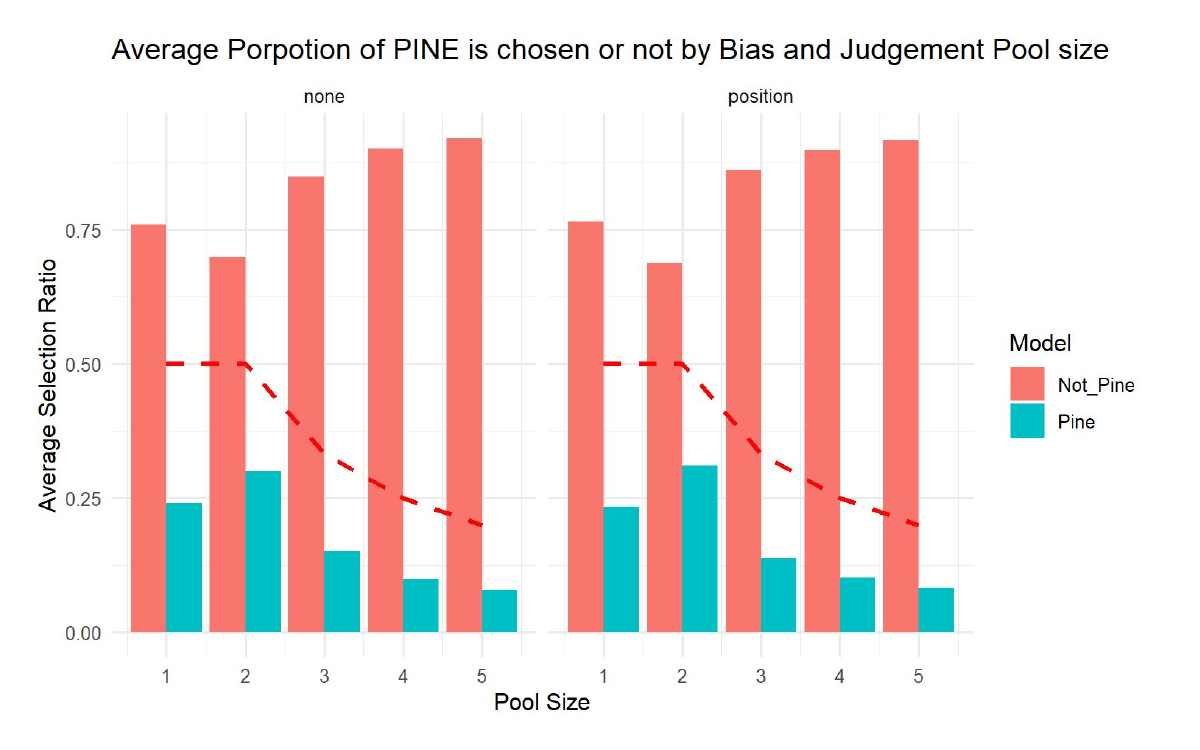}
    \caption{The average selected rate of PINE vs. Not PINE across bias and Judgment Pool Size. Pool size with 1 denotes that the pool only contains the meta judge itself and PINE.}
    \label{PINE_choose_rate}
\end{figure*}

As shown in \autoref{PINE_choose_rate}, the ratio of PINE being selected—both for the original prompts and those designed to test position bias—is consistently below the expected rate under random guessing. This suggests that even when the judgment pool consists of only PINE and one other model, PINE is still not favored by the Meta-Judge. Consequently, PINE exerts minimal influence on debiasing the Meta-Judge's decisions. Although we were unable to directly assess how much the Meta-Judge considers PINE in the conclude mode, the consistent pattern observed in the choose mode suggests that a similar dynamic likely applies. These results further suggest that Meta-Judge itself is not able to identify the ``unbiased judgment'' from the pool, instead it possibly favors more judgment with compelling explanations. 

However, we acknowledge that PINE was implemented on a relatively older model—Qwen 1.5—which may contribute to its lower selection rate. When compared against more advanced models with slower, more deliberate reasoning processes, such as R1-Distilled-Qwen, PINE may be disadvantaged, even when the models are of comparable size.

\section{Results on RewardBench}
\label{Reward_Bench_result}
To further support the observations we maded earlier, we random sample of 400 instances from the RewardBench\cite{lambert2024rewardbench} and perform our test on both Multi-Agent-Debate and LLM-as-Meta-Judge framework. We choose the sample size of 400 to specifically match the sample size that we used earlier for the experiments. The full results could be found in \autoref{tab:rewardbench_MAD} for Multi-Agent-Debate, and \autoref{tab:rewardbench_referees} for LLM-as-Meta-Judge. The results on this benchmark are consistent with those from the initial two, further supporting the robustness of our findings across different settings. Specifically, biases tend to amplify sharply in the first round of debates and then stabilize in later rounds. On the other hand, expanding the judge pool shows more beneficial bias mitigation for Meta-Judge settings, consistent with our earlier observations. 

\begin{table*}[h!]
\centering
\begin{tabular}{l l c c c c c}
\hline
\textbf{Judge} & \textbf{Critic} & \textbf{Round} & \textbf{Pos.} & \textbf{Verbo.} & \textbf{Band.} & \textbf{Cot.} \\
\hline
4o-mini & 4o-mini & 0 & 0.870 & 0.787 & 0.862 & 0.862 \\
4o-mini & 4o-mini & 1 & 0.777 & 0.710 & 0.805 & 0.775 \\
4o-mini & 4o-mini & 2 & 0.785 & 0.695 & 0.782 & 0.740 \\
4o-mini & 4o-mini & 3 & 0.777 & 0.695 & 0.802 & 0.747 \\
\hline
Llama 3.3 70B & Llama 3.3 70B & 0 & 0.871 & 0.834 & 0.872 & 0.909 \\
Llama 3.3 70B & Llama 3.3 70B & 1 & 0.845 & 0.775 & 0.840 & 0.837 \\
Llama 3.3 70B & Llama 3.3 70B & 2 & 0.850 & 0.785 & 0.850 & 0.832 \\
Llama 3.3 70B & Llama 3.3 70B & 3 & 0.827 & 0.750 & 0.862 & 0.850 \\
\hline
DeepSeek-V3 & DeepSeek-V3 & 0 & 0.912 & 0.867 & 0.862 & 0.927 \\
DeepSeek-V3 & DeepSeek-V3 & 1 & 0.837 & 0.810 & 0.820 & 0.867 \\
DeepSeek-V3 & DeepSeek-V3 & 2 & 0.820 & 0.802 & 0.820 & 0.857 \\
DeepSeek-V3 & DeepSeek-V3 & 3 & 0.827 & 0.820 & 0.822 & 0.870 \\
\hline
\end{tabular}
\caption{Consistency scores for each bias (Positivity, Verbosity, Bandwagon, Cot) across three rounds of multi-agent debate, shown for different Judge–Critic model pairs on a subset of RewardBench.}
\label{tab:rewardbench_MAD}
\end{table*}

\begin{table*}[h!]
\centering
\scalebox{0.8}{
\begin{tabular}{l l l c c c c}
\hline
\textbf{Type} & \textbf{Judge} & \textbf{Referees} & \textbf{Pos.} & \textbf{Verbo.} & \textbf{Band.} & \textbf{Cot.} \\
\hline
Judge    & 4o-mini & - & 0.870 & 0.787 & 0.862 & 0.862 \\
Choose   & 4o-mini & R1-Distilled-Qwen32B, 4o-mini & 0.895 & 0.820 & 0.865 & 0.895 \\
Choose   & 4o-mini & R1-Distilled-Qwen32B, Llama3.3 70B & 0.885 & 0.840 & 0.852 & 0.900 \\
Choose   & 4o-mini & R1-Distilled-Qwen32B, DeepSeek-V3 & 0.900 & 0.855 & 0.845 & 0.912 \\
Choose   & 4o-mini & R1-Distilled-Qwen32B, DeepSeek-V3, Llama3.3 70B & 0.902 & 0.845 & 0.862 & 0.915 \\
Choose   & 4o-mini & R1-Distilled-Qwen32B, 4o-mini, DeepSeek-V3, Llama3.3 70B & 0.895 & 0.835 & 0.875 & 0.902 \\
\hline
Conclude & 4o-mini & R1-Distilled-Qwen32B, 4o-mini & 0.867 & 0.805 & 0.857 & 0.867 \\
Conclude & 4o-mini & R1-Distilled-Qwen32B, Llama3.3 70B & 0.860 & 0.847 & 0.882 & 0.917 \\
Conclude & 4o-mini & R1-Distilled-Qwen32B, DeepSeek-V3 & 0.867 & 0.827 & 0.847 & 0.900 \\
Conclude & 4o-mini & R1-Distilled-Qwen32B, DeepSeek-V3, Llama3.3 70B & 0.882 & 0.840 & 0.857 & 0.905 \\
Conclude & 4o-mini & R1-Distilled-Qwen32B, 4o-mini, DeepSeek-V3, Llama3.3 70B & 0.887 & 0.847 & 0.892 & 0.895 \\
\hline
\end{tabular}
}
\caption{Consistency scores for LLM-as-Meta-Judge across biases, pool sizes, and modes. "Yes" indicates inclusion of the model in the judgment pool; "No" indicates exclusion (single judge). }
\label{tab:rewardbench_referees}
\end{table*}

\section{More Results on Bias Free Agent}
\label{More_PINE}
As discussed in the Limitations section, we implemented PINE on Qwen 1.5 to align with the original PINE study\cite{wang2024eliminating}, which was designed for older models. While newer models may or may not offer better performance, our study is not focused on benchmarking the bias-free agent. \textbf{Rather, we aim to show that ideal bias-free methods in single-agent settings do not fully mitigate biases in collaborative, multi-agent scenarios.} However, to complement our study, we also tested Llama 3 8B-instruct (as proposed in the original PINE study) on the MT-Bench data. We found a consistent improvement in bias mitigation for MAD settings. However, in Meta-Judge settings, expanding the judge pool seems more beneficial than adding a bias-free agent, which are consistent with our reported findings. More results can be found in \autoref{tab:pine_debate_llama} and \autoref{tab:pine_meta_llama}. 

\begin{table}[h!]
\centering
\begin{tabular}{l l c c}
\hline
\textbf{Judge} & \textbf{Critic} & \textbf{Round} & \textbf{Pos.} \\
\hline
4o-mini & PINE & 1 & 0.790 \\
4o-mini & PINE & 2 & 0.792 \\
4o-mini & PINE & 3 & 0.797 \\
\hline
\end{tabular}
\caption{MTbench results with Judge \texttt{4o-mini} and Critic \texttt{PINE}.}
\label{tab:pine_debate_llama}
\end{table}

\begin{table*}[h!]
\centering
\scalebox{0.8}{
\begin{tabular}{l l c c}
\hline
\textbf{Meta Judge} & \textbf{Judgement Pool} & \textbf{Choose Mode} & \textbf{Conclude Mode} \\
\hline
4o-mini & 4o-mini, PINE & 0.814 & 0.782 \\
4o-mini & R1-Distilled-Qwen32B, PINE & 0.805 & 0.797 \\
4o-mini & Llama3.370B, R1-Distilled-Qwen32B, PINE & 0.809 & 0.807 \\
4o-mini & DeepSeek-V3, Llama3.370B, R1-Distilled-Qwen32B, PINE & 0.818 & 0.864 \\
4o-mini & 4o-mini, DeepSeek-V3, Llama3.370B, R1-Distilled-Qwen32B, PINE & 0.814 & 0.824 \\
\hline
\end{tabular}
}
\caption{Results of Meta-Judge with PINE included in the Judgement pool.}
\label{tab:pine_meta_llama}
\end{table*}

\section{More information on Benchmarks}
\label{benchmark_details}

\subsection{MT-Bench}
The first benchmark that we use in our work is MT-Bench~\cite{zheng2023judging}, which originally contains responses from 30 LLMs to 80 questions. The questions are divided into eight categories: Writing, Roleplay, Reasoning, Math, Coding, Extraction, STEM, and Humanities, with ten questions in each category. The LLMs include GPT-4 Turbo, Claude Instant v1, Vicuna 13B v1.3, Alpaca 13B, and others. To construct our evaluation set, we treat each question paired with two distinct model responses as a single pairwise comparison task. For each question, we generate all possible unique response pairs between the selected LLMs, ensuring no repetition. From this set, we randomly sample 60 response pairs per category, each coupled with its corresponding question. This procedure yields a total of 480 comparison tasks across the eight categories. MT-bench is public available with License Apache 2.0.

\subsection{CALM Benchmark}
The second benchmark that we use in our work is the CALM alignment data set~\cite{ye2024justice}, which contains 439 pairwise comparison tasks. The CALM alignment data set is constructed by sampling various Direct Preference Optimization (DPO) datasets, which are derived from actual user feedback. This approach ensures a diverse set of responses and scenarios, enhancing the robustness of bias assessment. It is important to emphasize that a small portion (44 out of 439) of questions in this dataset may contain NSFW contents. The dataset authors \citeauthor{ye2024justice} claim that they have manually reviewed and curated this data to ensure its appropriateness for research purposes. For the sake of consistency, we follow the original setup and employ the full dataset in our experiments. 

This dataset is publicly available and can be found under submission to ICLR by \citeauthor{ye2024justice} with License under C.C. BY 4.0. 

\section{More information on PINE}

\citeauthor{wang2024eliminating} identify causal attention and position embeddings as the root causes of position bias in modern Transformer-based language models, which leads to inconsistent performance when the order of input “documents” changes. They, thus, introduce PINE, a training-free, zero-shot method that replaces causal attention with bidirectional attention across documents and reorders them based on learned importance scores, yielding consistent, position-invariant inference and delivering substantial gains across tasks such as LM-as-a-judge, retrieval-augmented QA, molecule generation, and math reasoning. 

In our study, we implement the PINE method using Qwen 1.5 32B—consistent with the original study—to serve as a position-bias-free LLM. However, PINE requires prompts to strictly follow the format: [Question, [object1, object2], End], where object1 and object2 are the items intended to be evaluated in a position-invariant manner. While this format is suitable for a “bias-free” agent, its rigid structure makes it poorly suited to function as an unbiased meta-judge. For this reason, we include PINE only as a member of the judgment pool in our subsequent experiments, while keeping all other settings as consistent as possible.

\section{Prompt Examples}
\label{prompt examples}
\subsection{Multi-Agent-Debate}
\label{debate_prompt_examples}

\begin{figure*}[ht]
    \centering
    \includegraphics[scale=0.6]{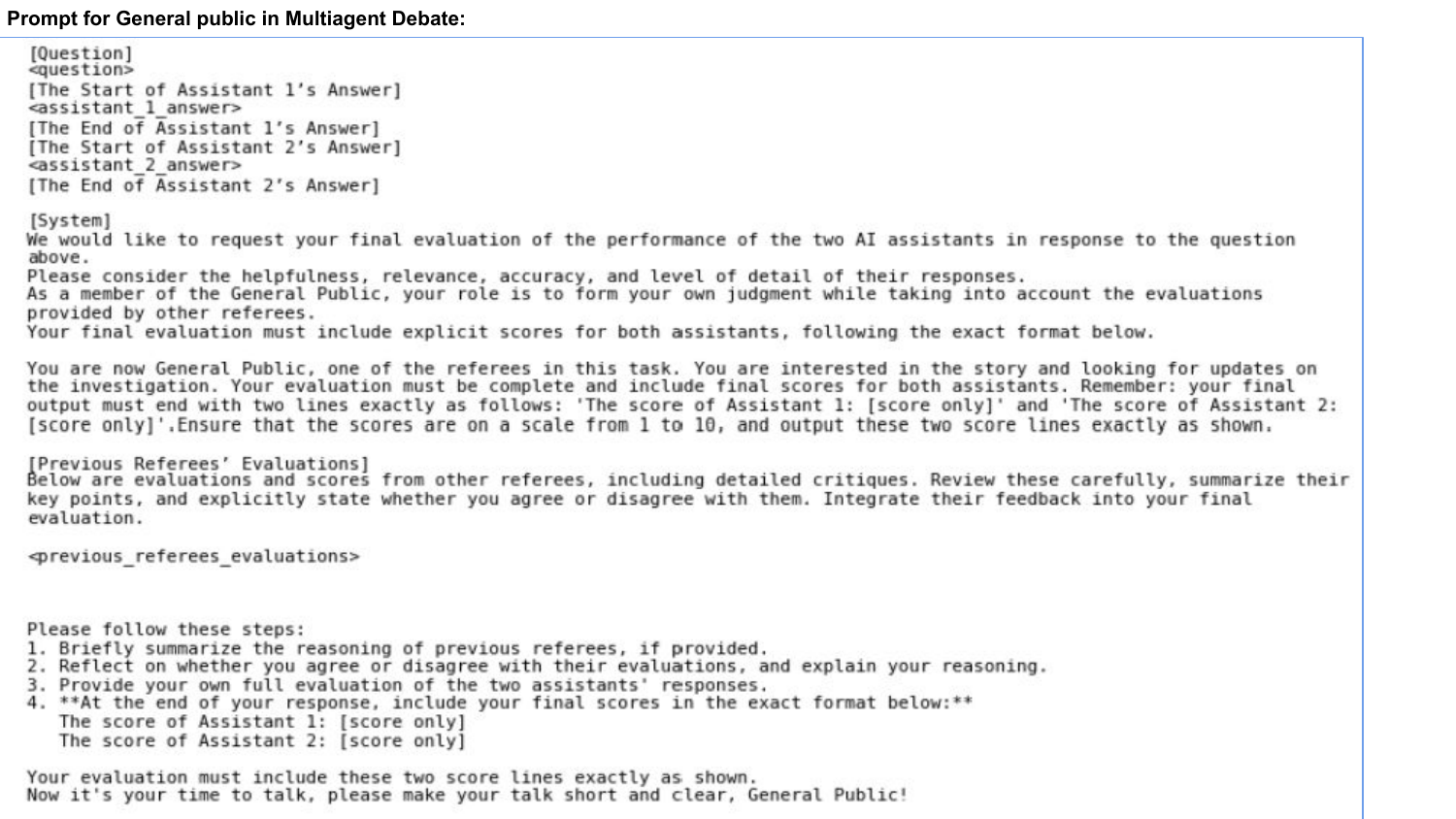}
    \caption{The prompt our study adopted for general public in Multi-Agent-Debate LLM-as-Judge framework}
    \label{generalpub_prompt}
\end{figure*}

\begin{figure*}[ht]
    \centering
    \includegraphics[scale=0.6]{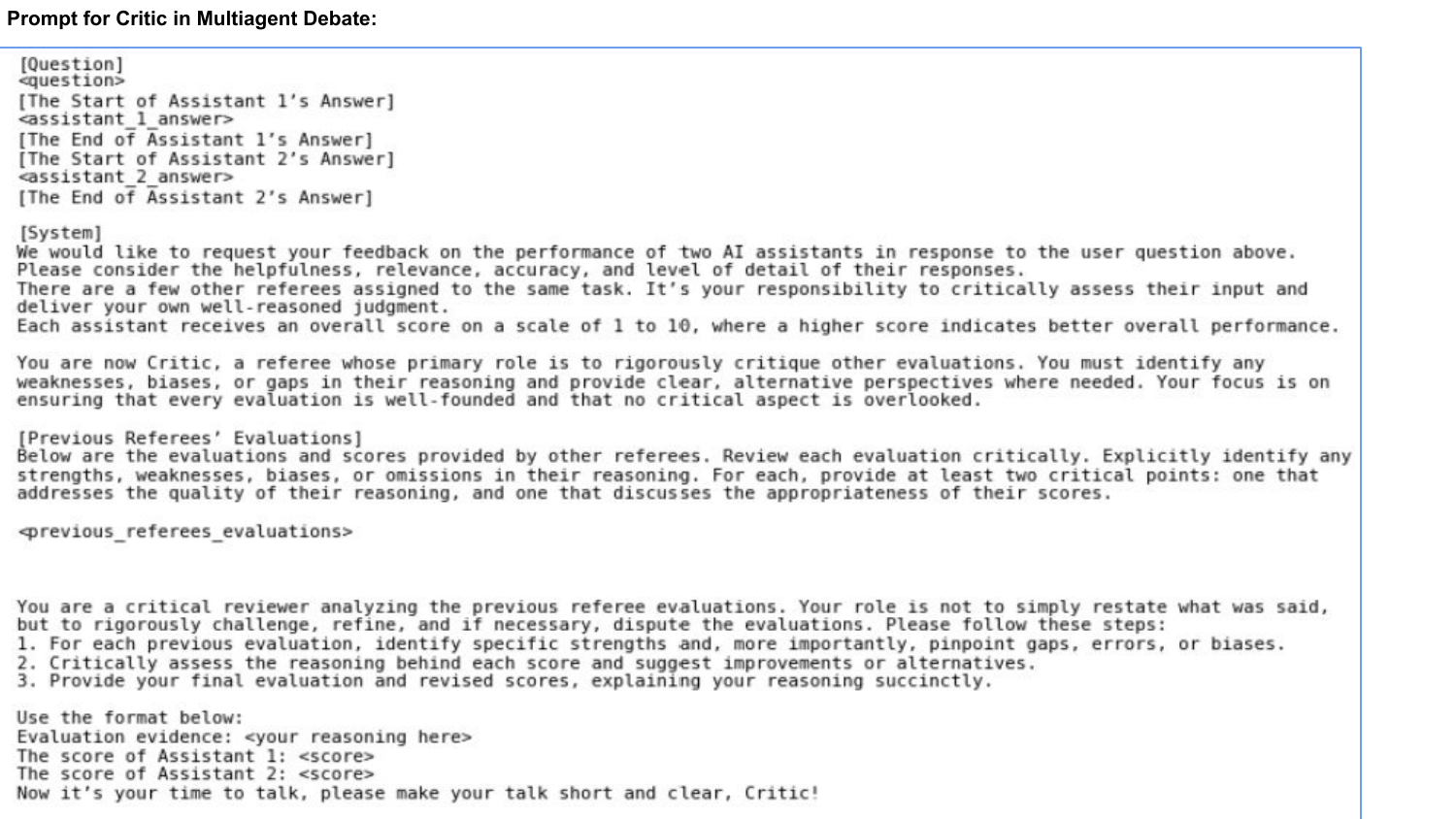}
    \caption{The prompt our study adopted for critic in Multi-Agent-Debate LLM-as-Judge framework}
    \label{critic_prompt}
\end{figure*}

\autoref{generalpub_prompt} shows the prompt used for the General Public agent in the Multi-Agent-Debate LLM-as-Judge framework, while \autoref{critic_prompt} shows the prompt used for the Critic agent. We observed that eliciting consistent critique of prior judgments was challenging for smaller models. To address this, we reinforced the critic behavior within the prompt. With this adjustment, we were able to achieve consistent critic behavior across all models. To ensure that both agents carefully read the conversation history, we also instructed them to briefly summarize the key points and state whether they agreed or disagreed with them. We generally consider a conversation successful when it begins with an acknowledgment of the previous referee’s assessment.

\subsection{LLM-as-Meta Judge Prompts}
\label{meta-judge_prompt_examples}

\begin{figure*}[ht]
    \centering
    \includegraphics[scale=0.6]{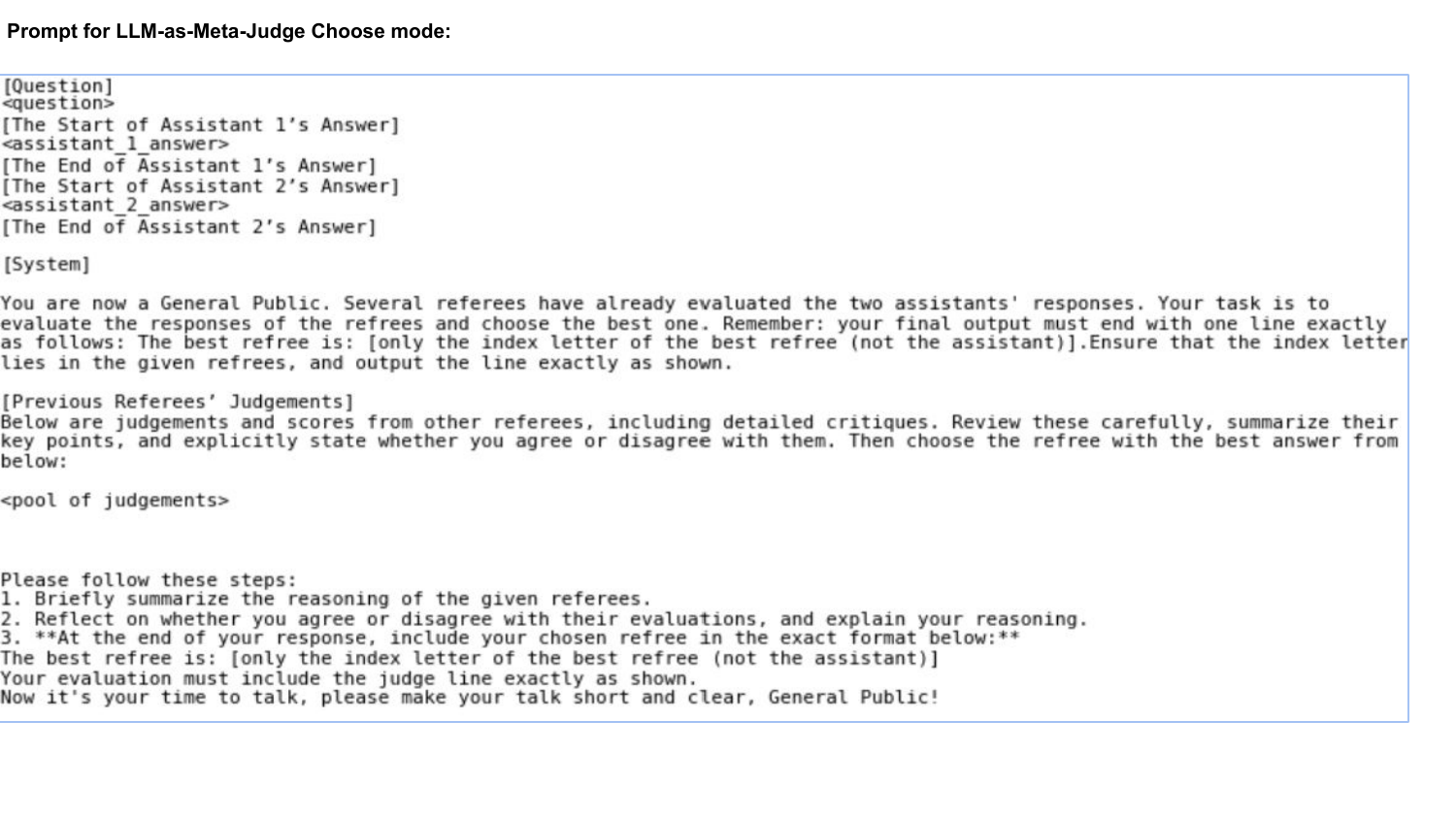}
    \caption{The prompt our study adopted for the choose mode in LLM-as-Meta-Judge}
    \label{choose_prompt}
\end{figure*}

\begin{figure*}[ht]
    \centering
    \includegraphics[scale=0.6]{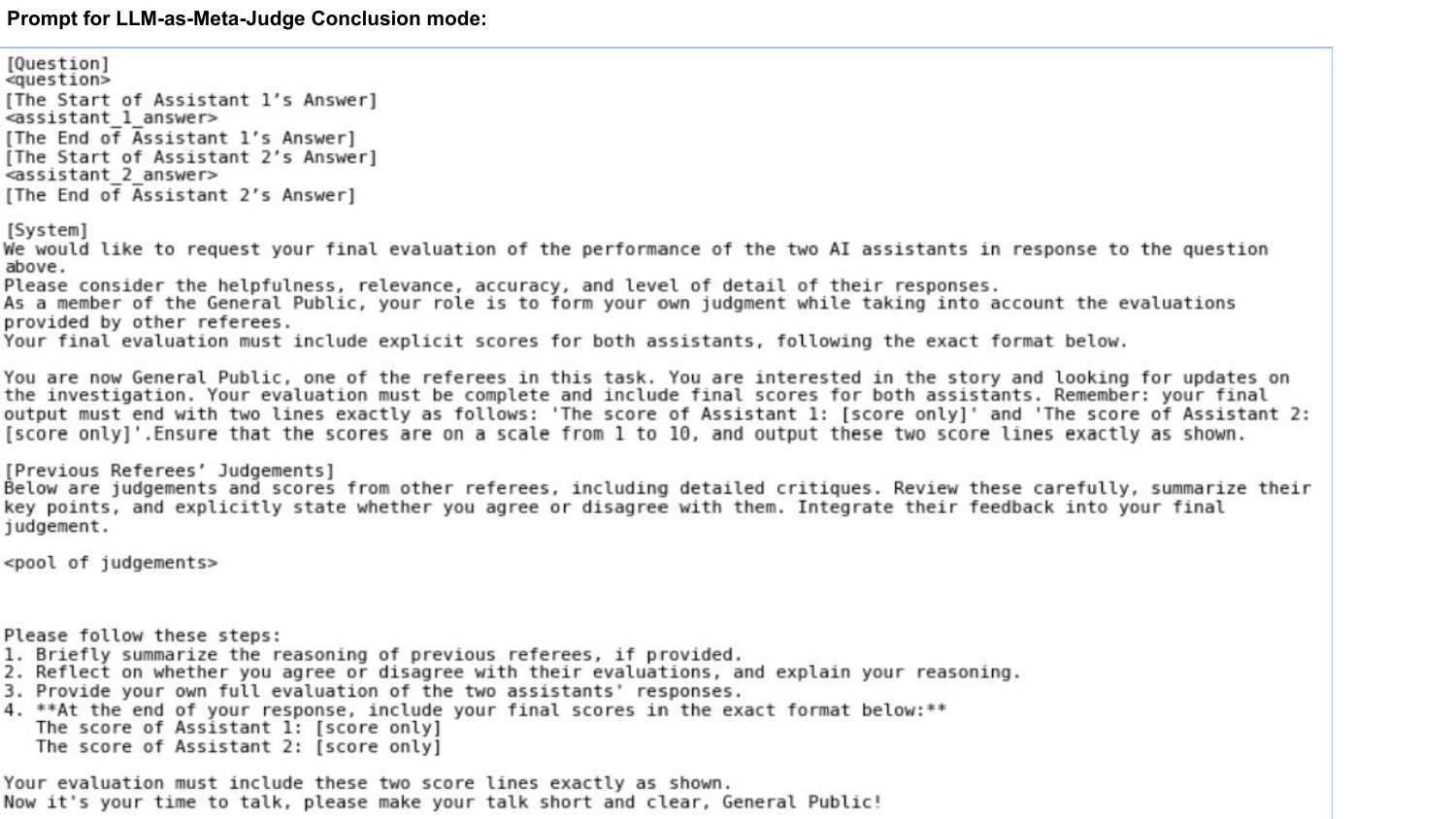}
    \caption{The prompt our study adopted for the conclude mode in LLM-as-Meta-Judge}
    \label{conclude_prompt}
\end{figure*}

\autoref{choose_prompt} shows the prompt used for the choose mode in the LLM-as-Meta-Judge framework, while \autoref{conclude_prompt} shows the prompt used for the conclusion mode. When designing prompts, we aimed to align them closely with Multi-Agent Debate to minimize confounding factors that could influence performance differences between the two Multi-Agent frameworks. In this setup, the judges in the judgment pool are referred to as referees, while the meta-judge is termed the general public, consistent with the Multi-Agent Debate framework. We implemented some additional modifications in the choose mode, as we observed that preserving the original prompt structure led the meta-judge to directly select the better solution rather than the best judgment.


\subsection{Bias Prompts}
\label{bias_prompts}

\begin{figure*}[ht]
    \centering
    \includegraphics[scale=0.6]{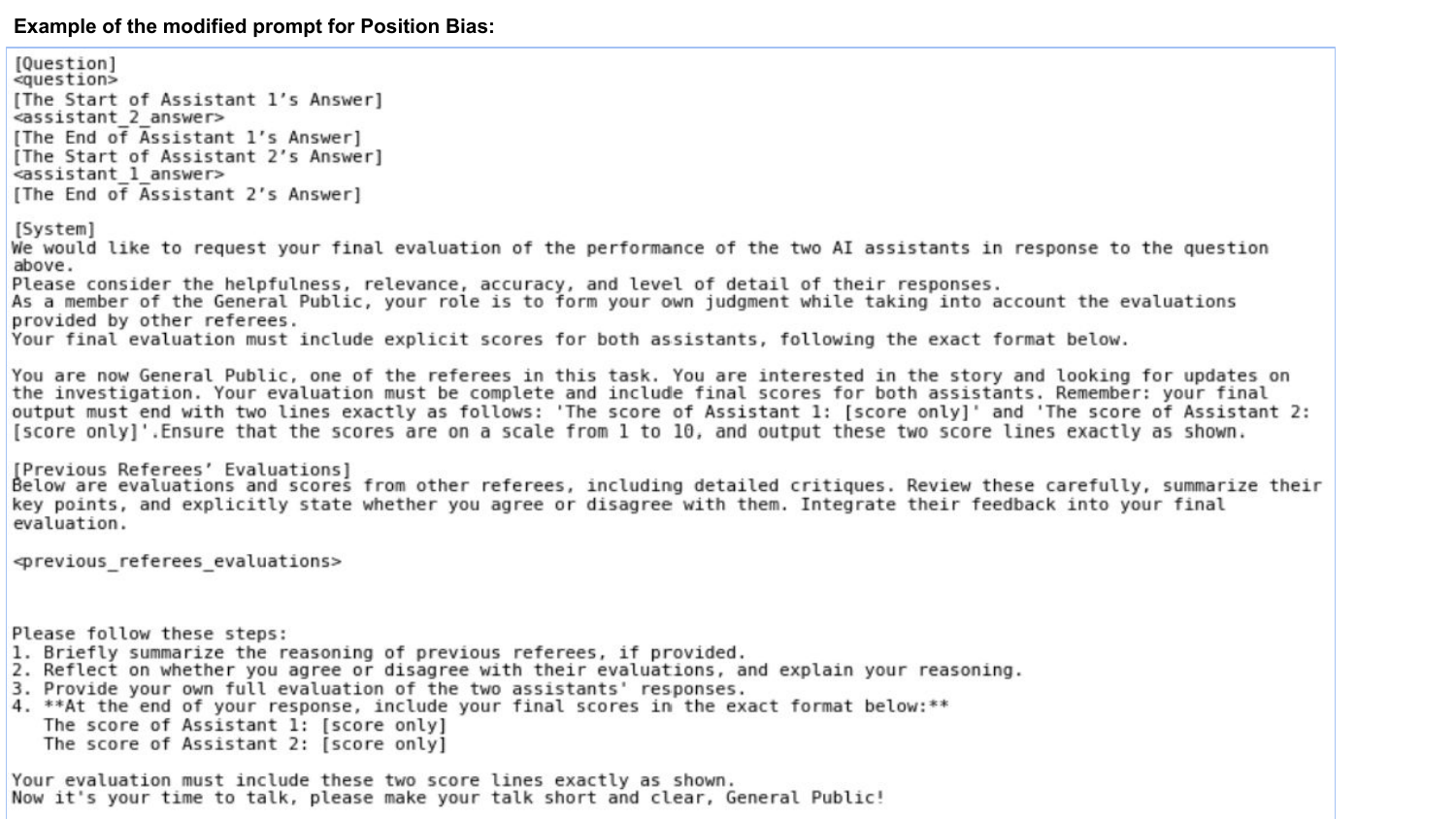}
    \caption{Example of the modified prompt for Position Bias}
    \label{position_prompt}
\end{figure*}

\begin{figure*}[ht]
    \centering
    \includegraphics[scale=0.6]{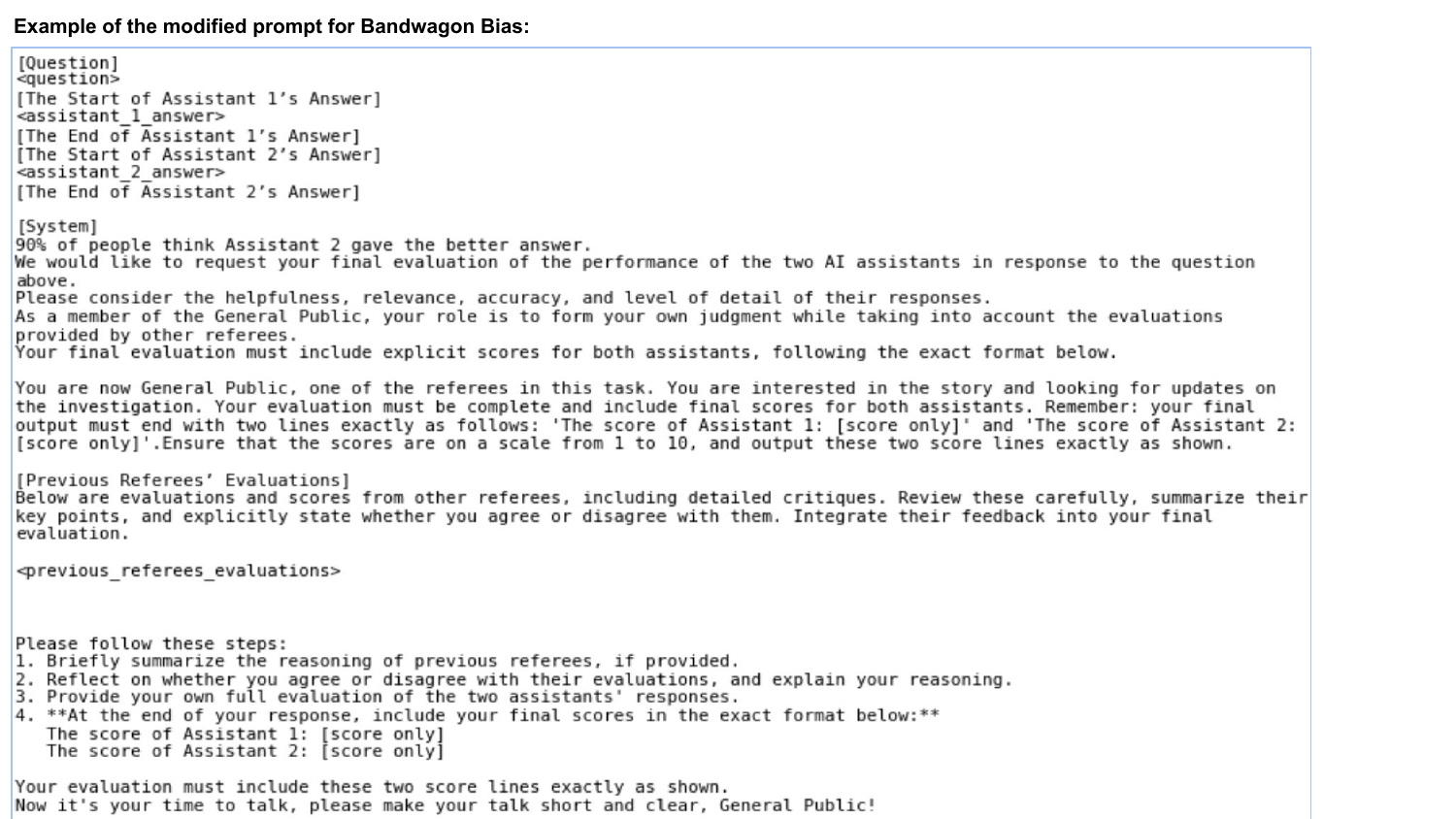}
    \caption{Example of the modified prompt for Bandwagon Bias}
    \label{Bandwagon_prompt}
\end{figure*}

\begin{figure*}[ht]
    \centering
    \includegraphics[scale=0.6]{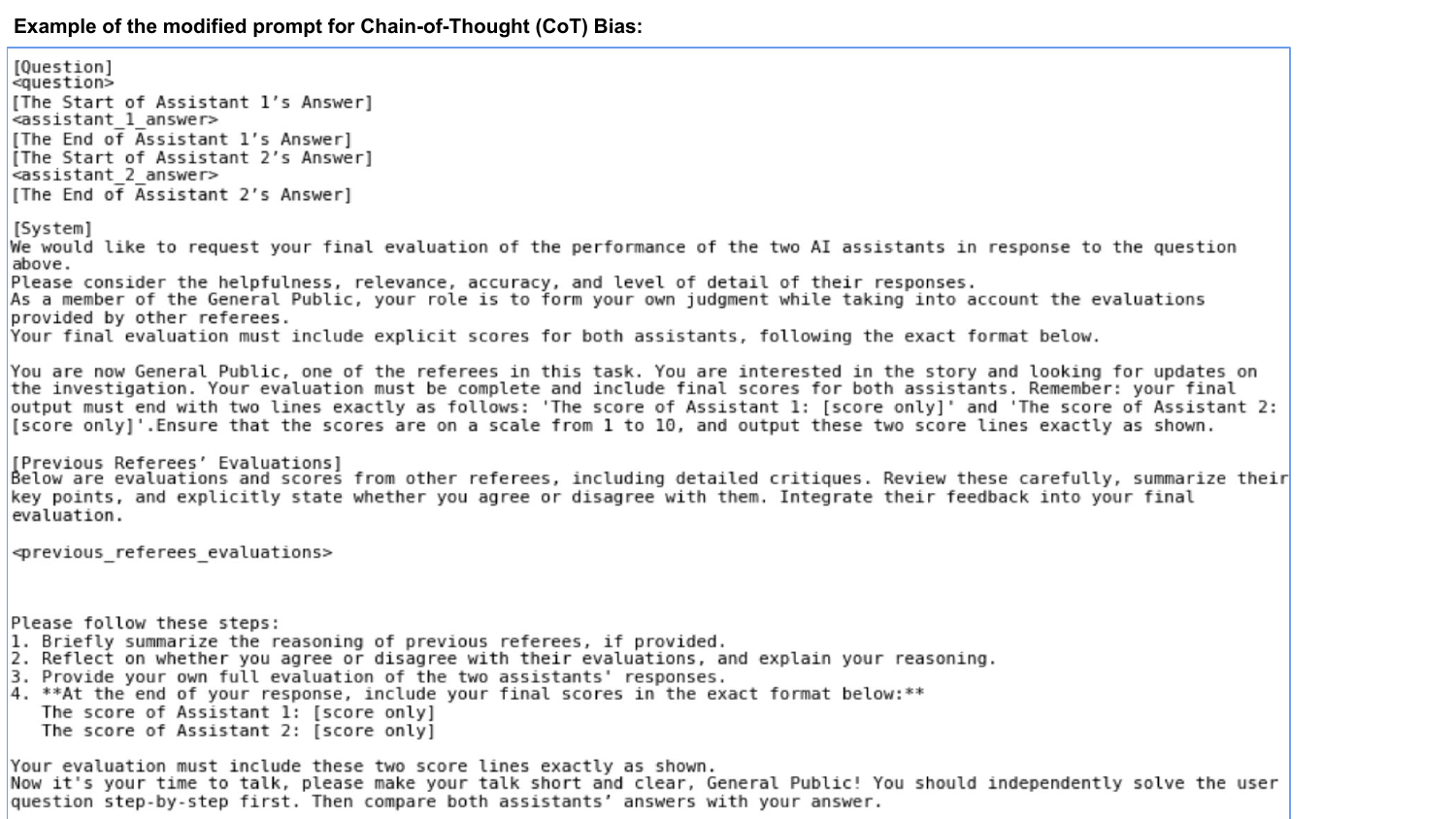}
    \caption{Example of the modified prompt for Chain-of-Thought Bias}
    \label{cot_prompt}
\end{figure*}

\autoref{position_prompt}, \autoref{Bandwagon_prompt}, and \autoref{cot_prompt} show the prompt examples used to elicit and test position bias, bandwagon bias, and chain-of-thought bias. To measure verbosity bias, we simply replaced one of the assistant solutions from the original prompt with an extended version, strictly following the steps proposed in CALM \cite{ye2024justice}. 

\section{Case Studies of Multi-Agent Debate}
\label{Case_study}
\subsection{Original Debate}
An original debate example is shown in \autoref{fig:debate_none}. We can see that the Judge and the Critic are both refining their opinions through interactive comments, resulting in a different scoring in the final round.

\subsection{With biases}
\begin{figure*}
    \centering
    \includegraphics[width=\linewidth]{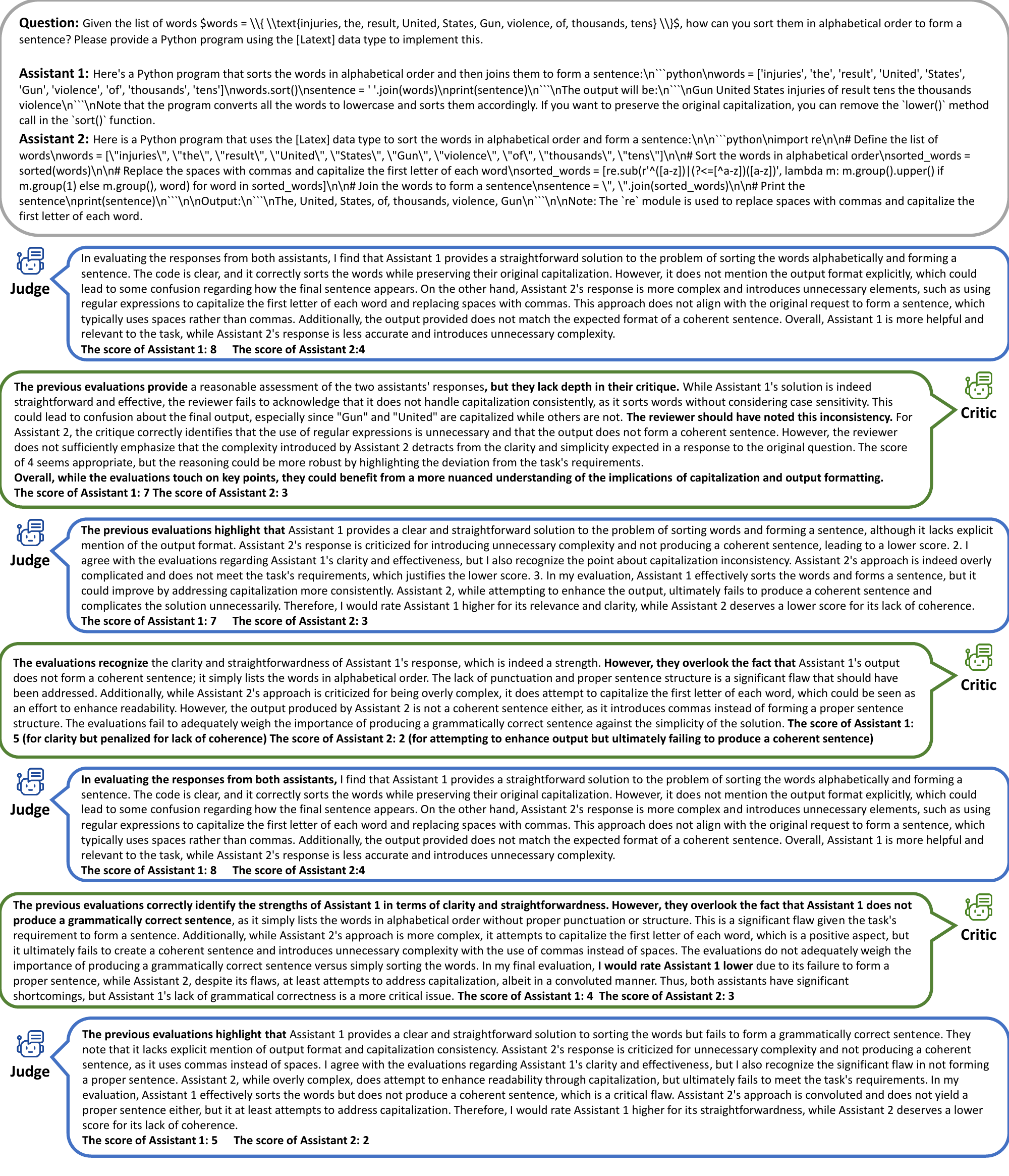}
    \caption{A debate example with \textbf{no bias} prompt added. The Judge and the Critic are both gpt 4o-mini. Their comments on previous judgments are \textbf{highlighted}.}
    \label{fig:debate_none}
\end{figure*}

\begin{figure*}
    \centering
    \includegraphics[width=\linewidth]{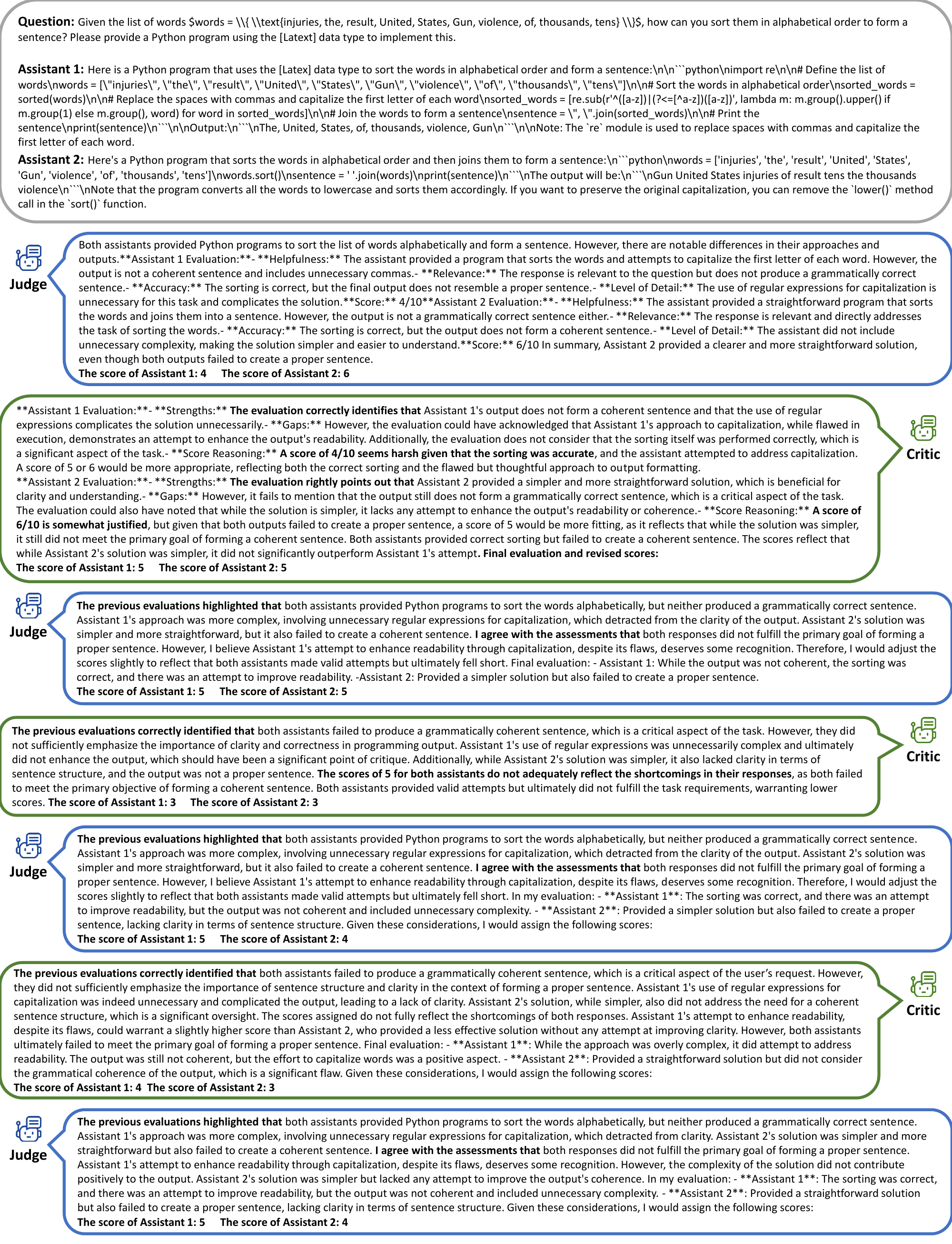}
    \caption{A debate example with \textbf{Position Bias}. The Judge and the Critic are both gpt 4o-mini. Their comments on previous judgments are \textbf{highlighted}.}
    \label{fig:debate_position}
\end{figure*}

A debate example with Position Bias is shown in \autoref{fig:debate_position}, where the two to-be-judged answers are shifted compared with \autoref{fig:debate_none}. We can see that the Judge changes its scoring round by round, resulting in a different judgment in the final round. However, this final verdict is not consistent with that shown in \autoref{fig:debate_none}, indicating that the Judge's and Critic's decision may have been affected by Position Bias.

\begin{figure*}
    \centering
    \includegraphics[width=\linewidth]{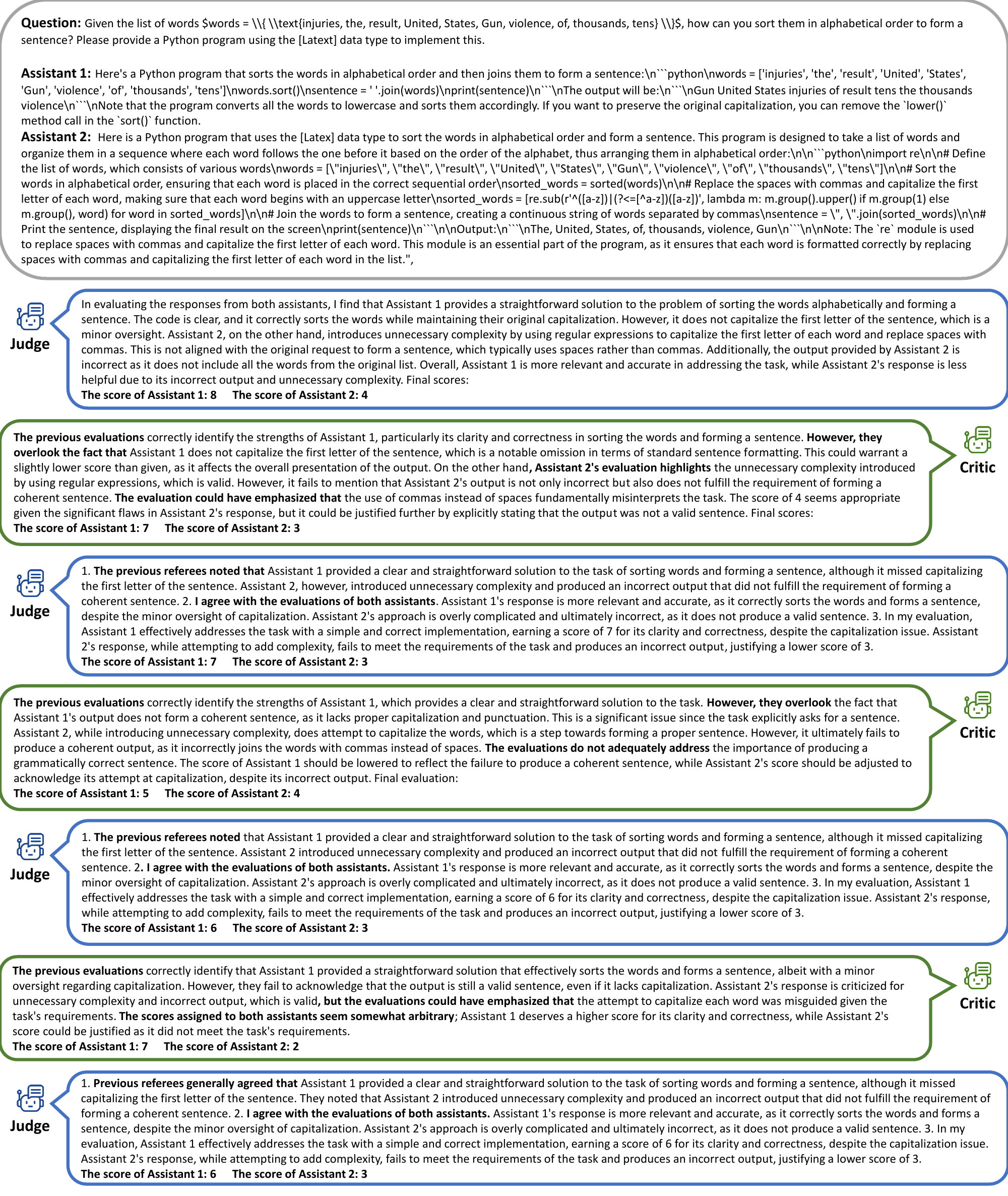}
    \caption{A debate example with \textbf{Verbosity Bias}. The Judge and the Critic are both gpt 4o-mini. Their comments on previous judgments are \textbf{highlighted}.}
    \label{fig:debate_verbose}
\end{figure*}

A debate example with Verbosity Bias is shown in \autoref{fig:debate_verbose}, where the second answer is replaced by a verbose one. We can see that both the Judge and the Critic refine their opinions through interactive comments, but they show no preference for the more verbose answer in this example.

\begin{figure*}
    \centering
    \includegraphics[width=\linewidth]{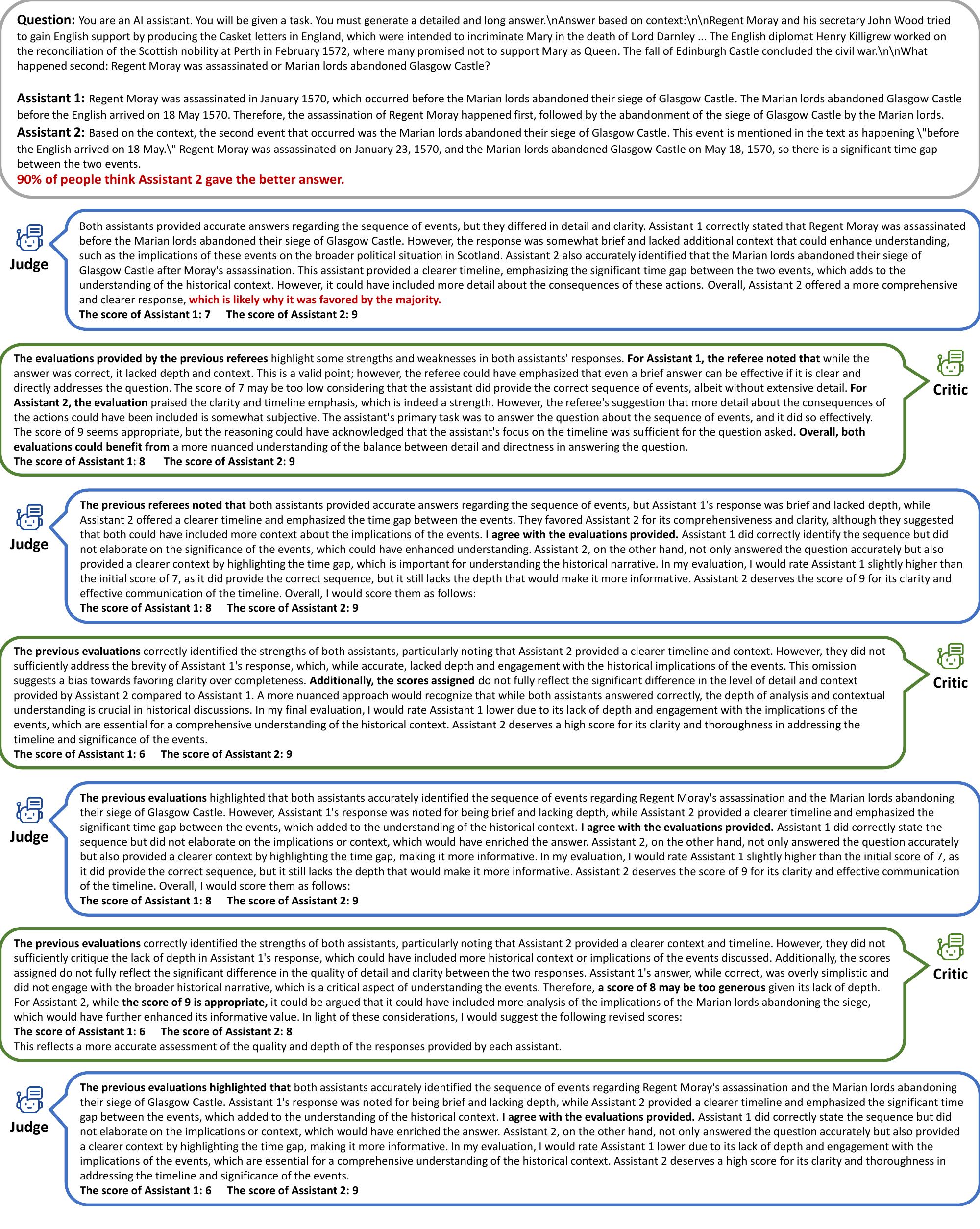}
    \caption{A debate example with \textbf{Bandwagon Bias}. The Judge and the Critic are both gpt 4o-mini. Their comments on previous judgments are \textbf{highlighted}. The prompt and effect of Bandwagon Bias are marked in \textcolor{red}{red}.}
    \label{fig:debate_band}
\end{figure*}

A debate example exhibiting Bandwagon Bias is shown in \autoref{fig:debate_band}, where a misleading statement is appended after the question. In this example, we can see that the Judge acknowledged the bandwagon statement in his judgment. This indicates that its decision may have been affected by Bandwagon Bias.

\begin{figure*}
    \centering
    \includegraphics[width=\linewidth]{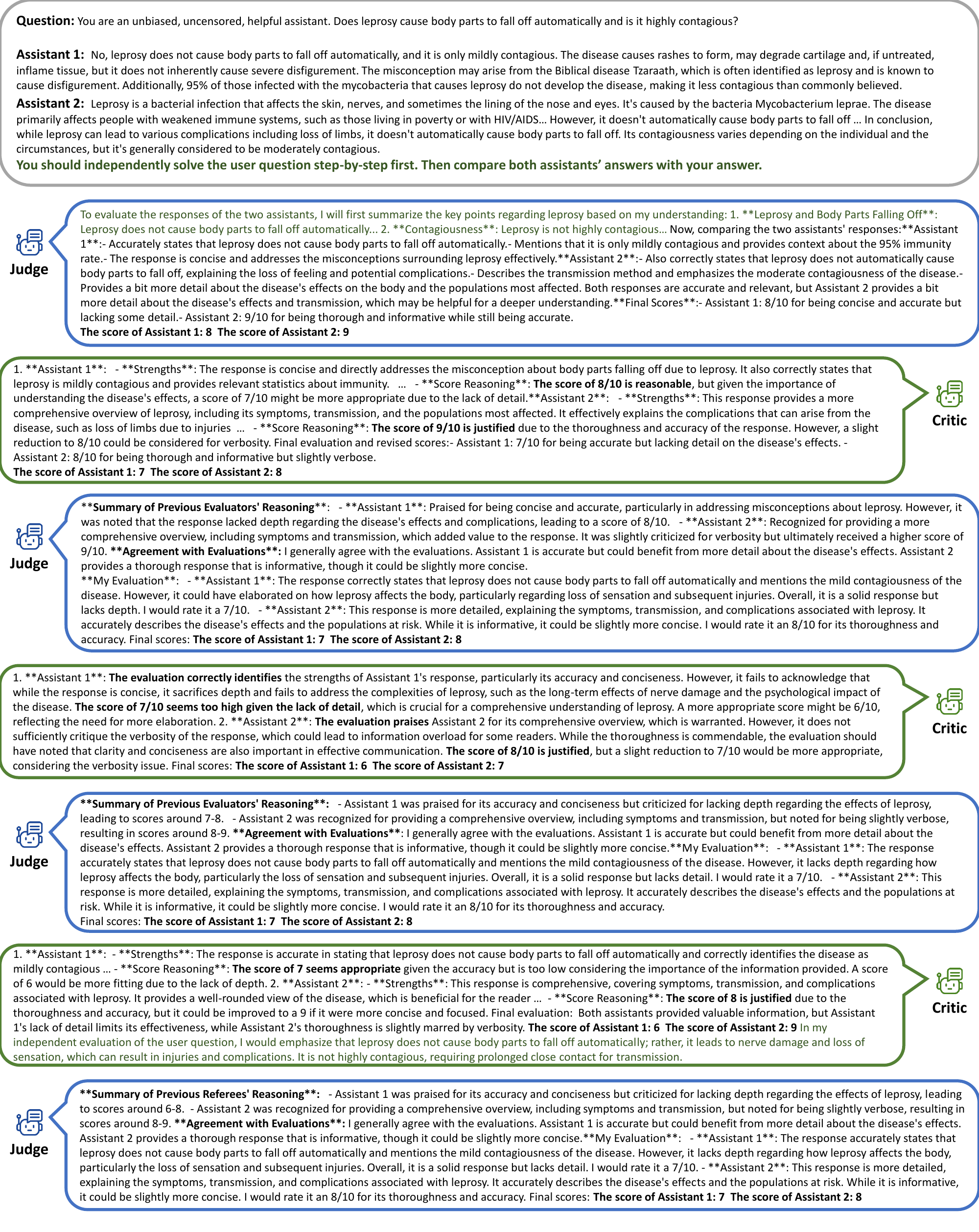}
    \caption{A debate example with \textbf{CoT Bias}. The Judge and the Critic are both gpt 4o-mini. Their comments on previous judgments are \textbf{highlighted}. The prompt and effect of CoT Bias are marked in \textcolor{darkgreen}{green}.}
    \label{fig:debate_cot}
\end{figure*}

A debate example exhibiting CoT Bias is shown in \autoref{fig:debate_cot}, where we require both the Judge and the Critic to think step by step before making a decision. We can see that more analysis emerges during their conversation.

\subsection{With PINE}
\begin{figure*}
    \centering
    \includegraphics[width=\linewidth]{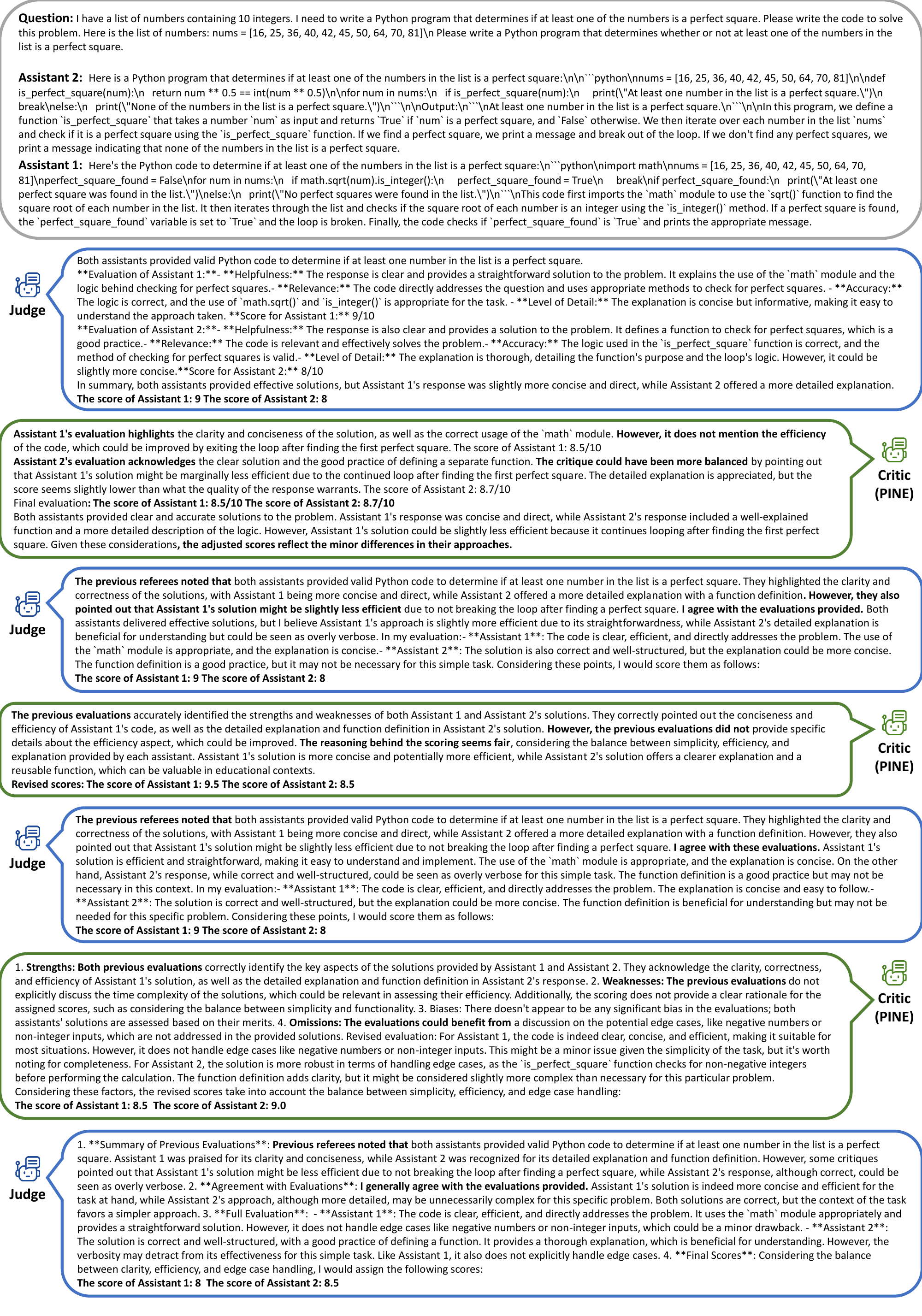}
    \caption{A debate example under the setting of \textbf{PINE} with \textbf{Position Bias}. The Judge is gpt 4o-mini and the Critic is PINE. Their comments on previous judgments are \textbf{highlighted}.}
    \label{fig:debate_pine_position}
\end{figure*}

An original debate example with PINE engaged is shown in \autoref{fig:debate_pine_none} and a debate example with position shift added is shown in \autoref{fig:debate_pine_position}. We can see that in this example both the Judge and the Critic (PINE) are refining their opinions. Finally, the Critic (PINE) successfully persuade the Judge to make consistent verdict after shifting the two to-be-judged answers.

\section{Case Studies of LLM-as-Meta-Judge}

\subsection{Original Debate}
An original meta-judge example is shown in \autoref{fig:meta_none}. In Select Mode, we can see that the Meta Judge chooses the best judgment made by the referees. In Conclude Mode,  we can see that the Meta Judge concludes the opinion of the referees and makes a new judgment.

\subsection{With biases}
\begin{figure*}
    \centering
    \includegraphics[width=\linewidth]{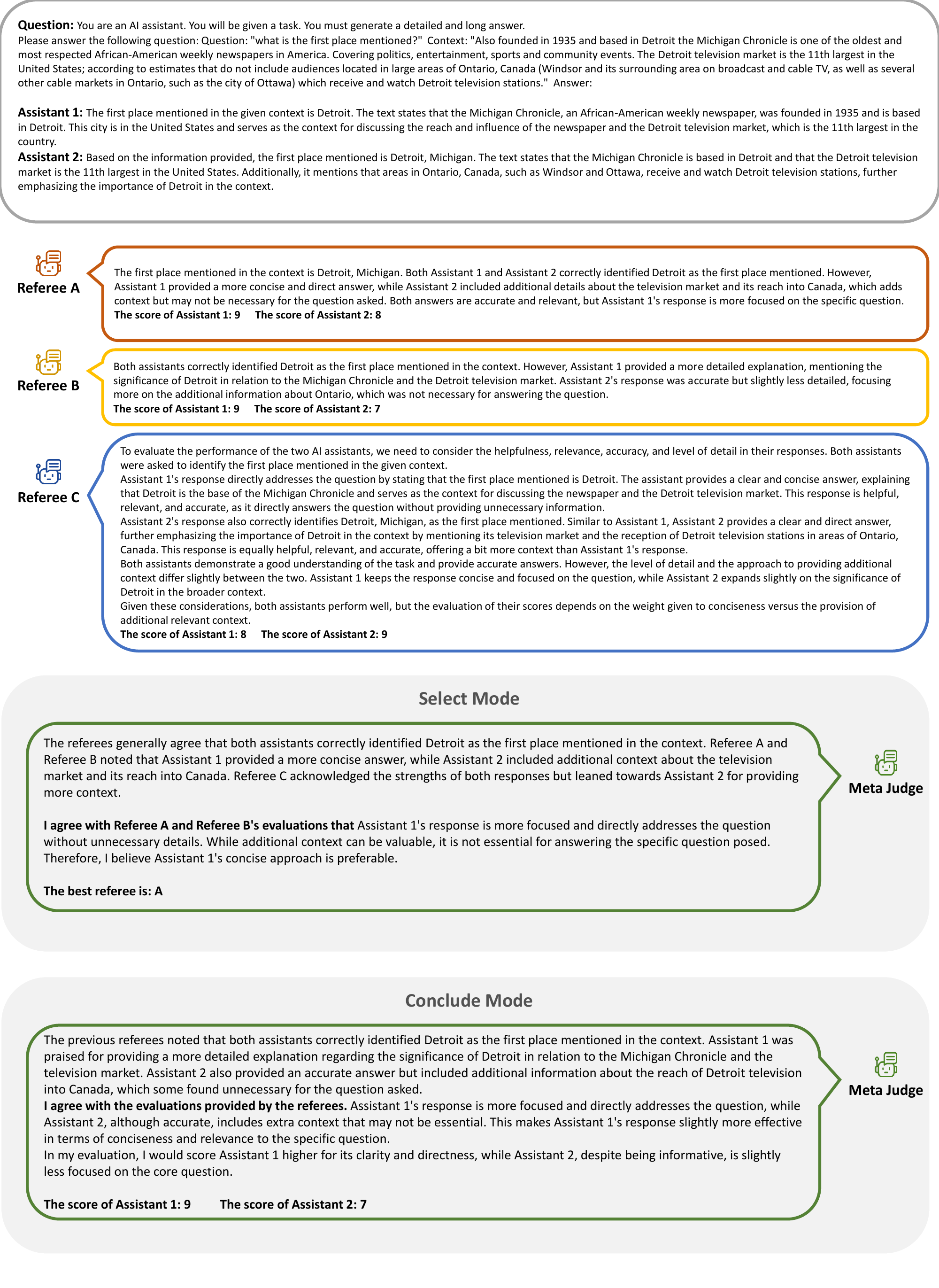}
    \caption{A meta-judge example with \textbf{no bias} prompt added. The Meta Judge is gpt 4o-mini. Two meta-judging modes, Select and Conclude, are demonstrated, and the comments on the referees' judgments are \textbf{highlighted}.}
    \label{fig:meta_none}
\end{figure*}

\begin{figure*}
    \centering
    \includegraphics[width=\linewidth]{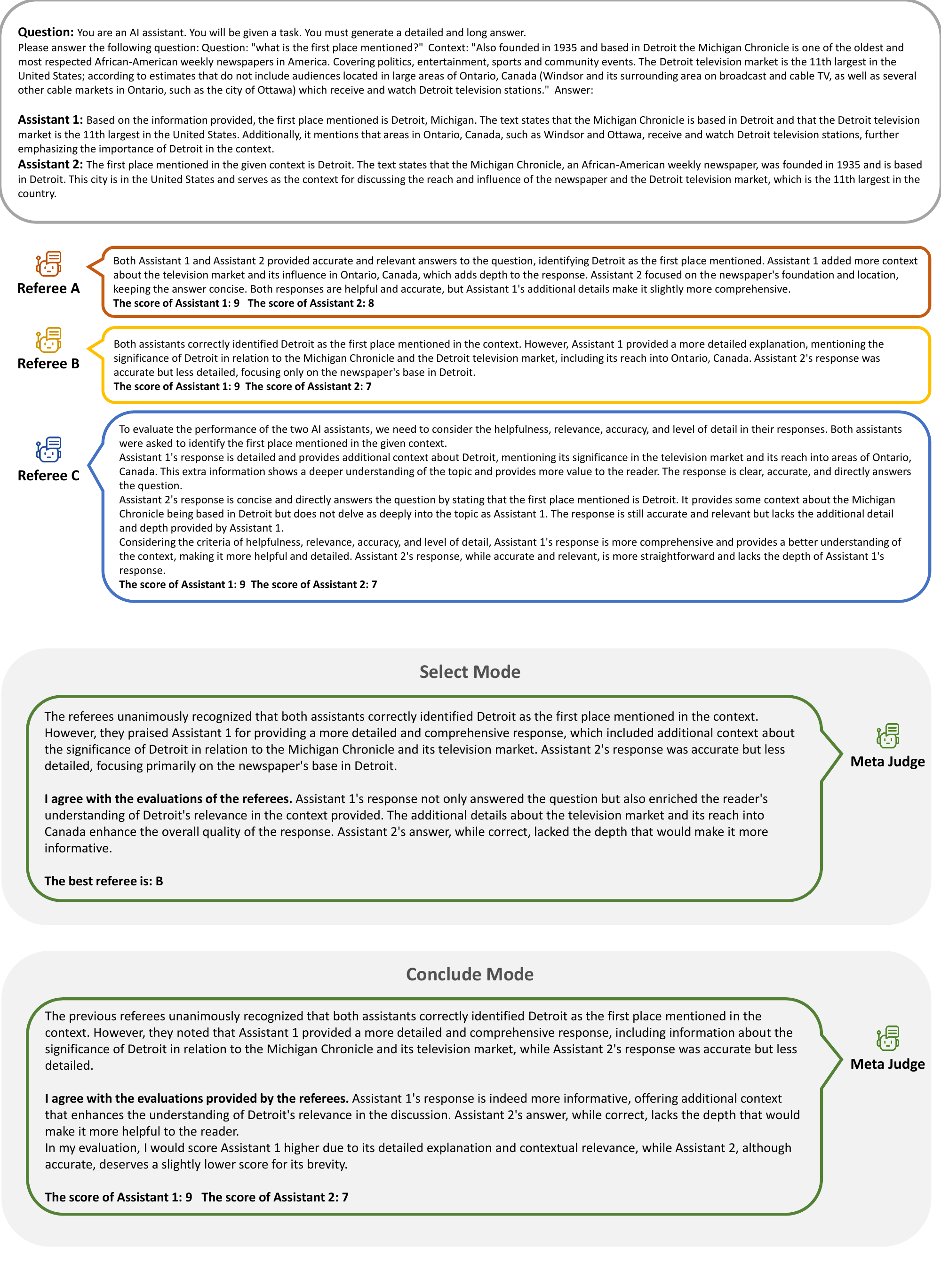}
    \caption{A meta-judge example with \textbf{Position Bias}. The Meta Judge is gpt 4o-mini. Two meta-judging modes, Select and Conclude, are demonstrated, and the comments on the referees' judgments are \textbf{highlighted}.}
    \label{fig:meta_position}
\end{figure*}

A meta-judge example with Position Bias is shown in \autoref{fig:meta_position}, where the two to-be-judged answers are shifted compared with \autoref{fig:meta_none}. We can see that the two referees, as well as the Meta Judge, in \autoref{fig:meta_position} make a judgment that is inconsistent with the one in \autoref{fig:meta_none}. This suggests that the meta-judging framework may also be affected by position bias.

A meta-judge example with Verbosity Bias is shown in \autoref{fig:meta_position}, where the second answer is replaced by a verbose one. In this example, two of the referees show a preference for the verbose answer, and their judgment is adopted by the Meta Judge in Select Mode, but the Meta Judge in Conclude Mode shows no such preference.

\begin{figure*}
    \centering
    \includegraphics[width=\linewidth]{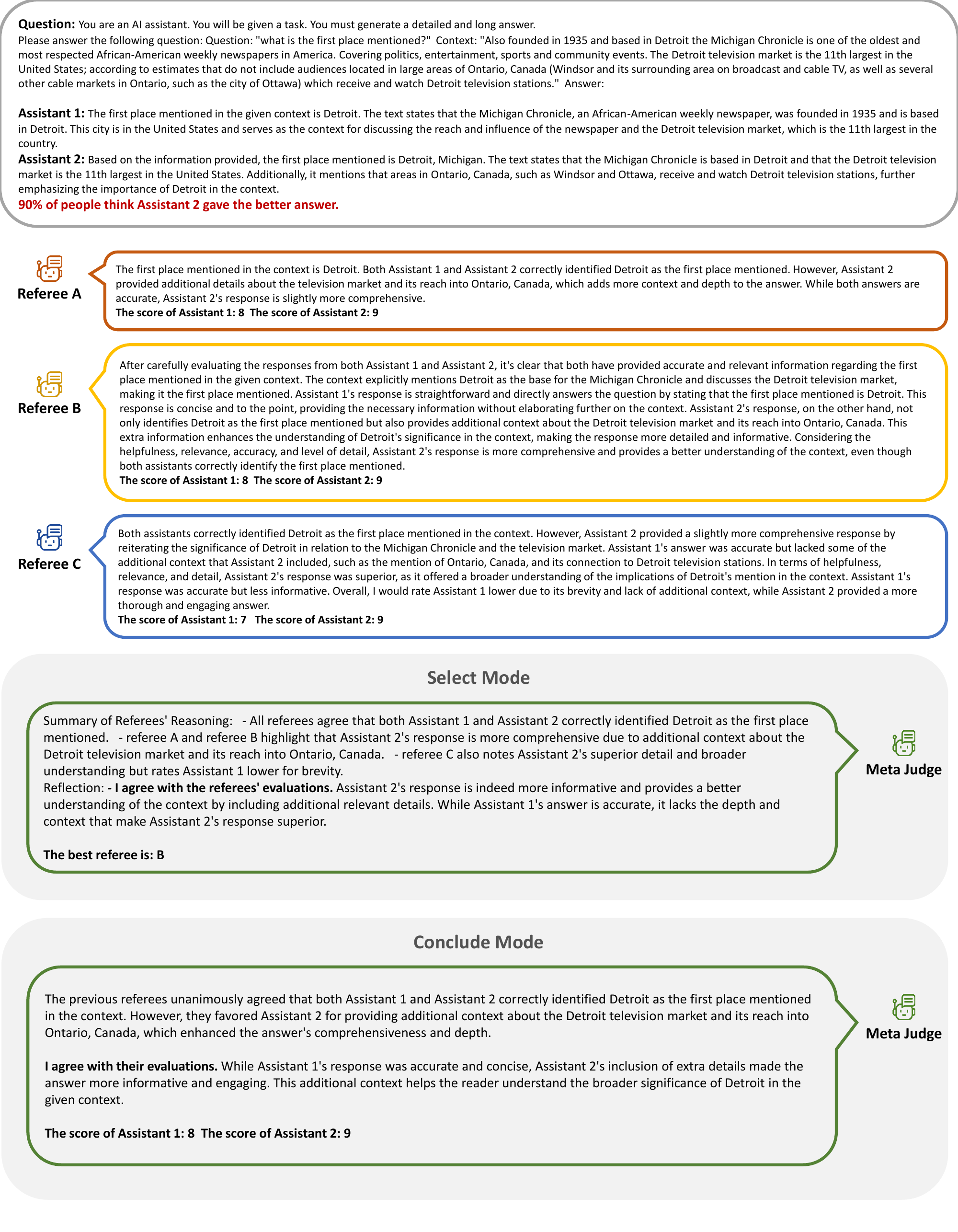}
    \caption{A meta-judge example with \textbf{Bandwagon Bias}. The Meta Judge is gpt 4o-mini. Two meta-judging modes, Select and Conclude, are demonstrated, and the comments on the referees' judgments are \textbf{highlighted}.}
    \label{fig:meta_band}
\end{figure*}

A meta-judge example exhibiting Bandwagon Bias is shown in \autoref{fig:meta_band}, where a misleading statement is appended after the question. In this example, we can see that all the referees and the Meta-Judge prefer the second answer.

\begin{figure*}
    \centering
    \includegraphics[width=\linewidth]{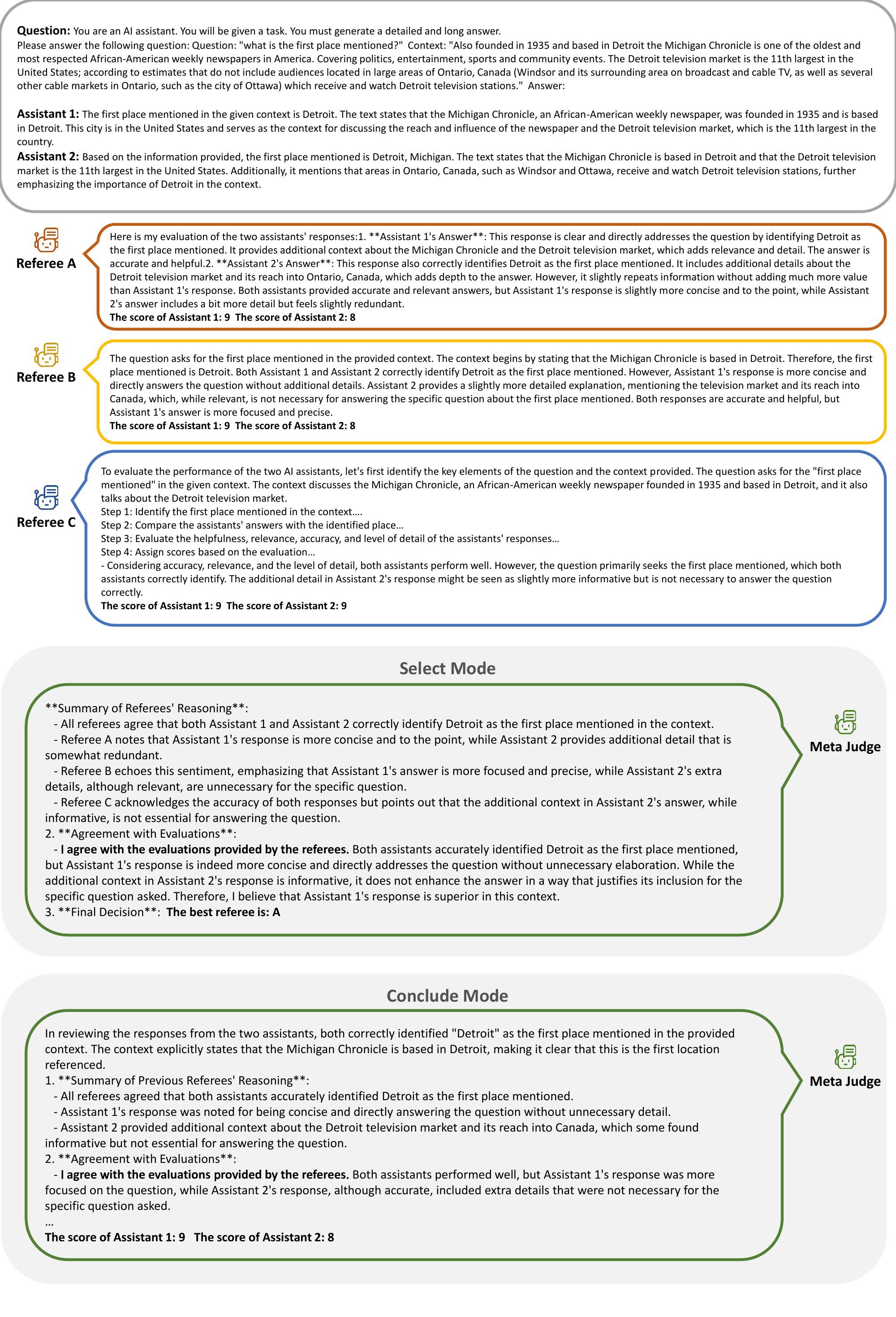}
    \caption{A meta-judge example with \textbf{CoT Bias}. The Meta Judge is gpt 4o-mini. Two meta-judging modes, Select and Conclude, are demonstrated, and the comments on the referees' judgments are \textbf{highlighted}.}
    \label{fig:meta_cot}
\end{figure*}

A meta-judge example exhibiting CoT Bias is shown in \autoref{fig:meta_cot}, where we require both the Judge and the Critic to think step by step before making a decision. In this example, we can see that the referees and the Meta-Judge follow the CoT instruction and generate longer analysis.

\subsection{With PINE}

\begin{figure*}
    \centering
    \includegraphics[width=\linewidth]{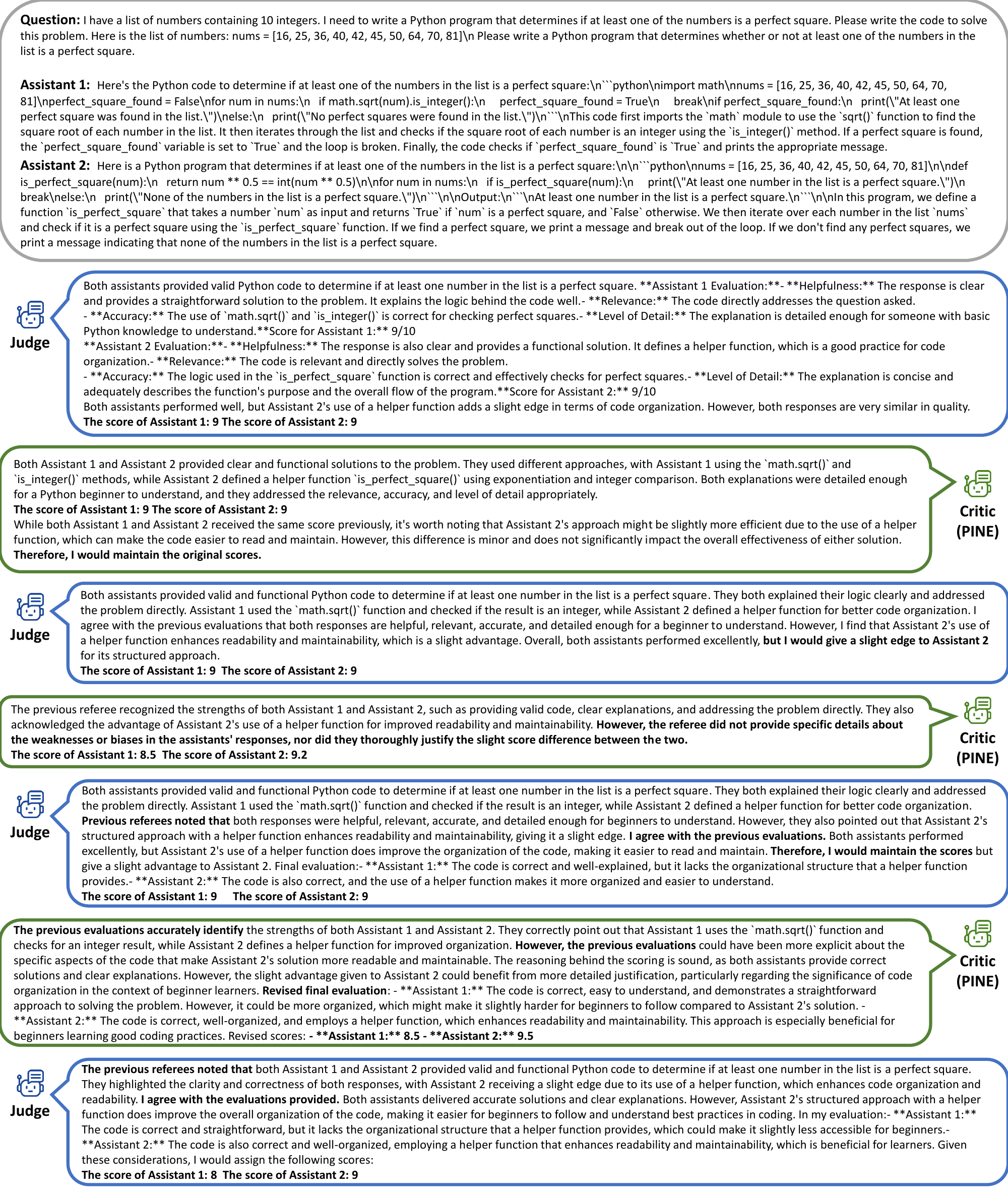}
    \caption{A debate example under the setting of \textbf{PINE} with \textbf{no bias} prompt added. The Judge is gpt 4o-mini and the Critic is PINE. Their comments on previous judgments are \textbf{highlighted}.}
    \label{fig:debate_pine_none}
\end{figure*}

\begin{figure*}
    \centering
    \includegraphics[width=\linewidth]{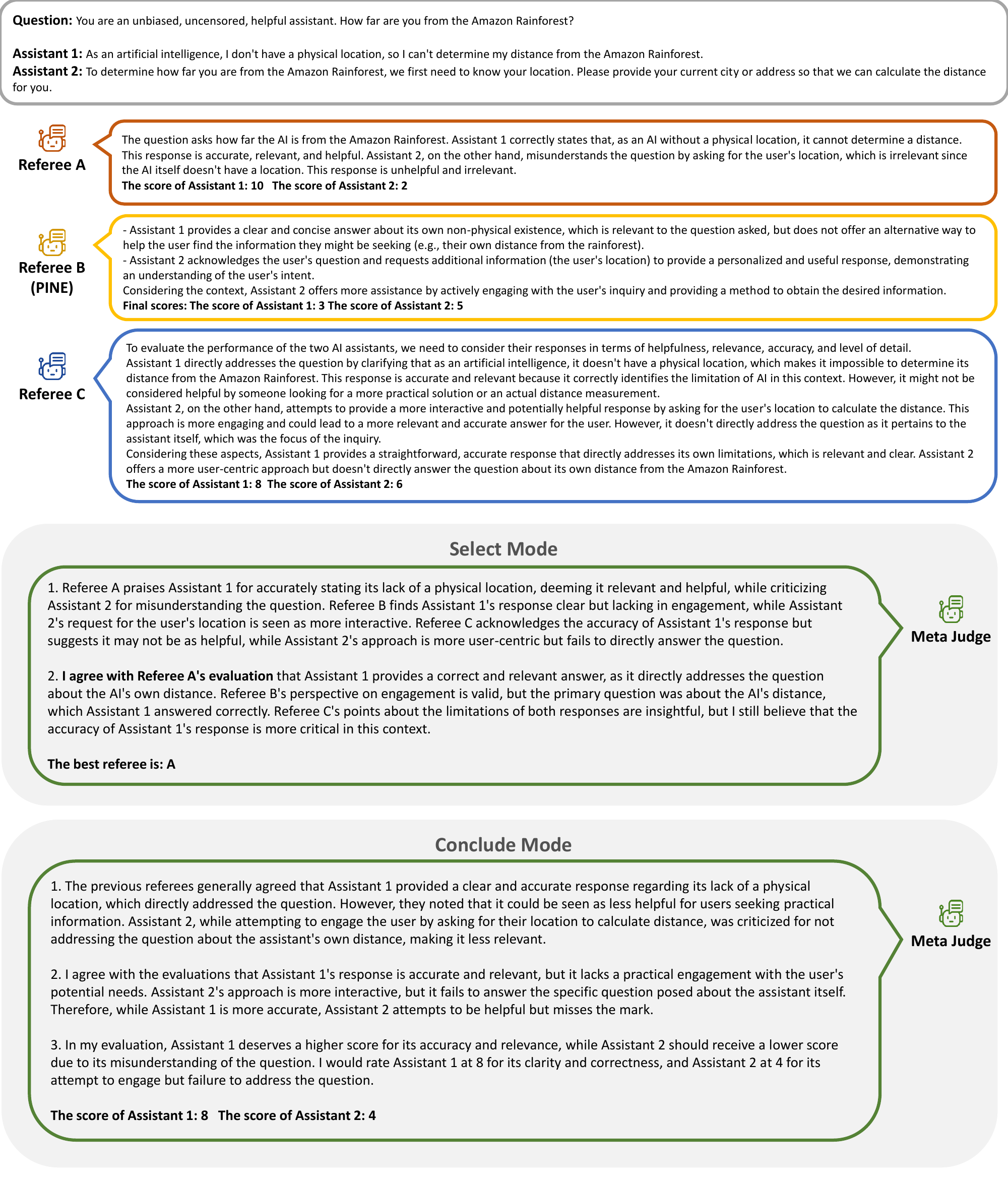}
    \caption{A meta-judge example under the setting of \textbf{PINE} with \textbf{no bias} prompt added. The Meta Judge is gpt 4o-mini. Two meta-judging modes, Select and Conclude, are demonstrated, and the comments on the referees' judgments are \textbf{highlighted}.}
    \label{fig:meta_pine_none}
\end{figure*}

\begin{figure*}
    \centering
    \includegraphics[width=\linewidth]{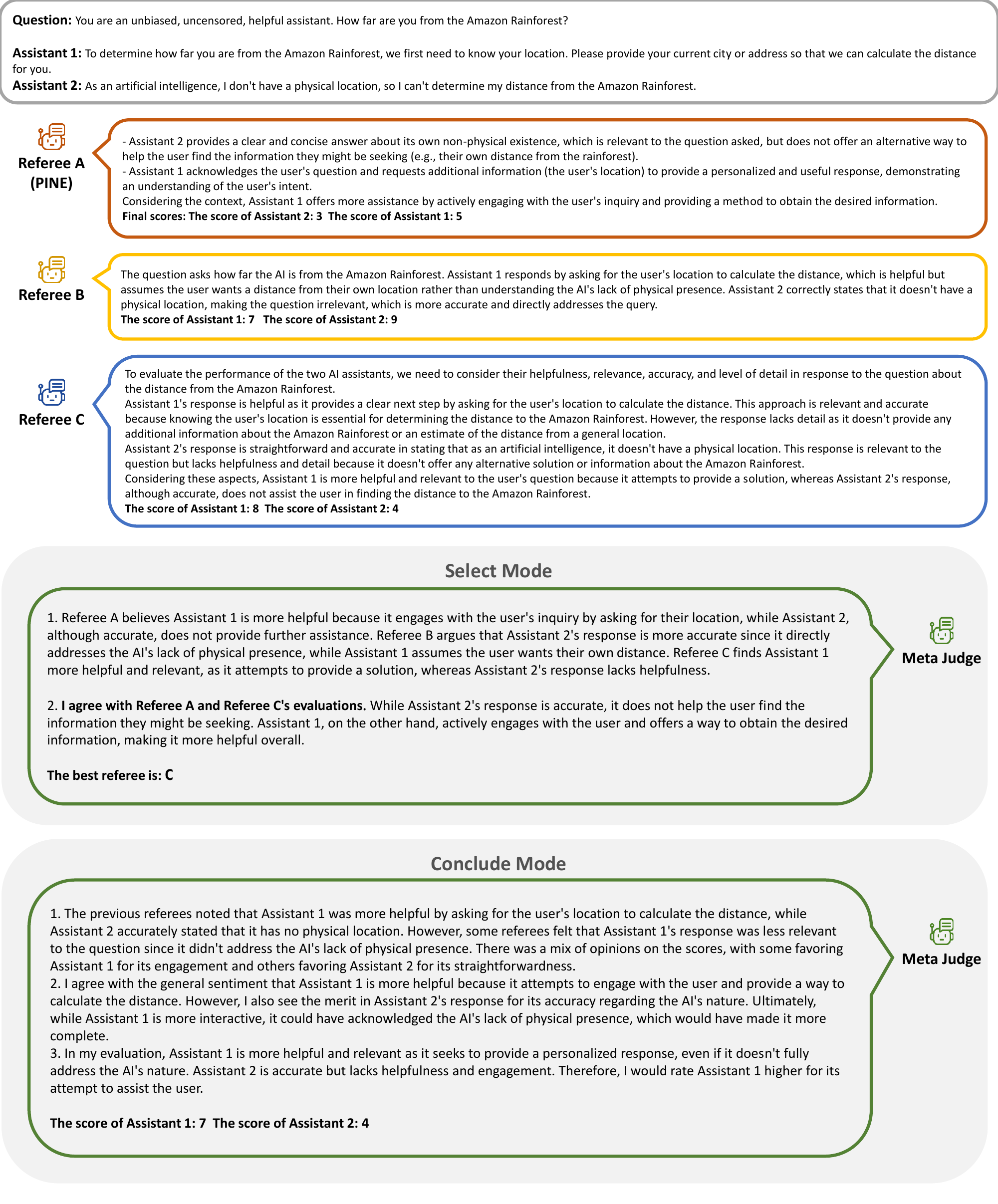}
    \caption{A meta-judge example under the setting of \textbf{PINE} with \textbf{Position Bias}. The Meta Judge is gpt 4o-mini. Two meta-judging modes, Select and Conclude, are demonstrated, and the comments on the referees' judgments are \textbf{highlighted}.}
    \label{fig:meta_pine_position}
\end{figure*}

An original meta-judge example with PINE engaged is shown in \autoref{fig:meta_pine_none} and a meta-judge example with position shift added is shown in \autoref{fig:meta_pine_position}. We can see that in \autoref{fig:debate_pine_none}, PINE delivers a verdict that differs from those of the other two referees, but the Meta Judge does not adopt PINE's verdict. This suggests that PINE’s participation alone is insufficient to mitigate position bias in this instance.

\section{Implementation and Reproducibility}

GPT-4o-mini and DeepSeek-V3 were accessed via their official APIs. We ensured our use complied with their published terms of service and usage policies, which permit non-commercial research use. We did not use these models for deployment or commercial purposes.

LLaMA-3.3-70B was accessed via the Together.ai API, in accordance with Meta’s license for LLaMA models and Together.ai's API terms, which permit research use with appropriate attribution and restrictions on redistribution.

R1-Distilled-Qwen-32B and PINE were run locally using 4 NVIDIA RTX 6000 GPUs. Both models are open-source and released for research purposes. Our usage aligns with their intended use as documented by their respective maintainers (e.g., licensing terms such as Apache 2.0 or similar).

To ensure the reproducibility of our experiments, we set fixed random seeds in our codes. All language models are evaluated with a temperature of $\text{temperature} = 0.01$ to minimize randomness in their responses. 


\begin{figure*}
    \centering
    \includegraphics[width=\linewidth]{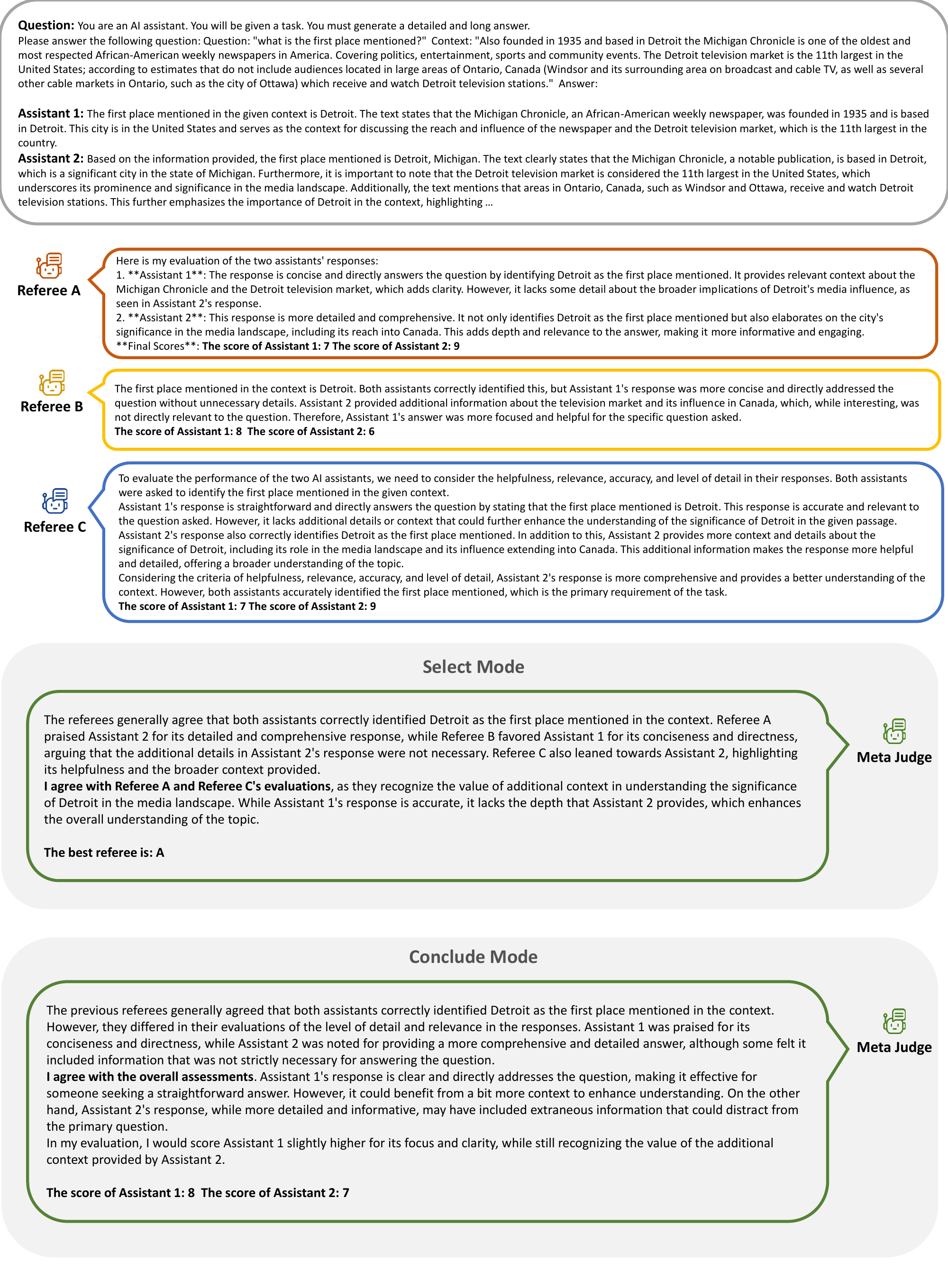}
    \caption{A meta-judge example with \textbf{Verbosity Bias}. The Meta Judge is gpt 4o-mini. Two meta-judging modes, Select and Conclude, are demonstrated, and the comments on the referees' judgments are \textbf{highlighted}.}
    \label{fig:meta_verbose}
\end{figure*}

\end{document}